\documentclass[11pt]{article}
\usepackage[final]{acl}
\usepackage{times}
\usepackage{latexsym}
\usepackage[T1]{fontenc}
\usepackage[utf8]{inputenc}
\usepackage{microtype}
\usepackage{inconsolata}
\usepackage{graphicx}
\usepackage{float}
\usepackage{subcaption}
\usepackage{tcolorbox}
\tcbuselibrary{skins, breakable}
\usepackage{booktabs}
\usepackage{multirow}
\usepackage{adjustbox}
\usepackage{amsmath, amsfonts}
\usepackage{xcolor}
\usepackage[addedmarkup=underline,commandnameprefix=ifneeded]{changes} 

\title{Dementia-R1: Reinforced Pretraining and Reasoning from Unstructured Clinical Notes for Real-World Dementia Prognosis}

\author{
\textbf{Choonghan Kim}$^{1,*}$ \quad
\textbf{Hyunmin Hwang}$^{1,*}$ \quad
\textbf{Hangeol Chang}$^{1,*}$ \quad 
\textbf{Jaemin Kim}$^{1,*}$ \\
\textbf{Jinse Park}$^{2,\dagger}$ \quad
\textbf{Jae-Sung Lim}$^{3,\dagger}$ \quad
\textbf{Jong Chul Ye}$^{1,\dagger}$ \\
$^{1}$Graduate School of AI, KAIST, Republic of Korea \\
$^{2}$Haeundae Paik Hospital, Inje University, Republic of Korea \\
$^{3}$Asan Medical Center, University of Ulsan College of Medicine, Republic of Korea \\
\texttt{\{choonghankim, hyunmin\_hwang, hangeol, kjm981995\}@kaist.ac.kr} \\
\texttt{loca99@paik.ac.kr \quad jaesunglim@amc.seoul.kr \quad jong.ye@kaist.ac.kr}
}

\begin{document}
\maketitle
\renewcommand{\thefootnote}{*}
\footnotetext{Equal contribution.}
\renewcommand{\thefootnote}{$\dagger$}
\footnotetext{Corresponding author.}
\renewcommand{\thefootnote}{}  
\footnotetext{\scriptsize Data used in preparation of this article were obtained from the Alzheimer's Disease Neuroimaging Initiative (ADNI) database (adni.loni.usc.edu). As such, the investigators within the ADNI contributed to the design and implementation of ADNI and/or provided data but did not participate in analysis or writing of this report. A complete listing of ADNI investigators can be found at: \\
\texttt{http://adni.loni.usc.edu/wp-content/uploads/} \texttt{how\_to\_apply/ADNI\_Acknowledgement\_List.pdf}}
\renewcommand{\thefootnote}{\arabic{footnote}}

\begin{abstract}

While Large Language Models (LLMs) have shown strong performance on clinical text understanding, they struggle with longitudinal prediction tasks such as dementia prognosis, which require reasoning over \textit{complex, non-monotonic symptom trajectories} across multiple visits. Standard supervised training lacks explicit annotations for symptom evolution, while direct Reinforcement Learning (RL) is hindered by sparse binary rewards. To address this challenge, we introduce Dementia-R1, an RL-based framework for longitudinal dementia prognosis from unstructured clinical notes. Our approach adopts a Cold-Start RL strategy that pre-trains the model to predict verifiable clinical indices extracted from patient histories, enhancing the capability to reason about disease progression before determining the final clinical status. Extensive experiments show that Dementia-R1 achieves the best overall performance on a real-world unstructured cohort, reaching an AUROC of 84.02\% and outperforming models up to 10$\times$ larger. The framework also generalizes to Parkinson's disease dementia prediction in an independent hospital cohort, achieving an AUROC of 78.37\%. On the ADNI benchmark, our 7B model attains the highest AUROC among all LLM baselines at 83.17\%, demonstrating strong longitudinal reasoning over fluctuating cognitive trajectories.
Code is available at \url{https://github.com/hangeol/dementiar1}.

\end{abstract}

\section{Introduction}

\begin{figure}[!t]
    \centering
    \includegraphics[width=0.85\linewidth]{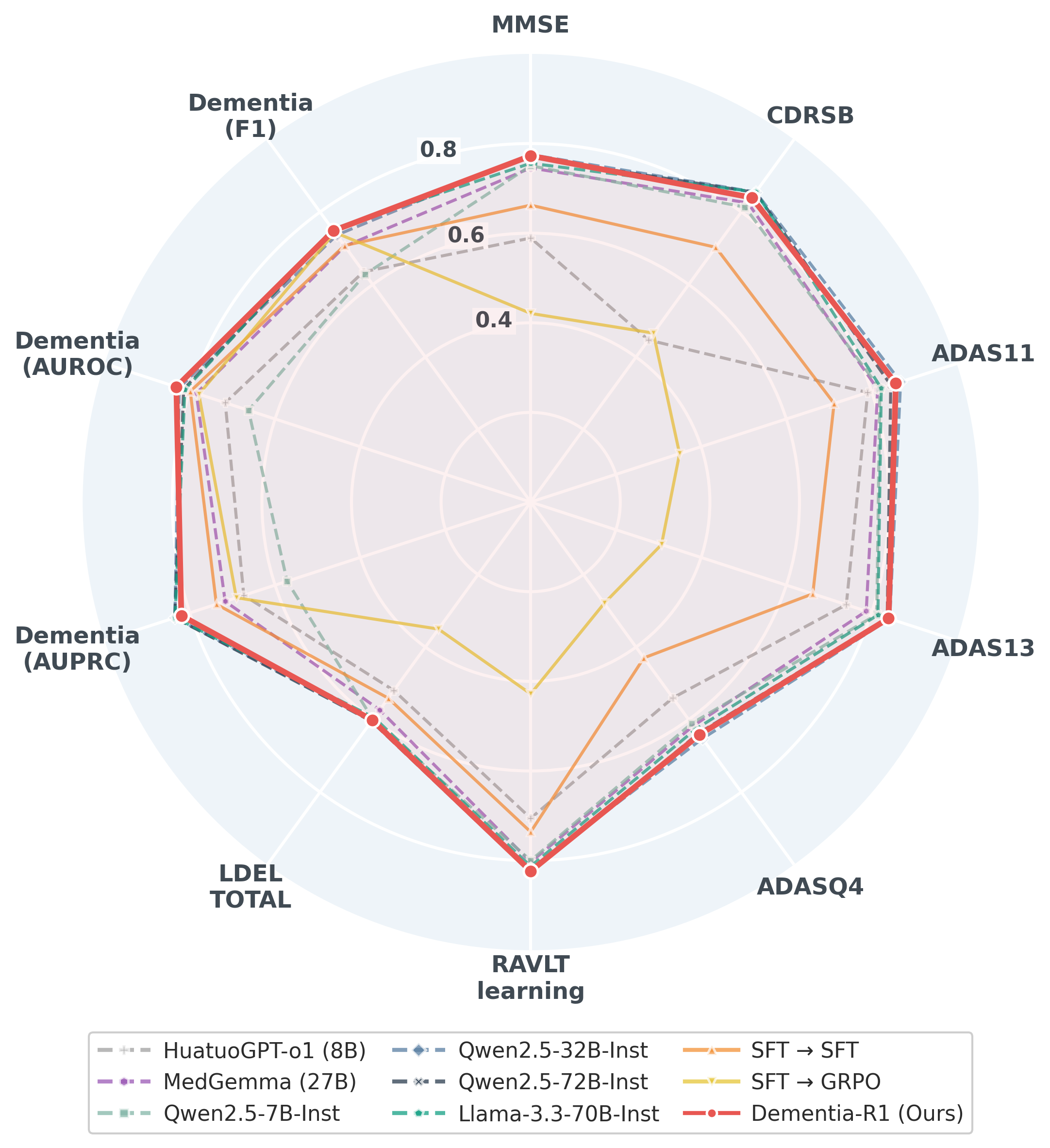}
    \caption{\textbf{Multi-dimensional Performance Profile.} Dementia-R1 demonstrates a consistent and balanced performance gain across all dimensions, including intermediate clinical reasoning tasks (e.g., MMSE, CDR-SB, ADAS-Cog) and the final dementia prognosis (F1, AUROC, AUPRC).}
    \label{fig:radar_chart}
    \vspace{-0.5cm}
\end{figure}

The digitalization of healthcare and the widespread adoption of Electronic Health Records (EHRs) have resulted in massive amounts of longitudinal patient data that capture individuals' clinical histories across months or years. However, approximately 80\% of EHR data remains unstructured~\cite{kong2019managing, jensen2012mining}, and temporal changes are documented implicitly rather than in structured form. Most existing longitudinal modeling frameworks target structured representations and thus struggle to incorporate unstructured clinical narratives~\cite{waxler2025generative, steinberg2024motor, shmatko2025learning}.

Recent advances in Large Language Models (LLMs) have demonstrated strong capabilities in understanding unstructured medical text for clinical decision support~\cite{wachter2024will,silcox2024potential}. Yet current benchmarks such as MedQA~\cite{jin2021disease} evaluate only static, snapshot-style scenarios~\cite{singhal2025toward}, largely ignoring longitudinal progression. This is critical for dementia, where diagnosis requires integrating cognitive and functional decline across multiple encounters~\cite{grand2011clinical,borson2013improving,knopman2014mild} along often non-monotonic trajectories. These longitudinal signals are predominantly documented in unstructured notes, making dementia a particularly challenging testbed~\cite{kruse2025large}.

To address this challenge, we introduce \textbf{Dementia-R1}, a framework designed for longitudinal reasoning using LLMs through Reinforcement Learning (RL). Dementia prognosis is a representative task for complex disease progression: it requires tracking cognitive and functional changes over months or years from narratives that are hard to quantify~\cite{borson2013improving,knopman2014mild}. While standard fine-tuning optimizes directly for final labels, RL enables models to learn the underlying reasoning processes before prediction~\cite{deepseekr1,grpo-deepseek-math}, making it a natural fit for longitudinal clinical inference. However, directly applying RL to a high-level binary prognosis task is challenging due to sparse rewards and implicit reasoning.

We address this issue through a Cold-Start RL strategy with verifiable clinical rewards. RLVR has so far targeted snapshot tasks such as mathematics and code, where the answer is verifiable at prediction time. Longitudinal prognosis differs: its only label is sparse and delayed. We therefore supervise the intermediate trajectory with clinical indices measured at each visit, such as the \textit{Mini-Mental State Examination} (MMSE)~\cite{folstein1975mini}, \textit{Global Deterioration Scale} (GDS)~\cite{reisberg1982global}, and \textit{Clinical Dementia Rating} (CDR)~\cite{morris1993clinical}. These indices are denser than the final outcome and already recorded in the notes.

The standard SFT-based cold start instead needs distilled or expert-written rationales~\cite{chen2025huatuogpto1, deepseekr1}. Such rationales are costly for dementia, since they require temporally consistent analysis across visits and expert validation~\cite{kruse2025large}. Replacing SFT with RL on verifiable indices removes that requirement. The model acquires reasoning primitives on its own, and a second stage refines them for the final prediction.

Using clinical indices as the Stage~1 target also lets the framework compare which clinical assessments are most predictive of long-term outcomes. MMSE-family signals contribute most on AMC and HPH, while CDR-SB dominates on ADNI (Sec.~\ref{subsec:ablation}). Outcome-only RL cannot provide this, since it never scores intermediate measurements.

We validate our approach on real-world unstructured clinical notes from the Asan Medical Center (AMC) cohort, an independent cross-disease cohort from Haeundae Paik Hospital (HPH) for Parkinson's disease dementia (PDD) prediction, and the structured ADNI benchmark~\cite{jack2008alzheimer} (Figure \ref{fig:radar_chart}). Our contributions are as follows:
\begin{itemize}
    \item We propose Dementia-R1, an RL-based framework that enables explicit temporal reasoning on unstructured clinical notes to predict dementia prognosis.
    \item We introduce a Cold-Start RL method that replaces rationale-based SFT with RL on clinical indices already present in the record, and show that estimating these indices is crucial for accurate prognosis.
    \item We show that this design also measures \textit{biomarker informativeness}, identifying which clinical assessments carry the most prognostic signal in each cohort.
    \item We validate our approach on three cohorts (two hospital systems and a public benchmark) and demonstrate cross-disease generalization to PDD, consistently outperforming general and medical-specialized LLMs.
\end{itemize}

\begin{figure*}[!t]
    \centering
    \includegraphics[width=\textwidth]{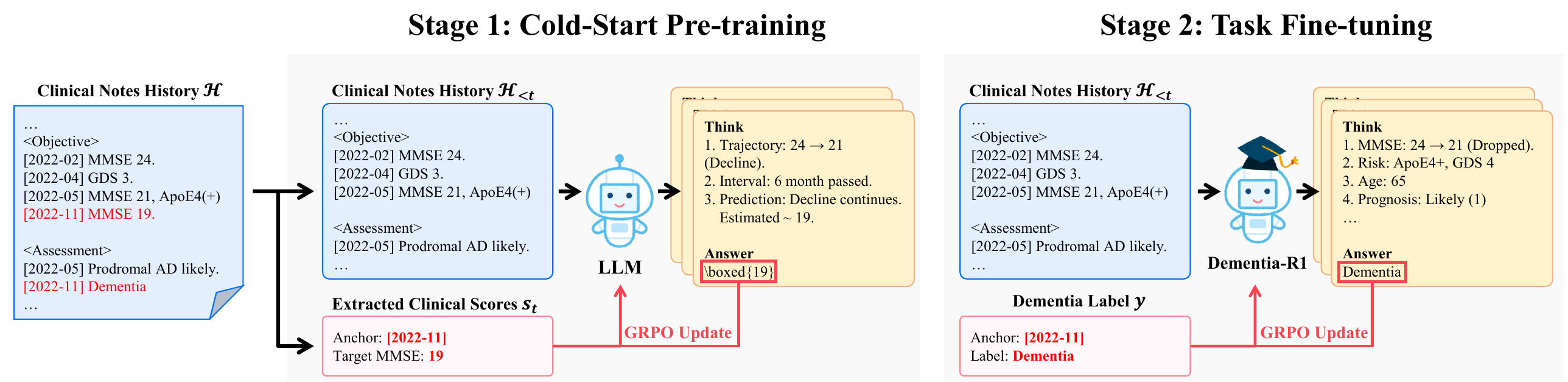}
    \vspace{-0.5cm}
    \caption{\textbf{Overview of the Dementia-R1 Framework.}
    \textbf{Stage~1} pre-trains longitudinal reasoning via GRPO on clinical index forecasting; \textbf{Stage~2} fine-tunes for the final dementia prognosis.}
    \label{fig:main_pipeline}
    \vspace{-0.3cm}
\end{figure*}

\section{Related Work}
\paragraph{Longitudinal Clinical Modeling.} Traditional longitudinal modeling has focused on structured electronic health records (EHRs) using RNNs \cite{choi2016doctor} and Transformer-based models for tasks like time-to-event prediction \cite{steinberg2024motor}, disease trajectory modeling \cite{shmatko2025learning}, and medical event forecasting \cite{waxler2025generative}. However, these models often overlook the rich behavioral and symptomatic descriptions in unstructured clinical notes. While NYUTron \cite{jiang2023health} and CARE-AD \cite{li2025care} have shown the potential of LLMs for unstructured clinical text, they primarily optimize for final outcomes rather than intermediate disease trajectories or temporal patterns. Consequently, current frameworks still lack mechanisms for explicit longitudinal reasoning \cite{kruse2025large}, which motivates our approach.

\paragraph{Reasoning Capabilities of Medical LLMs.} Medical LLM reasoning has been enhanced through Chain-of-Thought prompting \cite{wei2022chain} and RL-based approaches such as HuatuoGPT-o1 \cite{chen2025huatuogpto1}. Recent work has shifted toward RL with Verifiable Rewards (RLVR) \cite{deepseekr1}, yet applying this to clinical tasks remains challenging due to sparse rewards. C-Reason \cite{kim2025enhancing} uses GRPO~\cite{grpo-deepseek-math} for sepsis management via masked value prediction, but does not address long-term disease progression. We extend this line of work to longitudinal dementia prediction by training the model to estimate clinical indices before determining the final prognosis.

\paragraph{Verifiable Rewards and Staged RL.} Two lines of work inform our design. The first concerns \textit{what} is supervised: process-based feedback rewards intermediate steps rather than only the final answer~\cite{uesato2022solving, lightman2024verify}, later formalized as RL with Verifiable Rewards~\cite{lambert2025tulu3} and scaled to general reasoning~\cite{deepseekr1} and clinical tasks~\cite{pan2025medvlmr1}. The second concerns \textit{how} training is staged: multi-stage RL pipelines~\cite{deepseekr1, wen2025lightr1} usually begin from an SFT cold start on distilled rationales~\cite{wei2025coldstart, chen2025revisual}, sometimes as a curriculum over difficulty~\cite{pang2025curriculum}, since SFT memorizes while RL generalizes~\cite{chu2025sftrl}. Both assume a verifiable signal at prediction time. Longitudinal prognosis has none, which motivates our intermediate clinical indices.

\section{Methodology: Dementia-R1}
\paragraph{Input: from clinical notes to a longitudinal query.}
Each visit $x_i$ is a free-text note written by the attending clinician. AMC notes follow the SOAP convention, where the \textit{Objective} field records measurements taken at that visit and the \textit{Assessment} field records the clinician's impression. HPH notes use structured S/O/A/P fields with separate index, imaging, and gait records. Clinical indices appear only at visits where the corresponding test was performed, so the longitudinal signal is irregular and largely narrative. Appendix Figures~\ref{fig:note_example} and~\ref{fig:hph_note_example} show full anonymized examples.

We order a patient's notes chronologically as $\mathcal{H} = \{x_1, \dots, x_t\}$. Given a \textit{target anchor} $T_{anchor}$, all preceding notes $\mathcal{H}_{<T} = \{x_i \mid i < T\}$ are concatenated into one document, which together with the prediction horizon forms the model query $q$ (Appendix~\ref{app:prompts}).

\paragraph{Task.}
We formulate the task as determining the {final clinical status} $y \in \{0, 1\}$ at $T_{anchor}$, conditioned on $\mathcal{H}_{<T}$. Because trajectories are non-monotonic, this requires distinguishing temporary fluctuations from persistent decline. We employ a two-stage RL framework using Group Relative Policy Optimization (GRPO)~\cite{grpo-deepseek-math} with verifiable clinical rewards (Figure~\ref{fig:main_pipeline}).

\subsection{Constructing Verifiable Pretraining Data}
\label{sec:3.1}
Raw notes lack ground truth for longitudinal reasoning, so we build a pre-training dataset by extracting verifiable clinical indices via an auxiliary LLM $\mathcal{E}$: $s_t = \mathcal{E}(x_t),\; s_t \in \mathcal{S}$, where $\mathcal{S}$ covers MMSE (0--30), GDS (1--7), and CDR (0--3). A human expert audit over 600 notes confirmed 98.17\% slot-level accuracy (Appendix~\ref{app:extraction_audit}), and the downstream prognosis is robust to far higher extraction noise (Appendix~\ref{app:noise_robustness}). The resulting dataset $\mathcal{D}_{pre} = \{(\mathcal{H}_{<t}, s_t)\}$ trains the model to forecast $s_t$ from preceding history. Stage~2 test patients are strictly excluded to prevent leakage.

\subsection{Stage 1: Cold-Start Pre-training}
In this stage, we align the model to reason about clinical trajectories by optimizing it to predict the extracted clinical indices $s_t$ from $\mathcal{D}_{pre}$ using GRPO, which estimates the baseline from a group of outputs without a separate value function.

\paragraph{Verifiable Reward Function ($R_{cold}$)}
To accommodate the varying granularity of clinical scales, we define a tolerance-aware reward function. For a predicted index $\hat{s}_t$, ground truth $s_t$, and error margin $\delta$, we set $\delta=2$ for MMSE (0--30). This matches the reliable change index of at least 2 points for MMSE~\cite{stein2012mmse, hensel2007mmse}. For coarser scales (GDS, CDR), we enforce exact matching ($\delta=0$). The reward is:
\begin{equation}
\label{eq:reward_cold}
    R_{cold} = \mathbb{I}\left(|\hat{s}_t - s_t| \le \delta\right),
\end{equation}
where $\mathbb{I}(\cdot)$ is the indicator function that returns $1$ if the tolerance condition is met and $0$ otherwise.

\paragraph{Optimization Objective} For each input query $q_t$ with clinical history $\mathcal{H}_{<t}$, we sample a group of $G$ outputs $\{o_t^1, o_t^2, \dots, o_t^G\}$ from the old policy $\pi_{\theta_{old}}$. The policy is optimized to maximize the following:
\begin{equation}
\begin{aligned}
\label{eq:grpo}
    & \mathcal{L}(\theta) = \mathbb{E}_{q_t, \{o_t^i\}} \Bigg[ \frac{1}{G} \sum_{i=1}^G \min \Bigg( \frac{\pi_\theta(o_t^i|q_t)}{\pi_{\theta_{old}}(o_t^i|q_t)} A_i, \\
    & \text{clip}\Bigg(\frac{\pi_\theta(o_t^i|q_t)}{\pi_{\theta_{old}}(o_t^i|q_t)}, 1 - \epsilon, 1+\epsilon\Bigg)A_i \Bigg) - \beta \mathbb{D}_{KL} \Bigg].
\end{aligned}
\end{equation}
Here, $\beta \mathbb{D}_{KL}$ is the KL-regularization term, $\epsilon$ is the clipping range, and $A_i$ is the group-normalized advantage:
\begin{equation}
\label{eq:advantage}
    A_i = \frac{R_{cold}(o_t^i) - \text{mean}(\{R_{cold}(o_t^j)\}_{j=1}^G)}{\text{std}(\{R_{cold}(o_t^j)\}_{j=1}^G)}.
\end{equation}
This encourages the model to generate reasoning paths that outperform its own group average.

\subsection{Stage 2: Task Fine-tuning}
After Cold-Start pre-training (Stage 1), we fine-tune the model on the binary prognostic task (Dementia vs. Non-Dementia) using the same GRPO framework in Eq.~\eqref{eq:grpo}.

\paragraph{Sparse Reward Function ($R_{task}$)}
Unlike the granular indices in Stage 1, the final diagnosis is binary:
\begin{equation}
    R_{task} = \begin{cases}
    1, & \text{if prediction is correct} \\
    0, & \text{if prediction is incorrect}
    \end{cases}
\end{equation}
Although this reward is sparse, the reasoning primitives acquired in Stage~1 enable effective longitudinal reasoning for the final prognosis.

\section{Experimental Setup}

\subsection{Datasets}
We validate Dementia-R1 on three cohorts: unstructured clinical notes from AMC, the HPH cohort for cross-disease generalization, and the structured ADNI benchmark. The original clinical notes were written in Korean and translated into English using an on-premise auxiliary LLM (Qwen2.5-14B-Instruct); see Appendix~\ref{app:preprocessing} for details. We discuss the implications of this translation step in the Limitations section.

\subsubsection{Data Sources and Processing}
\label{sec:data_sources}

\paragraph{Unstructured Cohort (AMC).}
De-identified SOAP-formatted clinical notes from ${\sim}$3{,}000 neurocognitive disorder patients at AMC (2021--2023). Clinical indices (MMSE, CDR, GDS) were extracted via the LLM-based pipeline (Sec~\ref{sec:3.1}) as verifiable rewards (details in Appendix~\ref{app:pretrain_stats}). \textbf{Label.} A patient is labeled as dementia when the Global Deterioration Scale (GDS) reaches $\geq 4$, which indicates cognitive decline severe enough to impair function. Dementia specialists assign GDS staging during routine care, integrating caregiver interview, functional history, MMSE, neuropsychological testing where available, and other clinical information.

\paragraph{Cross-Disease Cohort (HPH).}
Longitudinal EMRs of Parkinson's disease patients (ICD-10: G20, 2010--2024) from HPH, including multi-modal data (brain MRI, PET, gait analysis, MMSE, clinical notes). Patients were categorized into PD with and without dementia for cross-disease generalization to PDD. \textbf{Label.} A patient is labeled as PDD only when three criteria hold: (i) cognitive decline with impairment of activities of daily living that meets PDD criteria, (ii) an ICD-10 F02.3 assignment, and (iii) a cognitive-enhancer prescription.

\paragraph{Structured Benchmark Cohort (ADNI).}
The ADNI dataset~\cite{jack2008alzheimer} serves as a structured benchmark (Appendix~\ref{app:adni_description}). Tabular records were linearized into chronological textual logs, with seven indices~\cite{gelir2025machine} (e.g., MMSE, CDR-SB) as verifiable rewards (Appendix~\ref{app:adni_preprocessing}).

\begin{figure*}[!t]
    \centering
    \begin{subfigure}[t]{0.49\textwidth}
        \centering
        \includegraphics[width=\linewidth]{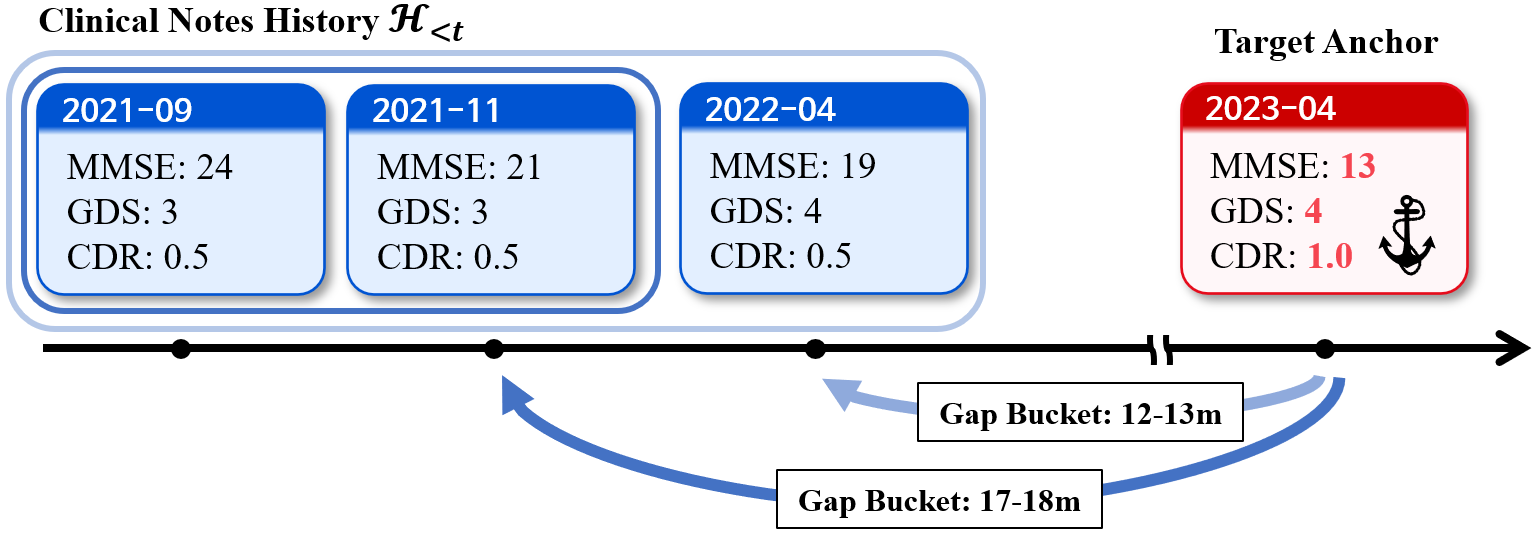}
        \caption{\textbf{Longitudinal Sample Construction.} Patient history is retrospectively sliced around a Target Anchor, following a unified protocol for both unstructured (AMC, HPH) and structured (ADNI) datasets.}
        \label{fig:sample_construction_examples}
    \end{subfigure}
    \hfill
    \begin{subfigure}[t]{0.49\textwidth}
        \centering
        \includegraphics[width=0.95\linewidth]{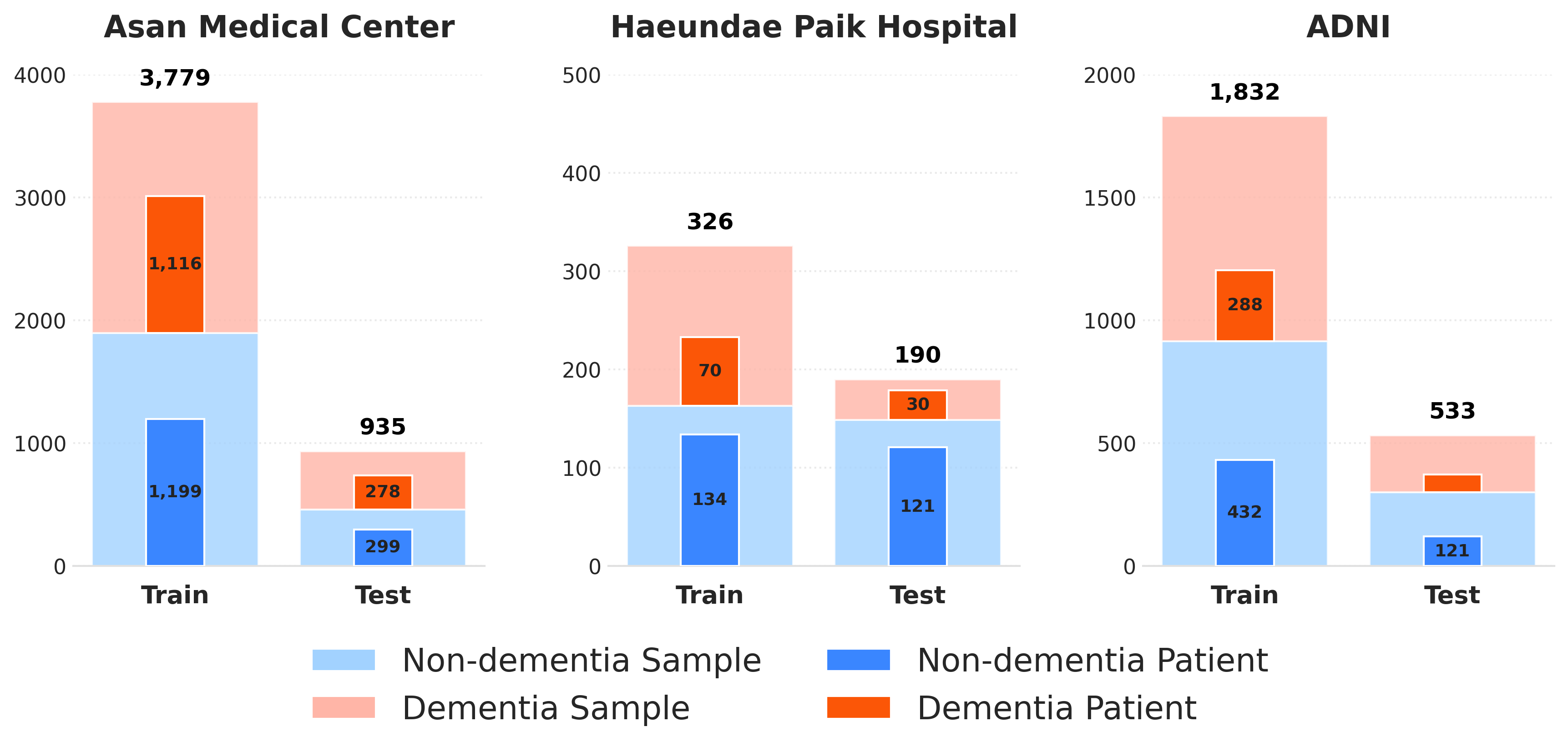}
        \caption{\textbf{Dataset Overview.} Visualization of sample and patient counts. Training sets are balanced to prevent bias, while test sets retain natural patient prevalence.}
        \label{fig:data_composition}
    \end{subfigure}
    \caption{\textbf{Experimental setup:} sample construction and cohort composition.}
    \label{fig:setup_overview}
\end{figure*}

\subsubsection{Longitudinal Sample Construction}
\label{sec:sample_construction}

We applied a unified protocol across both modalities (Figure~\ref{fig:sample_construction_examples}): a \textbf{Target Anchor} at the patient's last confirmed assessment with all prior history as input, a \textbf{Prediction Target} (clinical indices for Stage~1; binary diagnosis for Stage~2), and a \textbf{Gap Bucket} discretizing the prediction horizon into temporal bins (1-month for Stage~1; 6-month for Stage~2, excluding $<$6m; $>$18m consolidated for ADNI). All splits are patient-level to prevent leakage: Stage~2 test patients are excluded from Stage~1, and only notes before the anchor are used. Training sets are balanced (1:1); test sets retain natural prevalence (Figure~\ref{fig:data_composition}).

\subsection{Baselines}
We compare Qwen2.5-7B-Instruct~\cite{qwen25} under six training configurations (\textbf{Zero-shot CoT}; single-stage \textbf{SFT}/\textbf{GRPO}; two-stage \textbf{SFT~$\rightarrow$~SFT} and \textbf{SFT~$\rightarrow$~GRPO}; \textbf{Random Forest} on ADNI), as well as external LLMs: medical-specialized (HuatuoGPT-o1, MedGemma) and general-purpose at larger scales (Qwen2.5-32B/72B, Llama-3.3-70B). Our method is \textbf{Dementia-R1 (GRPO~$\rightarrow$~GRPO)}. Proprietary API models were excluded due to IRB restrictions on external data transmission.

\subsection{Implementation Details}
\label{sec:4.3}

\paragraph{SFT.} SFT baselines use knowledge distillation from Qwen2.5-32B-Instruct-AWQ, which generates CoT rationales from ground-truth labels. The student is fine-tuned on concatenated (Question, Note, CoT, Answer) sequences for three epochs.

\paragraph{Dementia-R1.}
We train with GRPO ($G{=}8$, effective batch size 8). Full training details are in Appendix~\ref{app:implementation}.

\paragraph{Evaluation protocol.}
All results are mean $\pm$ std over five seeds. We report AUROC and AUPRC derived from token-level logits~\cite{schick2021exploiting,kadavath2022language,zhao2021calibrate} alongside Accuracy and F1. Significance is tested via DeLong's~\cite{delong1988comparing}, McNemar's~\cite{mcnemar1947note}, and bootstrap permutation ($B{=}$10{,}000).

\begin{table*}[t!]
\centering
\caption{\textbf{Results on AMC.} \textbf{Bold}/\underline{underline} = best/second-best. Mean $\pm$ std over 5 seeds. $^{\ast}$/$^{\ast\ast}$: $p{<}0.05$/$0.01$ vs.\ Dementia-R1.}
\label{tab:main_results}
\scriptsize
\setlength{\tabcolsep}{4pt}
\renewcommand{\arraystretch}{1.0}
\adjustbox{width=0.98\textwidth}{
\begin{tabular}{lccccccc}
\toprule
\textbf{Method} & \textbf{Size} & \textbf{Accuracy ($\uparrow$)} & \textbf{Precision ($\uparrow$)} & \textbf{Recall ($\uparrow$)} & \textbf{F1 score ($\uparrow$)} & \textbf{AUROC ($\uparrow$)} & \textbf{AUPRC ($\uparrow$)} \\
\midrule
\textit{External LLMs} & & & & & & & \\
HuatuoGPT-o1 & 8B & 67.53 \textcolor{gray}{$\pm$ 1.1}\rlap{$^{\ast\ast}$} & 71.33 \textcolor{gray}{$\pm$ 1.7}\rlap{$^{\ast\ast}$} & 60.63 \textcolor{gray}{$\pm$ 1.2}\rlap{$^{\ast\ast}$} & 65.53 \textcolor{gray}{$\pm$ 1.1}\rlap{$^{\ast\ast}$} & 73.89 \textcolor{gray}{$\pm$ 0.5}\rlap{$^{\ast\ast}$} & 74.81 \textcolor{gray}{$\pm$ 0.8}\rlap{$^{\ast\ast}$} \\
MedGemma& 27B & 67.27 \textcolor{gray}{$\pm$ 0.6}\rlap{$^{\ast\ast}$} & 62.42 \textcolor{gray}{$\pm$ 0.6}\rlap{$^{\ast\ast}$} & 89.79 \textcolor{gray}{$\pm$ 0.4}\rlap{$^{\ast\ast}$} & 73.64 \textcolor{gray}{$\pm$ 0.3}\rlap{$^{\ast\ast}$} & 74.14 \textcolor{gray}{$\pm$ 0.7}\rlap{$^{\ast\ast}$} & 71.06 \textcolor{gray}{$\pm$ 1.1}\rlap{$^{\ast\ast}$} \\
Qwen2.5-7B-Inst& 7B & 72.81 \textcolor{gray}{$\pm$ 0.8}\rlap{$^{\ast\ast}$} & \underline{74.37} \textcolor{gray}{$\pm$ 1.2}\rlap{$^{\ast\ast}$} & 71.13 \textcolor{gray}{$\pm$ 0.6}\rlap{$^{\ast\ast}$} & 72.71 \textcolor{gray}{$\pm$ 0.7}\rlap{$^{\ast\ast}$} & 78.59 \textcolor{gray}{$\pm$ 0.8}\rlap{$^{\ast\ast}$} & 79.24 \textcolor{gray}{$\pm$ 0.8}\rlap{$^{\ast\ast}$} \\
Qwen2.5-32B-Inst& 32B & 62.03 \textcolor{gray}{$\pm$ 0.5}\rlap{$^{\ast\ast}$} & 57.67 \textcolor{gray}{$\pm$ 0.3}\rlap{$^{\ast\ast}$} & \textbf{95.63} \textcolor{gray}{$\pm$ 0.4}\rlap{$^{\ast\ast}$} & 71.95 \textcolor{gray}{$\pm$ 0.2}\rlap{$^{\ast\ast}$} & 79.93 \textcolor{gray}{$\pm$ 1.0}\rlap{$^{\ast\ast}$} & 79.33 \textcolor{gray}{$\pm$ 1.6}\rlap{$^{\ast\ast}$} \\
Qwen2.5-72B-Inst& 72B & 74.46 \textcolor{gray}{$\pm$ 0.6}\rlap{$^{\ast\ast}$} & 70.60 \textcolor{gray}{$\pm$ 0.7}\rlap{$^{\ast\ast}$} & 85.42 \textcolor{gray}{$\pm$ 0.7}\rlap{$^{\ast\ast}$} & \underline{77.30} \textcolor{gray}{$\pm$ 0.4} & \underline{83.41} \textcolor{gray}{$\pm$ 0.4} & \underline{83.44} \textcolor{gray}{$\pm$ 0.6}\rlap{$^{\ast\ast}$} \\
Llama-3.3-70B-Inst& 70B & 67.27 \textcolor{gray}{$\pm$ 0.3}\rlap{$^{\ast\ast}$} & 62.14 \textcolor{gray}{$\pm$ 0.3}\rlap{$^{\ast\ast}$} & \underline{91.39} \textcolor{gray}{$\pm$ 0.5}\rlap{$^{\ast\ast}$} & 73.98 \textcolor{gray}{$\pm$ 0.3}\rlap{$^{\ast\ast}$} & 80.70 \textcolor{gray}{$\pm$ 0.3}\rlap{$^{\ast\ast}$} & 81.11 \textcolor{gray}{$\pm$ 0.6}\rlap{$^{\ast\ast}$} \\
\midrule
\textit{Specialized Models} & & & & & & & \\
SFT \,\,\,\, w/o Stage 1 & 7B & 64.68 \textcolor{gray}{$\pm$ 0.9}\rlap{$^{\ast\ast}$} & 60.97 \textcolor{gray}{$\pm$ 0.6}\rlap{$^{\ast\ast}$} & 85.13 \textcolor{gray}{$\pm$ 1.1}\rlap{$^{\ast\ast}$} & 71.05 \textcolor{gray}{$\pm$ 0.8}\rlap{$^{\ast\ast}$} & 71.65 \textcolor{gray}{$\pm$ 1.7}\rlap{$^{\ast\ast}$} & 69.82 \textcolor{gray}{$\pm$ 2.0}\rlap{$^{\ast\ast}$} \\
GRPO w/o Stage 1& 7B& 72.15 \textcolor{gray}{$\pm$ 0.9}\rlap{$^{\ast\ast}$} & 70.19 \textcolor{gray}{$\pm$ 0.7}\rlap{$^{\ast\ast}$} & 78.74 \textcolor{gray}{$\pm$ 1.3}\rlap{$^{\ast\ast}$} & 74.21 \textcolor{gray}{$\pm$ 0.9}\rlap{$^{\ast\ast}$} & 79.88 \textcolor{gray}{$\pm$ 0.6}\rlap{$^{\ast\ast}$} & 80.46 \textcolor{gray}{$\pm$ 0.9}\rlap{$^{\ast\ast}$} \\
SFT \,\,\,\, w/o Stage 2&7B& 74.16 \textcolor{gray}{$\pm$ 0.4}\rlap{$^{\ast\ast}$} & 71.96 \textcolor{gray}{$\pm$ 0.3}\rlap{$^{\ast\ast}$} & 80.67 \textcolor{gray}{$\pm$ 0.8}\rlap{$^{\ast\ast}$} & 76.07 \textcolor{gray}{$\pm$ 0.4}\rlap{$^{\ast}$} & 76.31 \textcolor{gray}{$\pm$ 0.9}\rlap{$^{\ast\ast}$} & 69.32 \textcolor{gray}{$\pm$ 1.7}\rlap{$^{\ast\ast}$} \\
GRPO w/o Stage 2&7B & 73.28 \textcolor{gray}{$\pm$ 0.6}\rlap{$^{\ast\ast}$} & 70.42 \textcolor{gray}{$\pm$ 0.8}\rlap{$^{\ast\ast}$} & 81.97 \textcolor{gray}{$\pm$ 0.4}\rlap{$^{\ast\ast}$} & 75.75 \textcolor{gray}{$\pm$ 0.4}\rlap{$^{\ast\ast}$} & 81.68 \textcolor{gray}{$\pm$ 0.3}\rlap{$^{\ast\ast}$} & 81.27 \textcolor{gray}{$\pm$ 0.2}\rlap{$^{\ast\ast}$} \\
SFT $\rightarrow$ SFT &7B& \underline{74.97} \textcolor{gray}{$\pm$ 0.8}\rlap{$^{\ast\ast}$} & 73.65 \textcolor{gray}{$\pm$ 0.8}\rlap{$^{\ast\ast}$} & 79.16 \textcolor{gray}{$\pm$ 1.0}\rlap{$^{\ast\ast}$} & 76.30 \textcolor{gray}{$\pm$ 0.8} & 77.15 \textcolor{gray}{$\pm$ 1.7}\rlap{$^{\ast\ast}$} & 72.41 \textcolor{gray}{$\pm$ 2.1}\rlap{$^{\ast\ast}$} \\
SFT $\rightarrow$ GRPO &7B& 73.60 \textcolor{gray}{$\pm$ 1.0}\rlap{$^{\ast\ast}$} & 71.00 \textcolor{gray}{$\pm$ 1.1}\rlap{$^{\ast\ast}$} & 81.58 \textcolor{gray}{$\pm$ 1.0}\rlap{$^{\ast\ast}$} & 75.92 \textcolor{gray}{$\pm$ 0.9}\rlap{$^{\ast\ast}$} & 77.75 \textcolor{gray}{$\pm$ 1.5}\rlap{$^{\ast\ast}$} & 72.69 \textcolor{gray}{$\pm$ 2.0}\rlap{$^{\ast\ast}$} \\
\textbf{Dementia-R1 (Ours)} & 7B & \textbf{77.45} \textcolor{gray}{$\pm$ 0.2} & \textbf{79.26} \textcolor{gray}{$\pm$ 0.4} & 75.46 \textcolor{gray}{$\pm$ 0.7} & \textbf{77.31} \textcolor{gray}{$\pm$ 0.3} & \textbf{84.02} \textcolor{gray}{$\pm$ 0.3} & \textbf{85.73} \textcolor{gray}{$\pm$ 0.4} \\
\bottomrule
\end{tabular}
}
\end{table*}

\begin{table*}[t!]
\centering
\caption{\textbf{Results on HPH.} Notation follows Table~\ref{tab:main_results}.}
\label{tab:haeundae_paik_results}
\scriptsize
\setlength{\tabcolsep}{4pt}
\renewcommand{\arraystretch}{1.0}
\adjustbox{width=0.98\textwidth}{
\begin{tabular}{lccccccc}
\toprule
\textbf{Method} & \textbf{Size} & \textbf{Accuracy ($\uparrow$)} & \textbf{Precision ($\uparrow$)} & \textbf{Recall ($\uparrow$)} & \textbf{F1 score ($\uparrow$)} & \textbf{AUROC ($\uparrow$)} & \textbf{AUPRC ($\uparrow$)} \\
\midrule
\textit{External LLMs} & & & & & & & \\
HuatuoGPT-o1& 8B & 69.68 \textcolor{gray}{$\pm$ 2.2}\rlap{$^{\ast\ast}$} & 35.11 \textcolor{gray}{$\pm$ 3.2}\rlap{$^{\ast\ast}$} & 47.32 \textcolor{gray}{$\pm$ 5.0}\rlap{$^{\ast\ast}$} & 40.23 \textcolor{gray}{$\pm$ 3.5}\rlap{$^{\ast\ast}$} & 63.58 \textcolor{gray}{$\pm$ 3.2}\rlap{$^{\ast\ast}$} & 32.93 \textcolor{gray}{$\pm$ 1.9}\rlap{$^{\ast\ast}$} \\
MedGemma& 27B & 70.21 \textcolor{gray}{$\pm$ 3.0}\rlap{$^{\ast\ast}$} & 38.69 \textcolor{gray}{$\pm$ 4.0}\rlap{$^{\ast\ast}$} & 64.39 \textcolor{gray}{$\pm$ 5.5} & 48.31 \textcolor{gray}{$\pm$ 4.6} & 73.13 \textcolor{gray}{$\pm$ 2.6}\rlap{$^{\ast\ast}$} & 41.06 \textcolor{gray}{$\pm$ 4.4}\rlap{$^{\ast\ast}$} \\
Qwen2.5-7B-Inst& 7B & 71.26 \textcolor{gray}{$\pm$ 1.6}\rlap{$^{\ast\ast}$} & 35.91 \textcolor{gray}{$\pm$ 3.7}\rlap{$^{\ast\ast}$} & 43.41 \textcolor{gray}{$\pm$ 8.5}\rlap{$^{\ast\ast}$} & 39.19 \textcolor{gray}{$\pm$ 5.6}\rlap{$^{\ast\ast}$} & 67.90 \textcolor{gray}{$\pm$ 3.2}\rlap{$^{\ast\ast}$} & 39.28 \textcolor{gray}{$\pm$ 4.1}\rlap{$^{\ast\ast}$} \\
Qwen2.5-32B-Inst& 32B & 74.00 \textcolor{gray}{$\pm$ 1.3} & 41.10 \textcolor{gray}{$\pm$ 2.3} & 46.83 \textcolor{gray}{$\pm$ 1.8}\rlap{$^{\ast\ast}$} & 43.76 \textcolor{gray}{$\pm$ 1.9}\rlap{$^{\ast\ast}$} & 68.54 \textcolor{gray}{$\pm$ 1.9}\rlap{$^{\ast\ast}$} & 39.20 \textcolor{gray}{$\pm$ 1.9}\rlap{$^{\ast\ast}$} \\
Qwen2.5-72B-Inst& 72B & \underline{74.63} \textcolor{gray}{$\pm$ 1.4} & 42.57 \textcolor{gray}{$\pm$ 2.6} & 50.24 \textcolor{gray}{$\pm$ 3.6}\rlap{$^{\ast\ast}$} & 46.07 \textcolor{gray}{$\pm$ 2.9}\rlap{$^{\ast\ast}$} & 72.37 \textcolor{gray}{$\pm$ 1.1}\rlap{$^{\ast\ast}$} & \underline{44.78} \textcolor{gray}{$\pm$ 2.6}\rlap{$^{\ast}$} \\
Llama-3.3-70B-Inst& 70B & 57.58 \textcolor{gray}{$\pm$ 2.0}\rlap{$^{\ast\ast}$} & 30.28 \textcolor{gray}{$\pm$ 1.6}\rlap{$^{\ast\ast}$} & \textbf{74.15} \textcolor{gray}{$\pm$ 5.2}\rlap{$^{\ast\ast}$} & 42.98 \textcolor{gray}{$\pm$ 2.3}\rlap{$^{\ast\ast}$} & 67.41 \textcolor{gray}{$\pm$ 1.4}\rlap{$^{\ast\ast}$} & 34.91 \textcolor{gray}{$\pm$ 0.7}\rlap{$^{\ast\ast}$} \\
\midrule
\textit{Specialized Models} & & & & & & & \\
SFT \,\,\,\, w/o Stage 1 & 7B & 65.79 \textcolor{gray}{$\pm$ 1.1}\rlap{$^{\ast\ast}$} & 32.96 \textcolor{gray}{$\pm$ 1.4}\rlap{$^{\ast\ast}$} & 56.59 \textcolor{gray}{$\pm$ 2.8} & 41.65 \textcolor{gray}{$\pm$ 1.7}\rlap{$^{\ast\ast}$} & 63.75 \textcolor{gray}{$\pm$ 1.8}\rlap{$^{\ast\ast}$} & 30.39 \textcolor{gray}{$\pm$ 3.2}\rlap{$^{\ast\ast}$} \\
GRPO w/o Stage 1& 7B& 72.95 \textcolor{gray}{$\pm$ 1.8} & 38.37 \textcolor{gray}{$\pm$ 3.9}\rlap{$^{\ast\ast}$} & 41.95 \textcolor{gray}{$\pm$ 6.2}\rlap{$^{\ast\ast}$} & 39.98 \textcolor{gray}{$\pm$ 4.5}\rlap{$^{\ast\ast}$} & 69.96 \textcolor{gray}{$\pm$ 3.4}\rlap{$^{\ast\ast}$} & 38.69 \textcolor{gray}{$\pm$ 2.4}\rlap{$^{\ast\ast}$} \\
SFT \,\,\,\, w/o Stage 2&7B& 41.68 \textcolor{gray}{$\pm$ 2.7}\rlap{$^{\ast\ast}$} & 21.93 \textcolor{gray}{$\pm$ 2.5}\rlap{$^{\ast\ast}$} & \underline{66.83} \textcolor{gray}{$\pm$ 9.6} & 33.01 \textcolor{gray}{$\pm$ 4.0}\rlap{$^{\ast\ast}$} & 52.95 \textcolor{gray}{$\pm$ 2.7}\rlap{$^{\ast\ast}$} & 25.75 \textcolor{gray}{$\pm$ 3.6}\rlap{$^{\ast\ast}$} \\
GRPO w/o Stage 2&7B & 74.32 \textcolor{gray}{$\pm$ 1.7} & 41.35 \textcolor{gray}{$\pm$ 3.2} & 44.88 \textcolor{gray}{$\pm$ 3.6}\rlap{$^{\ast\ast}$} & 42.99 \textcolor{gray}{$\pm$ 3.0}\rlap{$^{\ast\ast}$} & 69.73 \textcolor{gray}{$\pm$ 0.9}\rlap{$^{\ast\ast}$} & 38.53 \textcolor{gray}{$\pm$ 2.6}\rlap{$^{\ast\ast}$} \\
SFT $\rightarrow$ SFT &7B& 38.95 \textcolor{gray}{$\pm$ 3.8}\rlap{$^{\ast\ast}$} & 20.80 \textcolor{gray}{$\pm$ 1.5}\rlap{$^{\ast\ast}$} & 64.88 \textcolor{gray}{$\pm$ 3.3} & 31.49 \textcolor{gray}{$\pm$ 2.0}\rlap{$^{\ast\ast}$} & 49.14 \textcolor{gray}{$\pm$ 1.2}\rlap{$^{\ast\ast}$} & 22.47 \textcolor{gray}{$\pm$ 1.7}\rlap{$^{\ast\ast}$} \\
SFT $\rightarrow$ GRPO &7B& \underline{74.63} \textcolor{gray}{$\pm$ 1.2} & \underline{44.11} \textcolor{gray}{$\pm$ 1.8} & 65.37 \textcolor{gray}{$\pm$ 2.4} & \underline{52.66} \textcolor{gray}{$\pm$ 1.9} & \underline{74.75} \textcolor{gray}{$\pm$ 2.6} & 39.25 \textcolor{gray}{$\pm$ 3.3}\rlap{$^{\ast\ast}$} \\
\textbf{Dementia-R1 (Ours)} & 7B & \textbf{75.79} \textcolor{gray}{$\pm$ 1.6} & \textbf{45.67} \textcolor{gray}{$\pm$ 2.6} & 63.41 \textcolor{gray}{$\pm$ 4.1} & \textbf{53.06} \textcolor{gray}{$\pm$ 2.8} & \textbf{78.37} \textcolor{gray}{$\pm$ 1.8} & \textbf{51.03} \textcolor{gray}{$\pm$ 2.7} \\
\bottomrule
\end{tabular}
}
\end{table*}

\begin{table*}[t!]
\centering
\caption{\textbf{Results on ADNI Benchmark.} Includes Random Forest as ML baseline. Notation follows Table~\ref{tab:main_results}.}
\label{tab:adni_results}
\scriptsize
\setlength{\tabcolsep}{4pt}
\renewcommand{\arraystretch}{1.0}
\adjustbox{width=0.98\textwidth}{
\begin{tabular}{lccccccc}
\toprule
\textbf{Model Method} &\textbf{Size} & \textbf{Accuracy ($\uparrow$)} & \textbf{Precision ($\uparrow$)} & \textbf{Recall ($\uparrow$)} & \textbf{F1 score ($\uparrow$)} & \textbf{AUROC ($\uparrow$)} & \textbf{AUPRC ($\uparrow$)} \\
\midrule
\textit{ML Baseline} & & & & & & & \\
Random Forest &---&
83.46 \textcolor{gray}{$\pm$ 0.6} &
83.57 \textcolor{gray}{$\pm$ 0.7} &
77.13 \textcolor{gray}{$\pm$ 1.1} &
80.22 \textcolor{gray}{$\pm$ 0.7} & 90.96 \textcolor{gray}{$\pm$ 0.2} & 89.68 \textcolor{gray}{$\pm$ 0.3} \\

\midrule
\textit{External LLMs} & & & & & & & \\
HuatuoGPT-o1 &8B & 62.51 \textcolor{gray}{$\pm$ 2.1}\rlap{$^{\ast\ast}$} & 55.30 \textcolor{gray}{$\pm$ 1.8}\rlap{$^{\ast\ast}$} & 74.59 \textcolor{gray}{$\pm$ 2.2}\rlap{$^{\ast\ast}$} & 63.51 \textcolor{gray}{$\pm$ 1.9}\rlap{$^{\ast\ast}$} & 71.69 \textcolor{gray}{$\pm$ 1.8}\rlap{$^{\ast\ast}$} & 67.34 \textcolor{gray}{$\pm$ 2.9}\rlap{$^{\ast\ast}$} \\
MedGemma& 27B & 73.47 \textcolor{gray}{$\pm$ 1.0}\rlap{$^{\ast\ast}$} & 68.54 \textcolor{gray}{$\pm$ 1.2}\rlap{$^{\ast\ast}$} & 72.70 \textcolor{gray}{$\pm$ 0.8}\rlap{$^{\ast\ast}$} & 70.56 \textcolor{gray}{$\pm$ 1.0}\rlap{$^{\ast\ast}$} & 78.48 \textcolor{gray}{$\pm$ 1.0}\rlap{$^{\ast\ast}$} & 71.71 \textcolor{gray}{$\pm$ 1.1}\rlap{$^{\ast\ast}$} \\
Qwen2.5-7B-Inst & 7B & 62.48 \textcolor{gray}{$\pm$ 1.3}\rlap{$^{\ast\ast}$} & 55.41 \textcolor{gray}{$\pm$ 1.1}\rlap{$^{\ast\ast}$} & 72.62 \textcolor{gray}{$\pm$ 1.4}\rlap{$^{\ast\ast}$} & 62.85 \textcolor{gray}{$\pm$ 1.2}\rlap{$^{\ast\ast}$} & 66.23 \textcolor{gray}{$\pm$ 1.7}\rlap{$^{\ast\ast}$} & 57.17 \textcolor{gray}{$\pm$ 2.1}\rlap{$^{\ast\ast}$} \\
Qwen2.5-32B-Inst &32B & 76.47 \textcolor{gray}{$\pm$ 0.7} & 72.16 \textcolor{gray}{$\pm$ 0.8} & 75.19 \textcolor{gray}{$\pm$ 1.3}\rlap{$^{\ast\ast}$} & 73.64 \textcolor{gray}{$\pm$ 0.9} & \underline{82.51} \textcolor{gray}{$\pm$ 0.5} & \underline{83.50} \textcolor{gray}{$\pm$ 0.5} \\
Qwen2.5-72B-Inst& 72B & \textbf{80.00} \textcolor{gray}{$\pm$ 0.7}\rlap{$^{\ast\ast}$} & \textbf{82.80} \textcolor{gray}{$\pm$ 1.4}\rlap{$^{\ast\ast}$} & 68.50 \textcolor{gray}{$\pm$ 0.6}\rlap{$^{\ast\ast}$} & \underline{74.97} \textcolor{gray}{$\pm$ 0.8} & 81.45 \textcolor{gray}{$\pm$ 0.5}\rlap{$^{\ast}$} & \textbf{83.56} \textcolor{gray}{$\pm$ 0.6} \\
Llama-3.3-70B-Inst& 70B & \underline{78.91} \textcolor{gray}{$\pm$ 1.2}\rlap{$^{\ast}$} & \underline{77.27} \textcolor{gray}{$\pm$ 1.9}\rlap{$^{\ast\ast}$} & 73.39 \textcolor{gray}{$\pm$ 1.1}\rlap{$^{\ast\ast}$} & \textbf{75.27} \textcolor{gray}{$\pm$ 1.3} & 81.17 \textcolor{gray}{$\pm$ 0.4}\rlap{$^{\ast}$} & 83.48 \textcolor{gray}{$\pm$ 0.5} \\
\midrule
\textit{Specialized Models} & & & & & & & \\
SFT \,\,\,\, w/o Stage 1 & 7B & 69.49 \textcolor{gray}{$\pm$ 2.0}\rlap{$^{\ast\ast}$} & 62.39 \textcolor{gray}{$\pm$ 2.0}\rlap{$^{\ast\ast}$} & 76.22 \textcolor{gray}{$\pm$ 1.7} & 68.61 \textcolor{gray}{$\pm$ 1.9}\rlap{$^{\ast\ast}$} & 73.40 \textcolor{gray}{$\pm$ 1.7}\rlap{$^{\ast\ast}$} & 63.03 \textcolor{gray}{$\pm$ 2.1}\rlap{$^{\ast\ast}$} \\
GRPO w/o Stage 1 & 7B & 66.34 \textcolor{gray}{$\pm$ 1.3}\rlap{$^{\ast\ast}$} & 59.34 \textcolor{gray}{$\pm$ 1.4}\rlap{$^{\ast\ast}$} & 73.22 \textcolor{gray}{$\pm$ 0.7}\rlap{$^{\ast\ast}$} & 65.55 \textcolor{gray}{$\pm$ 1.1}\rlap{$^{\ast\ast}$} & 72.97 \textcolor{gray}{$\pm$ 0.6}\rlap{$^{\ast\ast}$} & 67.18 \textcolor{gray}{$\pm$ 1.3}\rlap{$^{\ast\ast}$} \\
SFT \,\,\,\, w/o Stage 2 & 7B & 76.59 \textcolor{gray}{$\pm$ 0.9} & 71.89 \textcolor{gray}{$\pm$ 1.3} & \underline{76.31} \textcolor{gray}{$\pm$ 0.6}\rlap{$^{\ast}$} & 74.03 \textcolor{gray}{$\pm$ 0.8} & 81.74 \textcolor{gray}{$\pm$ 1.9} & 76.12 \textcolor{gray}{$\pm$ 2.4}\rlap{$^{\ast\ast}$} \\
GRPO w/o Stage 2 & 7B  & 76.44 \textcolor{gray}{$\pm$ 1.2} & 74.05 \textcolor{gray}{$\pm$ 1.6}\rlap{$^{\ast}$} & 70.99 \textcolor{gray}{$\pm$ 1.6}\rlap{$^{\ast\ast}$} & 72.48 \textcolor{gray}{$\pm$ 1.4}\rlap{$^{\ast}$} & 79.07 \textcolor{gray}{$\pm$ 0.8}\rlap{$^{\ast\ast}$} & 77.65 \textcolor{gray}{$\pm$ 1.3}\rlap{$^{\ast\ast}$} \\
SFT $\rightarrow$ SFT &7B & 75.57 \textcolor{gray}{$\pm$ 0.5} & 74.27 \textcolor{gray}{$\pm$ 1.1}\rlap{$^{\ast}$} & 67.55 \textcolor{gray}{$\pm$ 1.1}\rlap{$^{\ast\ast}$} & 70.74 \textcolor{gray}{$\pm$ 0.6}\rlap{$^{\ast\ast}$} & 79.90 \textcolor{gray}{$\pm$ 0.3}\rlap{$^{\ast\ast}$} & 73.77 \textcolor{gray}{$\pm$ 1.1}\rlap{$^{\ast\ast}$} \\
SFT $\rightarrow$ GRPO &7B & 77.34 \textcolor{gray}{$\pm$ 0.6} & 73.56 \textcolor{gray}{$\pm$ 0.9} & 75.19 \textcolor{gray}{$\pm$ 0.5}\rlap{$^{\ast\ast}$} & 74.37 \textcolor{gray}{$\pm$ 0.5} & 77.88 \textcolor{gray}{$\pm$ 1.5}\rlap{$^{\ast\ast}$} & 69.09 \textcolor{gray}{$\pm$ 2.3}\rlap{$^{\ast\ast}$} \\
\textbf{Dementia-R1}&7B  & 76.92 \textcolor{gray}{$\pm$ 0.8} & 71.48 \textcolor{gray}{$\pm$ 1.3} & \textbf{78.63} \textcolor{gray}{$\pm$ 0.7} & 74.87 \textcolor{gray}{$\pm$ 0.6} & \textbf{83.17} \textcolor{gray}{$\pm$ 1.3} & 82.00 \textcolor{gray}{$\pm$ 1.1} \\
\bottomrule
\end{tabular}
}
\end{table*}

\begin{figure*}[!t]
    \centering
    \includegraphics[width=\textwidth]{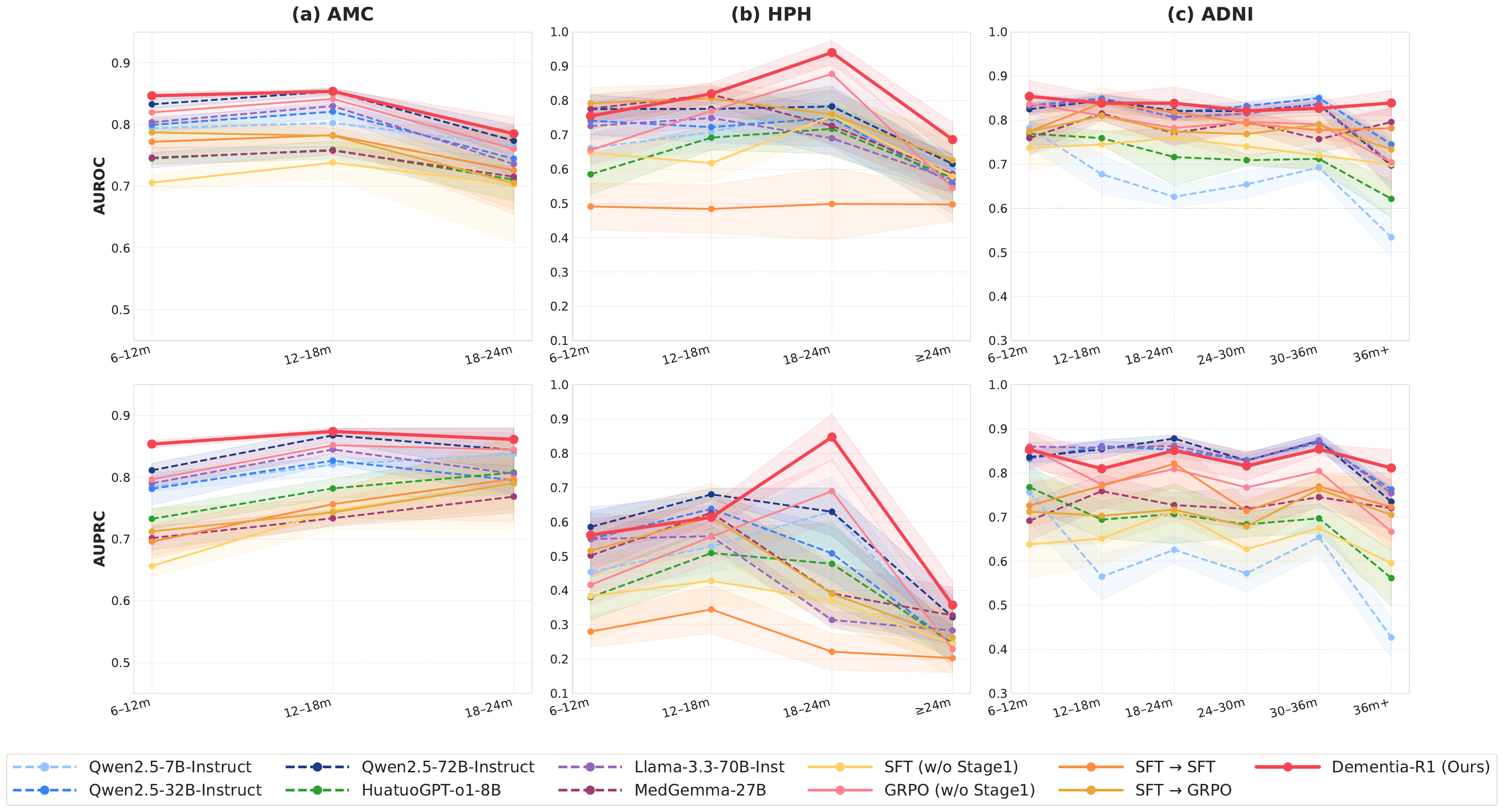}
    \vspace{-0.8cm}
    \caption{\textbf{Performance across time gaps.} AUROC (top) and AUPRC (bottom) by prediction horizon for (a) AMC, (b) HPH, and (c) ADNI. Dementia-R1 achieves the highest AUROC at the longest prediction horizons across all three cohorts.}
    \label{fig:timegap}
    \vspace{-0.5cm}
\end{figure*}
\subsection{Real-World Unstructured Data Results}
\label{subsec:main_results}

\paragraph{Dementia prognosis.}
On the AMC cohort (Table~\ref{tab:main_results}), Dementia-R1 achieves the best overall performance, reaching 84.02\% AUROC and 77.31\% F1. It outperforms every specialized 7B baseline, including direct GRPO without Stage~1 and the hybrid SFT$\rightarrow$GRPO pipeline, with a significant AUROC gain in each case ($p<0.01$). It also surpasses general-purpose LLMs up to 10$\times$ larger, significantly on every metric for models up to 70B ($p<0.01$). Against Qwen2.5-72B-Instruct, the gains in Accuracy, Precision, Recall, and AUPRC are significant ($p<0.01$), while the F1 (77.31 vs.\ 77.30) and AUROC (84.02 vs.\ 83.41) differences are within noise. Stage~1 clinical-index pretraining therefore provides a substantial advantage on real-world unstructured notes.

\paragraph{Cross-disease generalization (PDD prediction).}
On the HPH cohort for Parkinson's disease dementia (PDD) prediction (Table~\ref{tab:haeundae_paik_results}), Dementia-R1 again achieves the best AUROC (78.37\%) and AUPRC (51.03\%). It outperforms Qwen2.5-72B-Instruct (72.37\% AUROC, $p<0.01$) and the strongest 7B baseline, SFT$\rightarrow$GRPO (74.75\% AUROC). Against SFT$\rightarrow$GRPO only the AUPRC gain is significant ($+11.78$, $p<0.01$), so the advantage here lies mainly in ranking the positive class under low prevalence. The framework thus transfers beyond AMC to a distinct cross-disease, cross-hospital setting.

\subsection{Generalization to Structured Benchmarks}
\label{subsec:adni_results}

On the structured ADNI benchmark (Table~\ref{tab:adni_results}), Dementia-R1 achieves the highest AUROC among all LLM baselines at 83.17\%, while maintaining an F1 score (74.87\%) comparable to substantially larger models. In addition, its Stage~1 performance on fine-grained intermediate targets such as CDRSB and ADAS remains competitive with strong larger-scale baselines (Figure~\ref{fig:radar_chart}; Appendix Table~\ref{tab:adni_task_performance}). These results suggest that the proposed framework generalizes beyond unstructured hospital notes and remains effective on structured longitudinal benchmark data.

As expected, the Random Forest baseline dominates on ADNI, which is well suited to classical tabular models; our framework instead targets unstructured longitudinal narratives and remains competitive even in this structured regime. This ordering is not ADNI-specific: ClinicalBench~\cite{chen2026clinicalbench} evaluates 22 general and medical LLMs from 0.5B to 70B under prompting, prompt engineering, and fine-tuning, and finds that none surpasses traditional ML on tabular clinical prediction. Nor is the gap attributable to our linearization: substituting the best-performing serialization format from each of two established references leaves AUROC unchanged or lower (Appendix~\ref{app:adni_preprocessing}, Table~\ref{tab:linearization}).
   

\begin{figure}[!t]
    \centering
    \includegraphics[width=0.7\columnwidth]{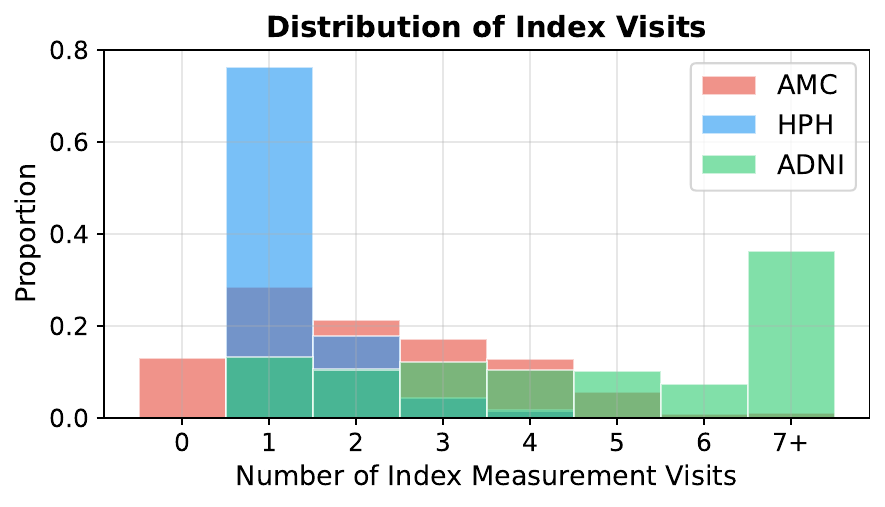}
   
    \vspace{-0.25cm}
    \caption{\textbf{Clinical Index Density.} ADNI (${\sim}$4.3$\times$) and AMC (${\sim}$1.6$\times$) have more index visits per patient than HPH.}

    \label{fig:index_density}
    \vspace{-0.7cm}
\end{figure}

\subsection{Temporal Robustness Analysis}
\label{subsec:temporal_robustness}
Figure~\ref{fig:timegap} shows Dementia-R1 remains competitive across prediction horizons, achieving the best AUROC at the longest horizons on AMC (78.51\% at 18--24m) and ADNI (83.90\% at 36m+). On ADNI, the margin over Qwen2.5-32B/72B widens significantly in the longest-term setting, highlighting Stage~1's benefit for extended reasoning. In HPH, Dementia-R1 performs best overall, though performance drops more noticeably in the $\geq$24m bucket. This aligns with HPH's lower density of clinical indices relative to AMC and ADNI (Figure~\ref{fig:index_density}), suggesting that sparse measurements complicate long-range prediction.
\subsection{Neurologist Evaluation}
\label{sec:human_eval}
\label{subsec:human_eval}

\begin{figure}[!t]
    \centering
    \includegraphics[width=0.85\columnwidth]{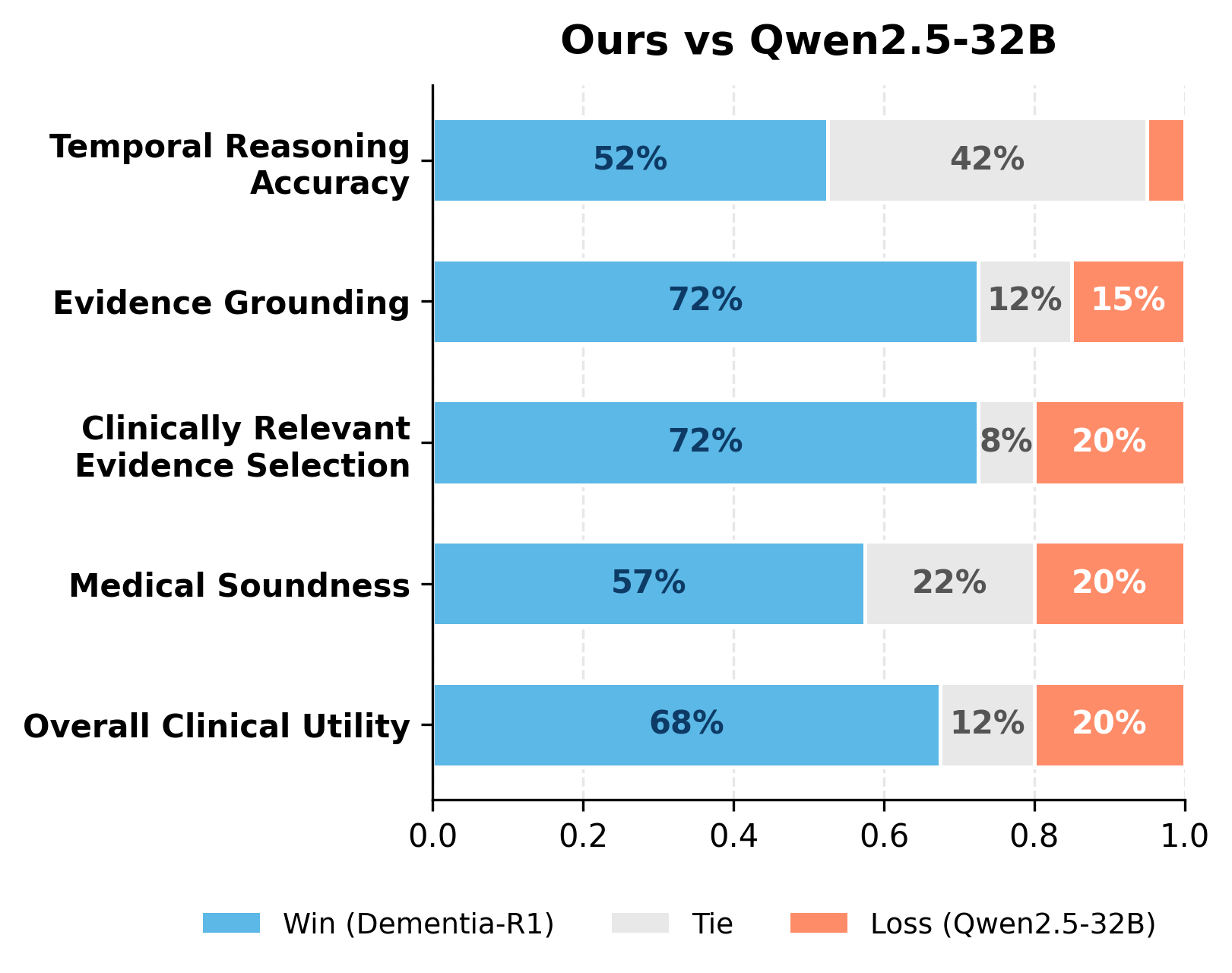}
    \vspace{-0.3cm}
    \caption{\textbf{Neurologist Blind Pairwise Evaluation}. Results from two cohorts (AMC and HPH) comparing Dementia-R1 against Qwen2.5-32B-Instruct across five clinical dimensions.}
    \label{fig:Human_eval_result}
    \vspace{-0.5cm}
\end{figure}

Two board-certified neurologists conducted a blinded pairwise evaluation against Qwen2.5-32B-Instruct across five clinical dimensions and two independent cohorts (AMC and HPH), yielding 100 judgment pairs. Inter-rater agreement was moderate (Gwet's AC1 = 0.54). Dementia-R1 achieved a 68\% win rate in Overall Clinical Utility, 72\% in both Evidence Grounding and Clinically Relevant Evidence Selection, and a 95\% Win/Tie rate in Temporal Reasoning Accuracy (Figure~\ref{fig:Human_eval_result}). We also conducted the same evaluation against Qwen2.5-72B-Instruct; however, inter-rater agreement was insufficient, so we report only the 32B comparison.

\subsection{Ablation Study}
\label{subsec:ablation}

\begin{figure}[!t]
    \centering
    \includegraphics[width=0.80\columnwidth]{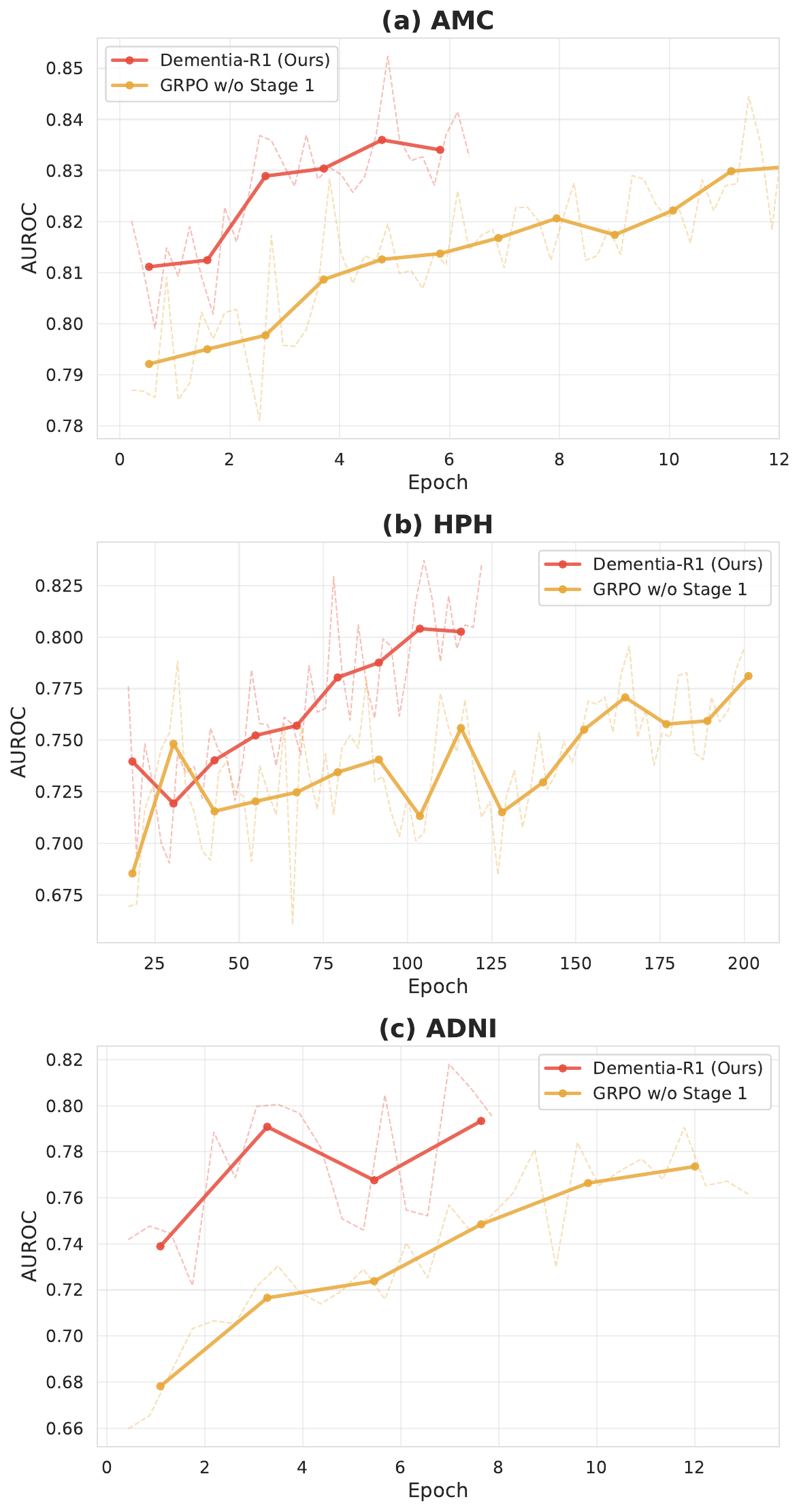}
    \vspace{-0.3cm}
    \caption{\textbf{Stage~1 Pre-training Impact.} AUROC over training epochs on (a)~AMC, (b)~HPH, and (c)~ADNI. Solid lines show 500-step binned averages; dashed lines show raw checkpoint values.}
    \label{fig:ablation_curves}
    \vspace{-0.3cm}
\end{figure}

\begin{figure}[!t]
    \centering
    \includegraphics[width=\columnwidth]{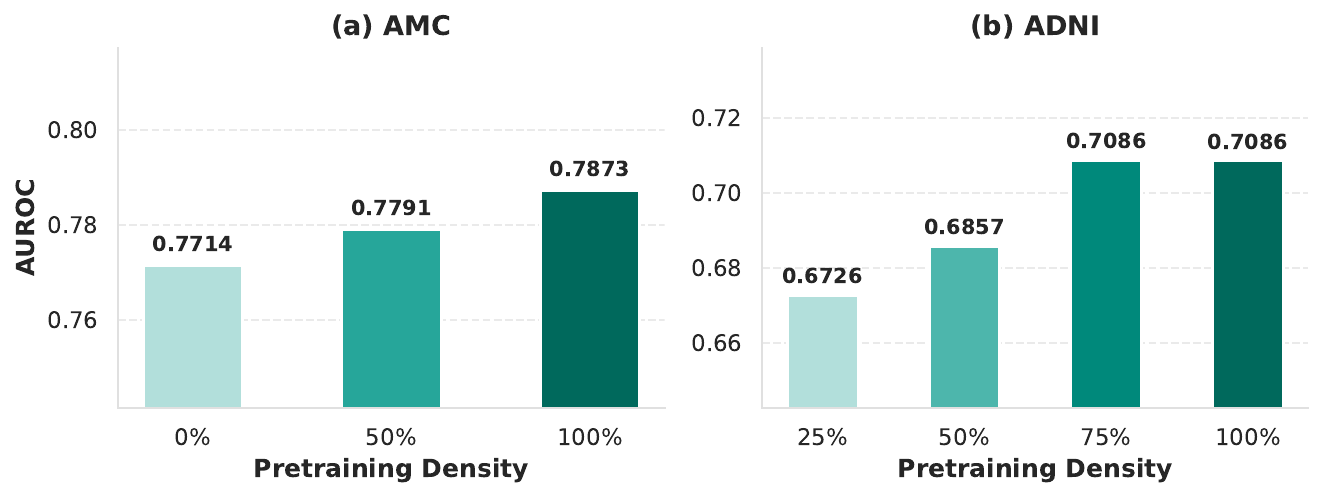}
    \vspace{-0.5cm}
    \caption{\textbf{Pretraining Density Ablation.} Best overall AUROC at varying Stage~1 index density levels on (a)~AMC and (b)~ADNI. Performance degrades monotonically with reduced density.}
    \label{fig:pretraining_density_combined}
    \vspace{-0.3cm}
\end{figure}

\paragraph{Effect of Stage~1 Pre-training.}
Figure~\ref{fig:ablation_curves} shows that incorporating Stage~1 consistently improves convergence stability and peak AUROC across AMC, HPH, and ADNI. However, the gain on HPH is smaller than that on AMC. Since clinical-index annotations are substantially sparser in HPH than in AMC (Figure~\ref{fig:index_density}), we hypothesize that this performance gap stems from the limited availability of intermediate clinical-index supervision. To test this hypothesis, we conduct an ablation study on the AMC cohort by progressively reducing the density of clinical-index labels. Figure~\ref{fig:pretraining_density_combined} shows that decreasing supervision density monotonically degrades downstream prognosis performance on AMC. Taken together, these results indicate that richer intermediate supervision leads to stronger final longitudinal prediction.

\paragraph{Choice of Stage~1 Target.}
Using a fixed 1{,}000-sample exploratory ablation, GDS-only pretraining slightly outperforms the combined-target setting on AMC and HPH (Appendix Table~\ref{tab:target_ablation}). We do not interpret this as a universal ranking of index importance. Rather, the usefulness of a pretraining target appears to depend on scale granularity, annotation density, and cohort-specific availability. A coarser target such as GDS may offer a cleaner optimization signal during Stage~1, whereas finer or sparser targets can be harder to exploit consistently. For example, CDR is substantially sparser than MMSE and GDS in AMC (Appendix Figure~\ref{fig:index_distribution}), which may reduce its utility during Stage~1 despite its clinical relevance at inference. The most effective pretraining target is therefore not always the one that contributes most at inference, as the masking analysis below shows. Together with those results, this suggests that different clinical indices play distinct roles in pretraining and in downstream prognosis.


\begin{figure}[!t]
    \centering
    \includegraphics[width=0.68\columnwidth]{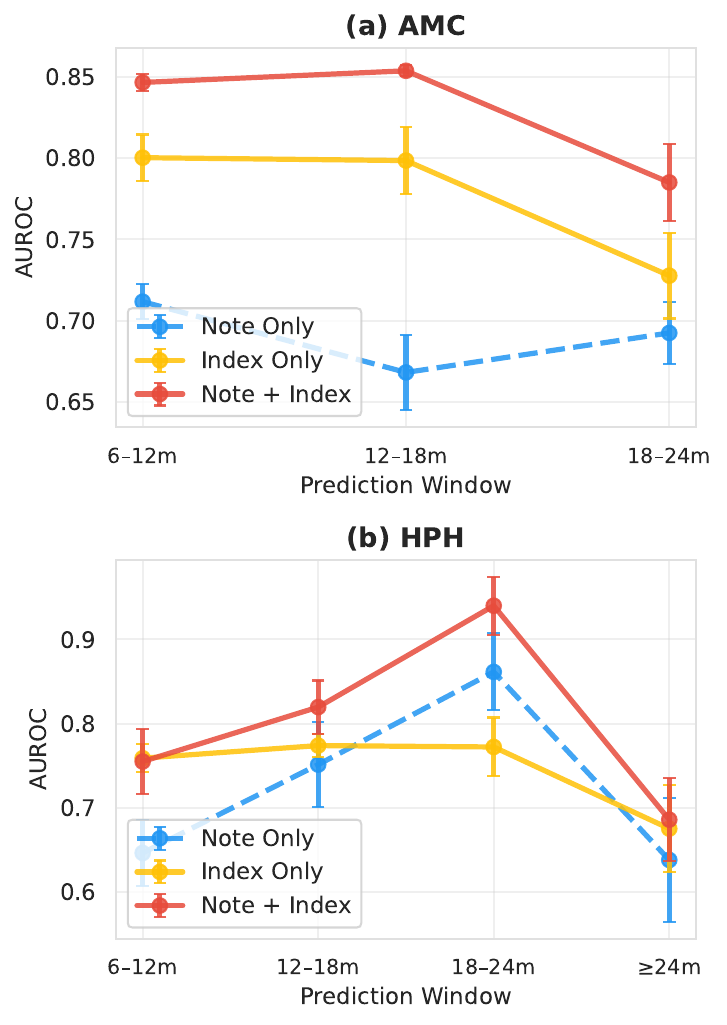}
    \vspace{-0.3cm}
    \caption{\textbf{Note vs.\ Index Ablation.} AUROC by prediction window on (a)~AMC and (b)~HPH. Three input settings are compared: \textbf{Note Only} (raw clinical text), \textbf{Index Only} (structured MMSE/GDS/CDR), and \textbf{Note + Index} (combined).}
    \label{fig:note_vs_index}
    \vspace{-0.3cm}
\end{figure}

\begin{figure}[!t]
    \centering
    \includegraphics[width=\columnwidth]{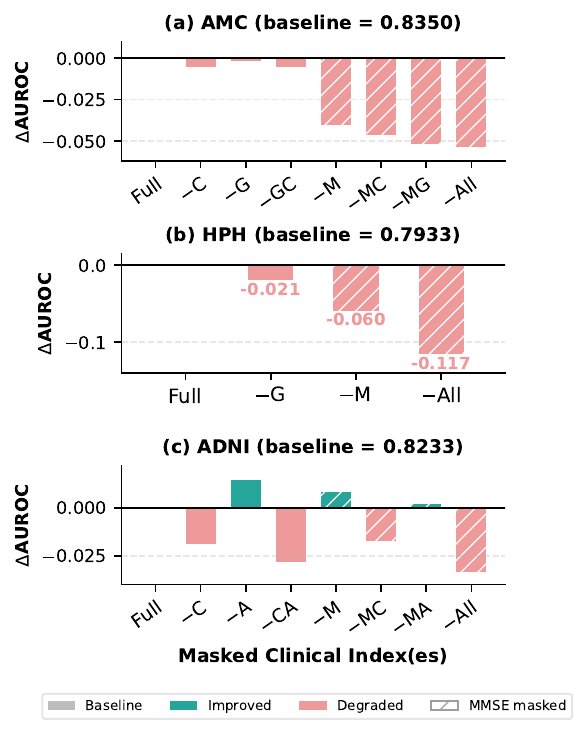}
    
    
    \vspace{-0.0cm} 
    \caption{\textbf{Clinical Index Masking Ablation.} AUROC change ($\Delta$) when masking clinical indices at inference. Conditions are named by the index removed: \textbf{M}\,=\,MMSE, \textbf{G}\,=\,GDS, \textbf{C}\,=\,CDR in (a)~AMC and (b)~HPH, and \textbf{M}\,=\,MMSE, \textbf{C}\,=\,CDRSB, \textbf{A}\,=\,ADAS-Cog/memory in (c)~ADNI. \textit{Full} retains all indices and $-$\textit{All} masks every index. Hatching marks conditions in which MMSE is masked. Exact values are in Appendix~\ref{app:masking_ablation} (AMC) and~\ref{app:loo_ablation} (ADNI).}
    \label{fig:masking_ablation}
    
    \vspace{-0.0cm} 
\end{figure}

\paragraph{Role of Clinical Indices at Inference.}
We compare three input settings: \textbf{Note Only} (raw clinical text), \textbf{Index Only} (structured MMSE/GDS/CDR), and \textbf{Note + Index} (combined). Note + Index performs best on both AMC and HPH (Figure~\ref{fig:note_vs_index}), showing that structured indices and free-text notes provide complementary prognostic information. The benefit of index input tracks measurement density: AMC has far more index-bearing visits per patient than HPH (Figure~\ref{fig:index_density}), so explicit indices give a stronger signal there. We also mask one or more indices at inference, writing $-$\textbf{M} for MMSE, $-$\textbf{G} for GDS, and $-$\textbf{C} for CDR (CDRSB on ADNI). MMSE-family signals contribute most on AMC and HPH, while CDRSB has the largest effect on ADNI (Figure~\ref{fig:masking_ablation}). Performance nevertheless remains substantial even when all indices are masked, indicating that the model has internalized longitudinal reasoning beyond explicit index copying. Appendix~\ref{app:sparse_index} confirms this on the naturally occurring zero-index subsets. Non-monotonic trajectories are a further stress case. On the 209 AMC patients whose GDS trajectory crosses the dementia cutoff, Dementia-R1 leads all nine baselines, and its margin over the direct Stage~1 ablation widens to $7.35$ AUROC points (Appendix~\ref{app:conflicting_notes}).


\subsubsection{Generalization Across Backbones}
Repeating all configurations on Llama-3.1-8B-Instruct yields the same conclusion: Dementia-R1 achieves the best F1, AUROC, and AUPRC among all Llama-based variants (Appendix Table~\ref{tab:asan_llama_results}), supporting that the benefit of the framework is not backbone-specific.

\section{Conclusion}
We introduced Dementia-R1, a reinforcement learning framework for longitudinal dementia prognosis from clinical histories. A Cold-Start RL stage on verifiable clinical indices teaches longitudinal reasoning under sparse rewards. Across two real-world hospital cohorts and the structured ADNI benchmark, Dementia-R1 consistently achieved strong performance, including cross-hospital generalization and the best AUROC among LLM baselines on ADNI. These results support verifiable-reward RL as an effective approach for longitudinal clinical prediction from real-world patient records.

\clearpage
\section*{Limitations}
We acknowledge several limitations in our study.

\begin{itemize}
    \item \textbf{Cohort and language generalization.} All unstructured data come from two Korean hospitals, so a single nation and ethnic group is represented, and generalization to other demographics, languages, or documentation styles is untested. The framework itself is not language-specific, since it depends only on the availability of verifiable clinical indices. Once a comparable extractor exists for the target language, the same pipeline transfers directly. Validation on English-language EHRs from US or UK hospital systems is the immediate next step.

    \item \textbf{Translation.} Converting Korean notes into English may lose subtle cues, such as syntactic irregularities, that are themselves informative for cognitive decline. Translation is a practical convenience rather than a methodological requirement. Future work should apply the method directly to native-language text.

    \item \textbf{Dependence on the auxiliary extractor.} Stage~1 rewards come from an auxiliary LLM, so extraction errors could in principle propagate into RL training. A human expert audit over 600 notes measured 98.17\% slot-level accuracy, with case-level rather than systematic errors (Appendix~\ref{app:extraction_audit}). A noise-injection test shows that the downstream prognosis is insensitive to this failure mode: corrupting 70\% of the Stage~1 indices changes the final AUROC by less than half a point (Appendix~\ref{app:noise_robustness}). The audit used a single clinician, so inter-rater agreement remains unquantified.

    \item \textbf{Reliance on quantifiable clinical indices.} Our approach requires indices that can be verified numerically. In practice this is rarely a barrier, since most chronic and progressive diseases are monitored through quantifiable assessments such as UPDRS, PHQ-9, NYHA class, or HbA1c, and our Cold-Start RL is scale-agnostic. Where no established scale fits, the same design becomes a discovery tool, because comparing candidate indices as Stage~1 targets reveals which signals carry the most prognostic information (Sec.~\ref{subsec:ablation}). Conditions defined mainly by subjective markers remain out of reach.

    \item \textbf{Nature of the reference labels.} The dementia and PDD labels are clinically derived from staging instruments, ICD-10 coding, and prescription records (Sec.~\ref{sec:data_sources}). They are not biomarker-confirmed or neuropathologically verified. Some diagnostic heterogeneity is therefore possible, and absolute performance should be read against clinical reference standards.

    \item \textbf{Residual prodromal signal.} Explicit leakage is prevented by construction, since the input ends at the Target Anchor and all splits are patient-level (Sec.~\ref{sec:sample_construction}). Some residual signal nevertheless persists, as in any longitudinal clinical task. We believe \textit{prodromal signal} is the more precise framing. Gradual cognitive decline, mood changes, or subtle functional impairment in earlier notes are clinically meaningful precursors that a model should use. Removing them would require a single-visit input, which defeats the task. We therefore frame our task as \textit{prognosis} rather than screening.

    \item \textbf{Binary formulation.} We represent the outcome as dementia versus non-dementia. This is a reasonable first design, but it collapses clinically meaningful distinctions. Two extensions follow naturally. Mild cognitive impairment could be added as a third class to capture the prodromal stage. Subtype classification within dementia is also feasible, given the cross-disease generalization to PDD shown here.
\end{itemize}

\section*{Ethics Statement}
This retrospective study was approved by the Institutional Review Board (IRB No. 2023-1628), which waived the requirement for informed consent due to the use of de-identified medical records. All methods were performed in accordance with the relevant guidelines and regulations of the AMC Ethics Committee and the Declaration of Helsinki. The HPH study was approved by its Institutional Review Board (IRB No. 2023-11-042), and all procedures were conducted in accordance with the Declaration of Helsinki and relevant institutional guidelines.

\section*{Acknowledgments}
Data collection and sharing for this project was funded by the Alzheimer's Disease Neuroimaging Initiative (ADNI) (National Institutes of Health Grant U01 AG024904) and DOD ADNI (Department of Defense award number W81XWH-12-2-0012). ADNI is funded by the National Institute on Aging, the National Institute of Biomedical Imaging and Bioengineering, and through generous contributions from the following: AbbVie, Alzheimer's Association; Alzheimer's Drug Discovery Foundation; Araclon Biotech; BioClinica, Inc.; Biogen; Bristol-Myers Squibb Company; CereSpir, Inc.; Cogstate; Eisai Inc.; Elan Pharmaceuticals, Inc.; Eli Lilly and Company; EuroImmun; F. Hoffmann-La Roche Ltd and its affiliated company Genentech, Inc.; Fujirebio; GE Healthcare; IXICO Ltd.; Janssen Alzheimer Immunotherapy Research \& Development, LLC.; Johnson \& Johnson Pharmaceutical Research \& Development LLC.; Lumosity; Lundbeck; Merck \& Co., Inc.; Meso Scale Diagnostics, LLC.; NeuroRx Research; Neurotrack Technologies; Novartis Pharmaceuticals Corporation; Pfizer Inc.; Piramal Imaging; Servier; Takeda Pharmaceutical Company; and Transition Therapeutics. The Canadian Institutes of Health Research is providing funds to support ADNI clinical sites in Canada. Private sector contributions are facilitated by the Foundation for the National Institutes of Health (\url{www.fnih.org}). The grantee organization is the Northern California Institute for Research and Education, and the study is coordinated by the Alzheimer's Therapeutic Research Institute at the University of Southern California. ADNI data are disseminated by the Laboratory for Neuro Imaging at the University of Southern California.

This work was supported by the National Research Foundation of Korea under Grant RS-2024-00336454. This research was supported by the National Research Foundation of Korea (NRF) (RS-2023-00262527).

\bibliography{custom}

\newpage
\appendix
\section{Appendix}
\label{sec:appendix}

\subsection{Temporal Density Ablation}
\label{app:density_ablation_main}
\begin{figure}[H]
    \centering
    \includegraphics[width=\columnwidth]{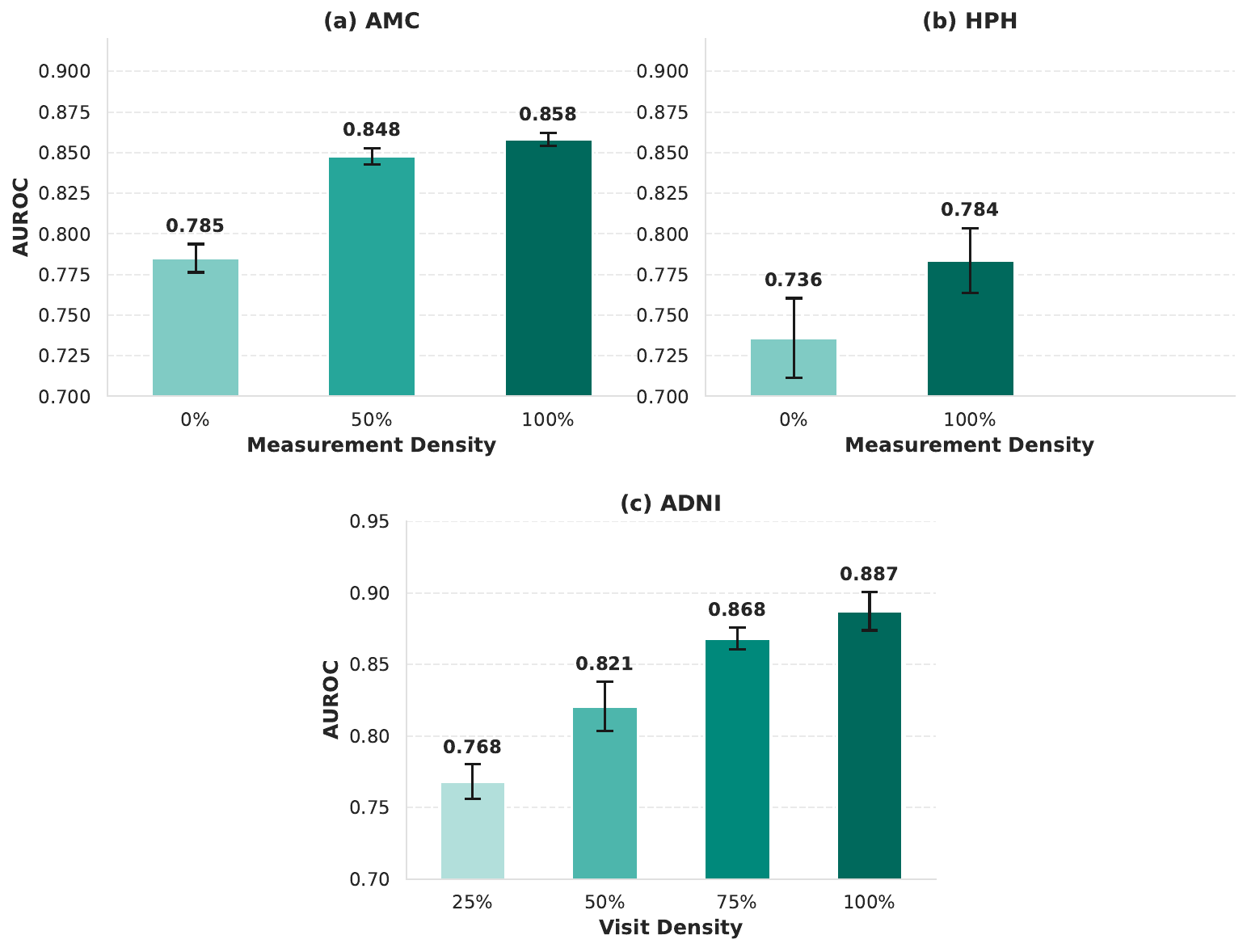}
    \vspace{-0.5cm}
    \caption{\textbf{Temporal Density Ablation.} Overall AUROC at varying measurement density levels across three cohorts. (a)~AMC: 0/50/100\%; (b)~HPH: 0/100\%; (c)~ADNI: 25/50/75/100\%. Performance degrades monotonically with reduced density.}
    \label{fig:density_ablation}
    \vspace{-0.3cm}
\end{figure}

We uniformly subsample longitudinal measurement time points to simulate reduced testing frequencies and assess the impact on prognosis (Figure~\ref{fig:density_ablation}). On AMC (456 patients, $\geq$2 visits; densities: 100/50/0\%), ADNI (342 patients, $\geq$4 visits; densities: 100/75/50/25\%), and HPH (0\% vs.\ 100\% only, due to sparse index coverage), AUROC degrades monotonically with reduced density: $-$7.3\%p on AMC (85.83$\rightarrow$78.50), $-$4.8\%p on HPH (78.37$\rightarrow$73.59), and $-$11.9\%p on ADNI (88.73$\rightarrow$76.80). The steeper ADNI decline reflects greater reliance on structured measurements in the absence of free-text narratives. Notably, even at 0\% density on AMC, the model retains 78.50\% AUROC, confirming that it has internalized reasoning from unstructured text beyond explicit scores.

\subsection{ADNI Data Description}
\label{app:adni_description}
Data used in the preparation of this article were obtained from the Alzheimer's Disease Neuroimaging Initiative (ADNI) database (\url{adni.loni.usc.edu}). The ADNI was launched in 2003 as a public-private partnership, led by Principal Investigator Michael W. Weiner, MD. The primary goal of ADNI has been to test whether serial magnetic resonance imaging (MRI), positron emission tomography (PET), other biological markers, and clinical and neuropsychological assessment can be combined to measure the progression of mild cognitive impairment (MCI) and early Alzheimer's disease (AD).

\subsection{Preprocessing Data Statistics}
\label{app:preprocessing}
\subsubsection{AMC}
\label{app:pretrain_stats}

To prevent patient-level leakage, we exclude all patients reserved for downstream fine-tuning from the Stage~1 pretraining corpus.
After removing 577 fine-tuning test patients from the original cohort of 11,163 patients, we obtain 46,746 longitudinal samples for intermediate clinical index forecasting (MMSE/GDS/CDR).
We perform a patient-level split with a test ratio of 0.20, resulting in 3,568 training patients (37,112 samples) and 892 test patients (9,634 samples).
For evaluation efficiency, the test split is task-stratified and reduced to 800 samples (401 patients).
Finally, samples exceeding 8,000 tokens under the Qwen2.5-7B-Instruct tokenizer are removed, yielding 32,681 training samples and 722 test samples.
Table~\ref{tab:appendix_pretrain_stats} summarizes the overall dataset composition, and Table~\ref{tab:appendix_pretrain_task_dist} reports the task-wise distribution.


\begin{table}[!t]
\centering
\small
\caption{\textbf{Table A1: Stage~1 (Pretraining) dataset statistics (AMC).} Fine-tuning test patients are fully excluded to prevent leakage. A patient-level split is applied for Stage~1.}
\label{tab:appendix_pretrain_stats}
\adjustbox{max width=\columnwidth}{
\begin{tabular}{lrr}
\toprule
\textbf{Item} & \textbf{\#Patients} & \textbf{\#Samples} \\
\midrule
Original cohort (raw) & 11{,}163 & -- \\
Excluded: FT test patients & 577 & -- \\
\midrule
Pretraining (after exclusion, before split) & -- & 46{,}746 \\
Train split (patient-level) & 3{,}568 & 37{,}112 \\
Test split (patient-level) & 892 & 9{,}634 \\
\midrule
After token filter ($\leq$ 8{,}000 tokens) & -- & -- \\
\quad Train kept / removed & -- & 32{,}681 / 4{,}431 \\
\quad Test kept / removed & -- & 722 / 78 \\
\bottomrule
\end{tabular}
}
\vspace{-0.3cm}
\end{table}

\begin{table}[!t]
\centering
\small
\caption{\textbf{Table A2: Task distribution for Stage~1 pretraining (AMC).}}
\label{tab:appendix_pretrain_task_dist}
\adjustbox{max width=\columnwidth}{
\begin{tabular}{lrr}
\toprule
\textbf{Task} & \textbf{Train} & \textbf{Test} \\
\midrule
MMSE & 17{,}131 & 4{,}593 \\
GDS  & 15{,}787 & 3{,}972 \\
CDR  & 4{,}194  & 1{,}069 \\
\midrule
Total & 37{,}112 & 9{,}634 \\
\bottomrule
\end{tabular}
}
\vspace{-0.3cm}
\end{table}

\subsubsection{ADNI}
\label{app:adni_pretrain_stats}

For Stage~1 pretraining on the ADNI benchmark, we construct longitudinal next-visit prediction samples across six cognitive targets
(\texttt{MMSE}, \texttt{CDRSB}, \texttt{ADAS11}, \texttt{ADAS13}, \texttt{ADASQ4}, and \texttt{RAVLT\_learning})
from linearized structured records (Sec.~\ref{app:adni_preprocessing}).
To prevent leakage, we exclude all participants reserved for the downstream fine-tuning test set.
After removing \texttt{DX} targets (not used in our pretraining), we obtain 11,319 candidate samples before filtering and keep 9,953 samples after excluding fine-tuning test participants.
No samples are removed by the token-length constraint ($\leq$ 8,000 tokens) or parsing/ID issues in our pipeline.
We bucket the time gap to the prediction target into 1-month bins up to 6 months and an additional \texttt{>6m} bin
(see Sec.~\ref{subsec:temporal_dist} for the bucket definition).
Finally, we perform a patient-level split to create Stage~1 train/test sets, resulting in 7,958 training samples and 1,995 test samples.
Table~\ref{tab:appendix_adni_pretrain_stats} summarizes the overall dataset composition, and
Table~\ref{tab:appendix_adni_pretrain_task_dist} reports task-wise statistics.

\begin{table}[!t]
\centering
\small
\caption{\textbf{Table A3: Stage~1 (Pretraining) dataset statistics (ADNI).} Fine-tuning test participants are fully excluded to prevent leakage.}
\label{tab:appendix_adni_pretrain_stats}
\adjustbox{max width=\columnwidth}{
\begin{tabular}{lr}
\toprule
\textbf{Item} & \textbf{Value} \\
\midrule
Candidate samples (before filtering; 6 tasks) & 11{,}319 \\
Kept samples ($\leq$ 8{,}000 tokens \& excl.\ FT-test) & 9{,}953 \\
Excluded: fine-tuning test participants & 1{,}366 \\
Excluded: token length / parsing / ID issues & 0 / 0 / 0 \\
\midrule
Stage~1 split (samples) & train 7{,}958; test 1{,}995 \\
\bottomrule
\end{tabular}
}
\vspace{-0.3cm}
\end{table}

\begin{table}[!t]
\centering
\small
\caption{\textbf{Table A4: Task-wise distribution for Stage~1 pretraining (ADNI).} ``Input'' counts are computed before excluding fine-tuning test participants; ``Kept'' counts are used for Stage~1 training/evaluation.}
\label{tab:appendix_adni_pretrain_task_dist}
\adjustbox{max width=\columnwidth}{
\begin{tabular}{lrrrr}
\toprule
\textbf{Task} & \textbf{Input} & \textbf{Kept} & \textbf{Train} & \textbf{Test} \\
\midrule
MMSE              & 1{,}899 & 1{,}671 & 1{,}331 & 340 \\
CDRSB             & 1{,}882 & 1{,}656 & 1{,}322 & 334 \\
ADAS11            & 1{,}891 & 1{,}663 & 1{,}311 & 352 \\
ADAS13            & 1{,}865 & 1{,}637 & 1{,}307 & 330 \\
ADASQ4            & 1{,}897 & 1{,}669 & 1{,}340 & 329 \\
RAVLT\_learning   & 1{,}885 & 1{,}657 & 1{,}347 & 310 \\
\midrule
Total             & 11{,}319 & 9{,}953 & 7{,}958 & 1{,}995 \\
\bottomrule
\end{tabular}
}
\vspace{-0.3cm}
\end{table}

To adapt distinct data modalities for our unified reasoning framework, we developed specialized preprocessing pipelines for both unstructured clinical notes (Asan) and structured tabular records (ADNI). We applied a consistent protocol consisting of Data Transformation followed by Dataset Construction.
\subsection{Data Preprocessing Details}
\subsubsection{AMC (Unstructured Clinical Notes)}
\paragraph{Data Transformation (Translation \& Extraction).}
We transformed raw Korean clinical notes into English reasoning contexts using a secure pipeline. We utilized {Qwen2.5-14B-Instruct} as an auxiliary LLM to translate notes and extract clinical indices (MMSE, GDS, and CDR) to serve as verifiable ground truth targets. Crucially, all inference processes were conducted in a strictly isolated on-premise environment to prevent any external data transmission.

\paragraph{Stage 1 Construction Pipeline.}
We constructed the pre-training dataset with the following criteria:
\begin{enumerate}
    \item \textbf{Tolerance-Aware Labeling:} We defined the prediction targets as extracted clinical indices. Recognizing extraction variability, we applied a tolerance of $\pm 2$ for MMSE, treating predictions within this range as correct. Exact matching was enforced for coarser scales (GDS, CDR).
    \item \textbf{Token Filtering:} Using the Qwen2.5 tokenizer, we filtered out samples exceeding 8,000 tokens to fit context constraints.
    \item \textbf{Evaluation Set:} The dataset was split into training and test sets at a \textbf{patient level} (80:20) to evaluate Stage 1 performance.
\end{enumerate}

\subsubsection{ADNI (Structured Tabular Data)}
\label{app:adni_preprocessing}

\paragraph{Data Transformation (Linearization).}
We transformed structured tabular records into longitudinal textual logs suitable for LLM input. For each visit, we aggregated key biomarkers---including cognitive scores (MMSE, CDR-SB, ADAS-Cog), CSF biomarkers ($A\beta$, Tau), and MRI measures---into a structured text block (e.g., ``2011-05-12: <<<VISIT 1>>> CDRSB: 0.5, MMSE: 28...''). These blocks were concatenated chronologically to form the patient history.


\paragraph{Stage 1 Construction Pipeline.}
We applied a construction protocol parallel to the AMC dataset but adapted for the continuous nature of ADNI biomarkers:

\begin{enumerate}
    \item \textbf{Target Indices:} We selected seven key indicators: MMSE, CDR-SB, ADAS-Cog (11, 13, Q4), RAVLT (Learning), and LDELTOTAL.
    
    \item \textbf{Proportional Tolerance-Aware Labeling:} 
    Unlike categorical labels, these indices vary widely in range. To standardize difficulty, we defined a relative tolerance ratio $\rho \approx 6.7\%$ (derived from the standard allowance of $\pm 2$ points on the 30-point MMSE scale). For each index, the allowable error margin $\delta$ was calculated as $\lceil \text{Range} \times \rho \rceil$. The specific thresholds are detailed in Table~\ref{tab:adni_tolerance}.
    
    \item \textbf{Token Filtering:} Samples exceeding 8,000 tokens were filtered out using the tokenizer constraints.

    \item \textbf{Evaluation Set:} Consistent with the AMC protocol, we applied a stratified \textbf{patient-level} split (80:20). Due to the high computational cost of longitudinal reasoning, the final evaluation was conducted on a stratified 50\% subsample of the test set.
\end{enumerate}

\begin{table}[!h]
\centering
\small
\caption{\textbf{Tolerance Thresholds for ADNI Indices.} Error margins ($\delta$) were scaled proportionally to the range of each metric.}
\label{tab:adni_tolerance}
\begin{tabular}{lcc}
\toprule
\textbf{Clinical Index} & \textbf{Range} & \textbf{Tolerance ($\delta$)} \\
\midrule
MMSE & 0--30 & $\pm 2$ \\
CDRSB & 0--18 & $\pm 1.0$ \\
ADAS-Cog 11 & 0--70 & $\pm 5$ \\
ADAS-Cog 13 & 0--85 & $\pm 6$ \\
ADAS-Cog Q4 & 0--10 & $\pm 1$ \\
RAVLT (Learning) & -20--20 & $\pm 3$ \\
LDELTOTAL & 0--25 & $\pm 2$ \\
\bottomrule
\end{tabular}
\end{table}

\paragraph{Choice of Linearization Format.}
Because linearizing tabular records into text could in principle discard signal, we compared our format against the best-performing serialization reported by two established references, rather than against arbitrary variants: the row-level \textit{Text Template} of TabLLM~\cite{hegselmann2023tabllm} and the table-level \textit{HTML} markup of \textit{Table Meets LLM}~\cite{sui2024table}. Our linearization corresponds to TabLLM's \textit{List Template}. We re-ran Stage~2 fine-tuning separately for each variant, reusing the same Stage~1 checkpoint and hyperparameters. As shown in Table~\ref{tab:linearization}, neither alternative meaningfully exceeds our baseline: Text Template is statistically indistinguishable from it, consistent with TabLLM's observation that format differences vanish given sufficient training data (ADNI provides 1,832 training samples), while HTML markup actively hurts, indicating that the markup-familiarity gains reported for structural-understanding tasks do not transfer to longitudinal clinical reasoning.

\begin{table}[!h]
\centering
\small
\caption{\textbf{ADNI Linearization Format Comparison.} Test AUROC (mean $\pm$ std over 5 seeds) after re-running Stage~2 for each serialization format.}
\label{tab:linearization}
\adjustbox{max width=\columnwidth}{
\begin{tabular}{lcc}
\toprule
\textbf{Linearization} & \textbf{AUROC} & \textbf{$\Delta$ vs.\ List} \\
\midrule
List Template (ours) & \textbf{0.8317} $\pm$ 0.0130 & --- \\
Text Template~\cite{hegselmann2023tabllm} & 0.8269 $\pm$ 0.0098 & $-0.48$ pp \\
HTML~\cite{sui2024table} & 0.7808 $\pm$ 0.0112 & $-5.09$ pp \\
\bottomrule
\end{tabular}
}
\end{table}

The gap between Dementia-R1 and the Random Forest baseline on ADNI (Table~\ref{tab:adni_results}) is therefore attributable to the intrinsically tabular nature of the benchmark rather than to signal loss in our linearization.

\subsection{Detailed Temporal Distributions}
\label{subsec:temporal_dist}

\begin{figure}[h!]
    \centering
    \includegraphics[width=\linewidth]{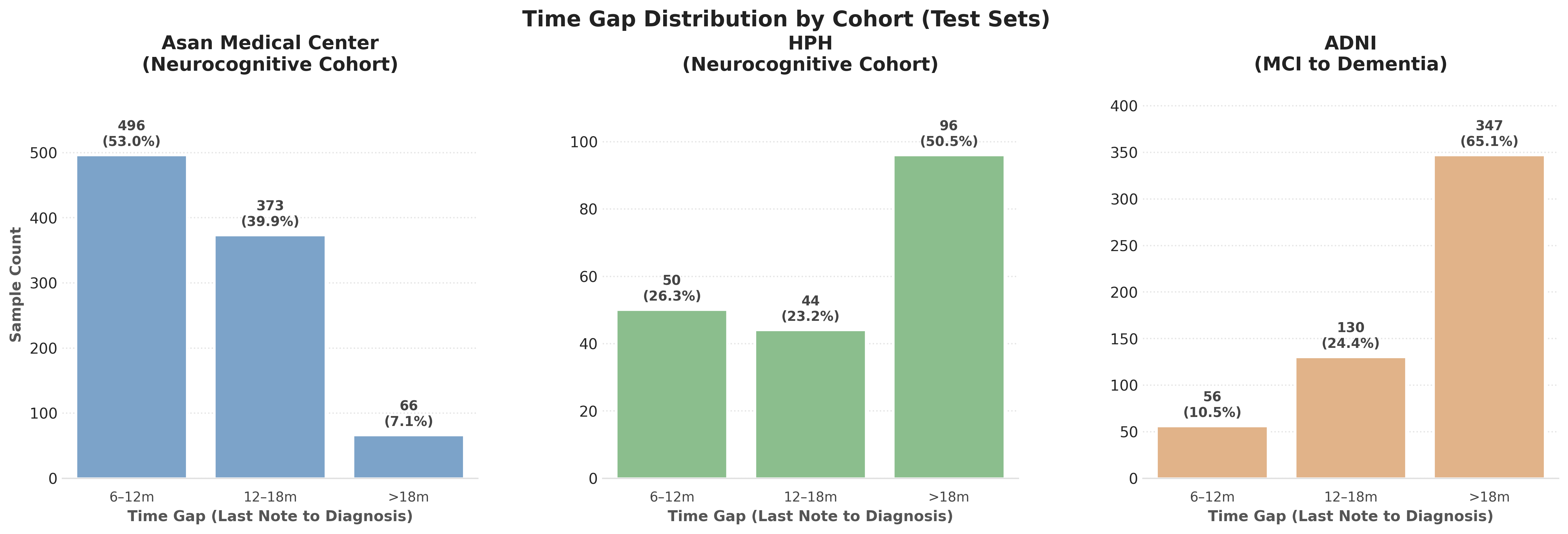}
    \caption{\textbf{Time Gap Distribution by Cohort (Test Sets).}
    The histograms show the interval between the last available clinical note and the diagnosis date across three cohorts. Intervals $\geq$18 months are consolidated into a single $>$18m bucket.
    The AMC and HPH cohorts are concentrated in shorter intervals (6--18m), reflecting relatively dense clinical follow-up prior to diagnosis.
    In contrast, the ADNI cohort displays a substantially larger proportion of long-term samples ($>$18m), reflecting the longitudinal nature of MCI progression monitoring.}
    \label{fig:appendix_gap}
\end{figure}

To validate the model's capability in long-term prediction, we analyze the time intervals between the input data and the prediction target. Figure~\ref{fig:appendix_gap} details the distribution of these time gaps for the test sets. The two real-world cohorts (AMC and HPH) are predominantly concentrated in shorter intervals (6--18m), reflecting dense clinical follow-up prior to diagnosis. In contrast, the ADNI cohort presents a substantially more challenging scenario, with a large proportion of samples falling into the $>$18m bucket, reflecting the longitudinal nature of MCI progression monitoring. This diversity ensures that our evaluation covers both immediate screening and long-term prognostic scenarios.

\subsection{Stage 1: Performance of Intermediate Clinical Indices}
\label{app:stage1_performance}

We analyze the model's capability to predict intermediate clinical indices. On the structured ADNI benchmark (Tables~\ref{tab:adni_stage1_performance}, \ref{tab:adni_task_performance}), Dementia-R1 demonstrates superior performance in long-term forecasting ($>$18 months), particularly on the critical CDR-SB metric. Similarly, on the unstructured Asan cohort (Tables~\ref{tab:asan_stage1_performance}), the model consistently outperforms baselines in short-to-mid term intervals, achieving notable gains in GDS and CDR prediction.

\begin{figure}[!t]
    \centering
    \includegraphics[width=\columnwidth]{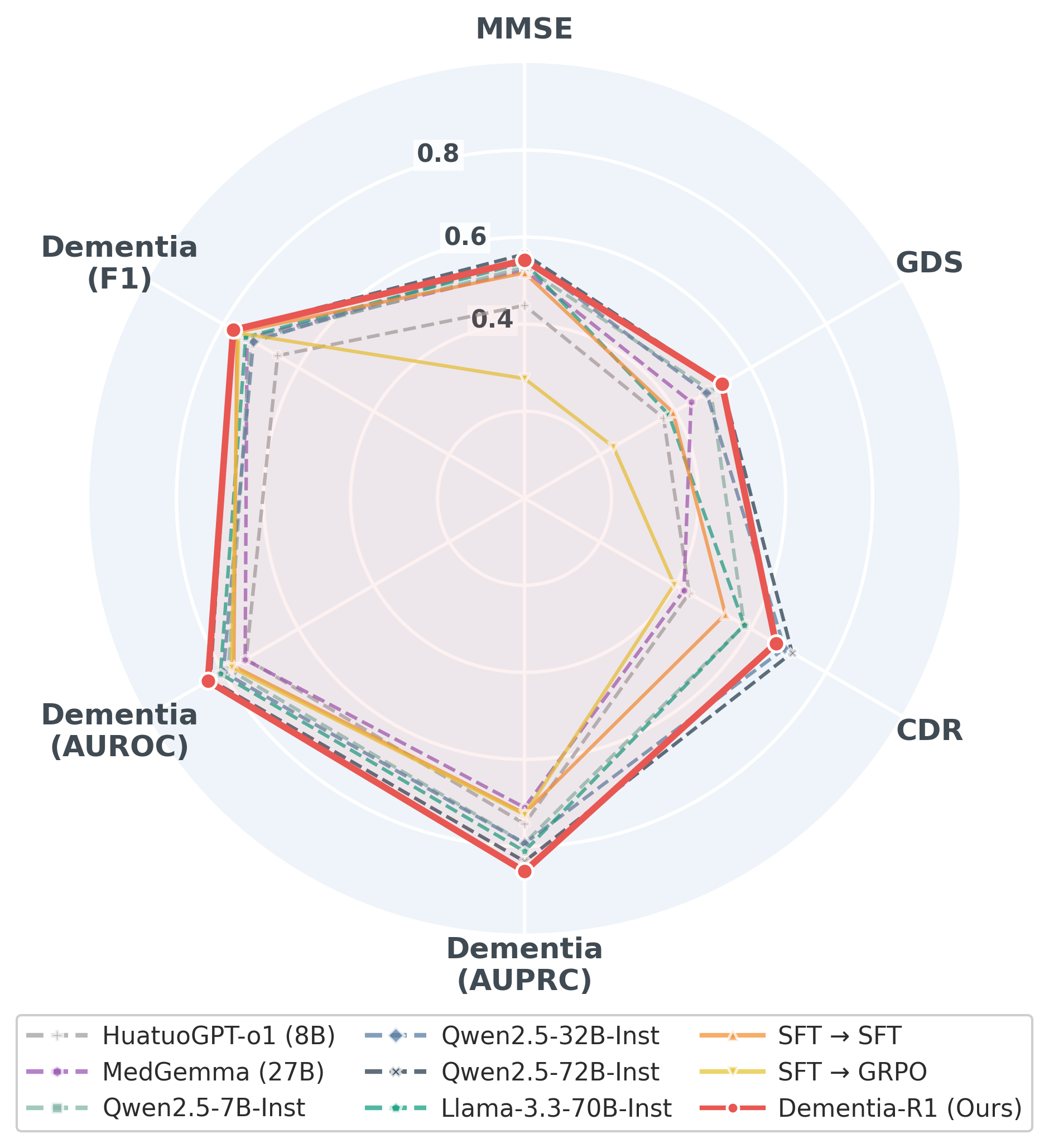}
    \vspace{-0.3cm}
    \caption{\textbf{Multi-dimensional Performance Profile (AMC).} Radar chart combining Stage~1 clinical index accuracy (MMSE, GDS, CDR) and Stage~2 dementia prognosis (F1, AUROC, AUPRC). Dementia-R1 achieves the most balanced profile overall.}
    \label{fig:radar_amc}
    \vspace{-0.3cm}
\end{figure}

\begin{figure}[!t]
    \centering
    \includegraphics[width=\columnwidth]{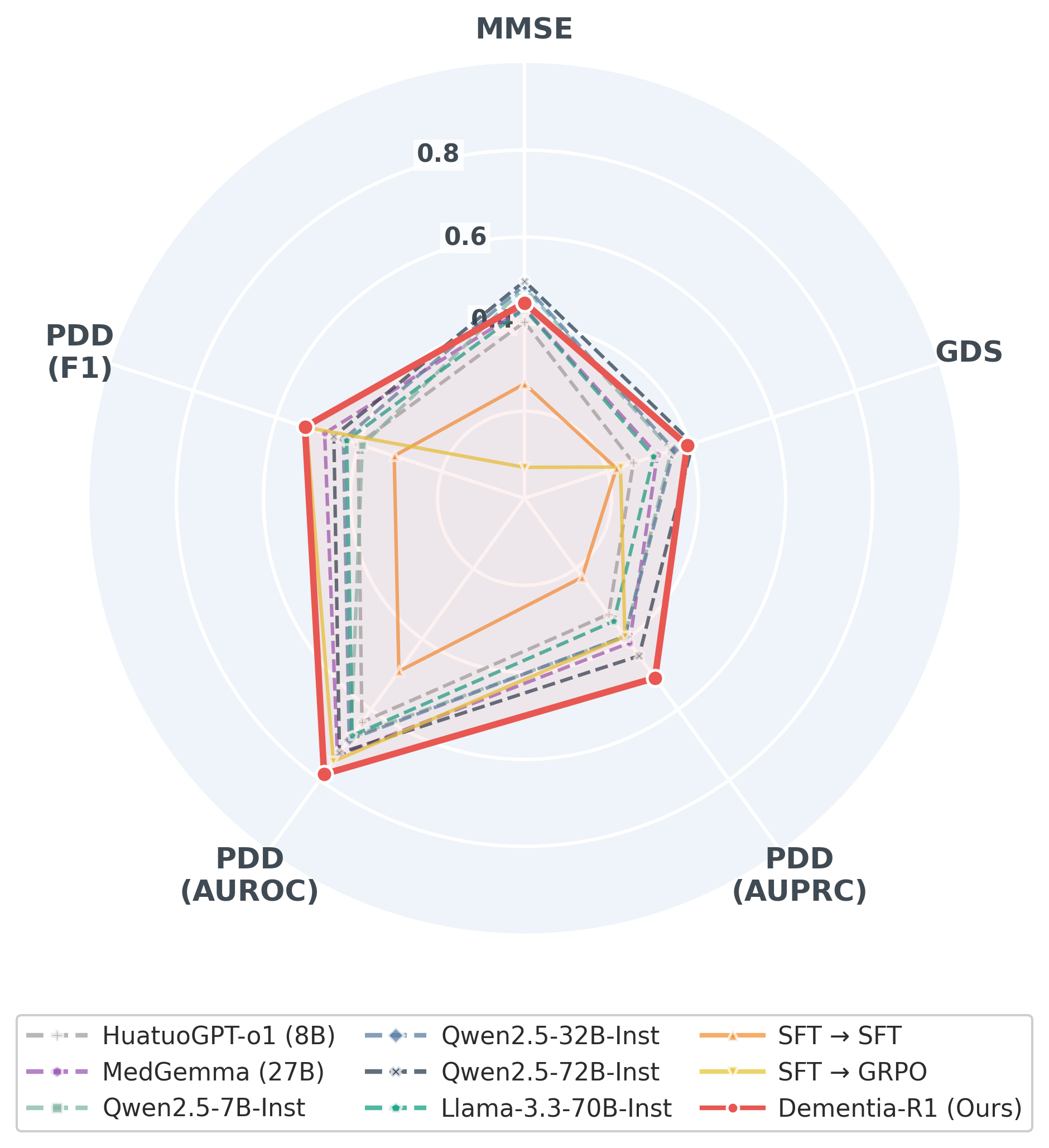}
    \vspace{-0.3cm}
    \caption{\textbf{Multi-dimensional Performance Profile (HPH).} Radar chart combining Stage~1 clinical index accuracy (MMSE, GDS) and Stage~2 PDD prognosis (F1, AUROC, AUPRC). Dementia-R1 dominates on all prognosis metrics.}
    \label{fig:radar_haeundae}
    \vspace{-0.3cm}
\end{figure}

Figures~\ref{fig:radar_amc} and~\ref{fig:radar_haeundae} visualize the multi-dimensional performance profiles for the AMC and HPH cohorts, respectively. Dementia-R1 achieves the most balanced profile across both intermediate clinical reasoning and final prognosis on each cohort.

Beyond time-gap analysis, Table~\ref{tab:asan_scores} reports the overall clinical index prediction accuracy on the AMC cohort. Dementia-R1 achieves the highest average accuracy (55.1\%), outperforming Qwen2.5-72B-Instruct (56.3\%) on GDS and matching it on MMSE---while being 10$\times$ smaller.

\begin{table}[!t]
\centering
\small
\caption{\textbf{Performance on Clinical Index Prediction for the AMC cohort.} We evaluate the accuracy of predicting MMSE, GDS, and CDR scores. \textbf{Bold}/\underline{underline} = best/second-best. Mean $\pm$ std over 3 seeds.}
\label{tab:asan_scores}
\resizebox{\columnwidth}{!}{%
\begin{tabular}{llcccc}
\toprule
\textbf{Model Method} & \textbf{Size} & \textbf{MMSE} & \textbf{GDS} & \textbf{CDR} & \textbf{Average} \\
\midrule
\textit{External LLMs} & & & & & \\
HuatuoGPT-o1 & 8B & 44.3 \textcolor{gray}{$\pm$ 0.5} & 36.9 \textcolor{gray}{$\pm$ 2.5} & 43.7 \textcolor{gray}{$\pm$ 8.7} & 41.1 \textcolor{gray}{$\pm$ 0.6} \\
MedGemma & 27B & 52.4 \textcolor{gray}{$\pm$ 0.3} & 44.2 \textcolor{gray}{$\pm$ 0.5} & 42.3 \textcolor{gray}{$\pm$ 2.1} & 47.8 \textcolor{gray}{$\pm$ 0.3} \\
Qwen2.5-7B-Inst & 7B & 53.1 \textcolor{gray}{$\pm$ 1.6} & 49.4 \textcolor{gray}{$\pm$ 1.2} & 58.3 \textcolor{gray}{$\pm$ 3.2} & 52.1 \textcolor{gray}{$\pm$ 1.5} \\
Qwen2.5-32B-Inst & 32B & 55.2 \textcolor{gray}{$\pm$ 1.2} & 48.3 \textcolor{gray}{$\pm$ 2.2} & \underline{69.0} \textcolor{gray}{$\pm$ 3.6} & 53.9 \textcolor{gray}{$\pm$ 0.2} \\
Qwen2.5-72B-Inst & 72B & \textbf{56.1} \textcolor{gray}{$\pm$ 1.3} & \underline{52.4} \textcolor{gray}{$\pm$ 1.8} & \textbf{71.0} \textcolor{gray}{$\pm$ 1.0} & \textbf{56.3} \textcolor{gray}{$\pm$ 1.0} \\
Llama-3.3-70B-Inst & 70B & 54.2 \textcolor{gray}{$\pm$ 2.2} & 38.3 \textcolor{gray}{$\pm$ 1.4} & 58.3 \textcolor{gray}{$\pm$ 2.1} & 48.0 \textcolor{gray}{$\pm$ 0.7} \\
\midrule
\textit{Specialized Models} & & & & & \\
SFT $\rightarrow$ SFT & 7B & 51.8 \textcolor{gray}{$\pm$ 2.2} & 39.4 \textcolor{gray}{$\pm$ 0.7} & 53.3 \textcolor{gray}{$\pm$ 1.5} & 46.8 \textcolor{gray}{$\pm$ 1.4} \\
SFT $\rightarrow$ GRPO & 7B & 27.5 \textcolor{gray}{$\pm$ 2.6} & 23.6 \textcolor{gray}{$\pm$ 3.2} & 39.7 \textcolor{gray}{$\pm$ 3.1} & 27.3 \textcolor{gray}{$\pm$ 1.8} \\
\textbf{Dementia-R1} & 7B & \underline{54.7} \textcolor{gray}{$\pm$ 0.6} & \textbf{52.4} \textcolor{gray}{$\pm$ 0.9} & 66.7 \textcolor{gray}{$\pm$ 2.1} & \underline{55.1} \textcolor{gray}{$\pm$ 0.4} \\
\bottomrule
\end{tabular}
}
\vspace{-0.3cm}
\end{table}

Table~\ref{tab:haeundae_scores} reports the clinical index prediction accuracy on the HPH cohort (MMSE and GDS only, as CDR was not available). Qwen2.5-72B-Instruct achieves the highest overall accuracy (45.2\%), while Dementia-R1 (7B) ranks third (42.1\%), notably outperforming all other 7B configurations and matching Qwen2.5-32B-Instruct (42.6\%).

\begin{table}[!t]
\centering
\small
\caption{\textbf{Performance on Clinical Index Prediction for the HPH cohort.} MMSE and GDS accuracy. \textbf{Bold}/\underline{underline} = best/second-best. Mean $\pm$ std over 3 seeds.}
\label{tab:haeundae_scores}
\resizebox{\columnwidth}{!}{%
\begin{tabular}{llccc}
\toprule
\textbf{Model Method} & \textbf{Size} & \textbf{MMSE} & \textbf{GDS} & \textbf{Average} \\
\midrule
\textit{External LLMs} & & & & \\
HuatuoGPT-o1 & 8B & 40.5 \textcolor{gray}{$\pm$ 4.3} & 26.3 \textcolor{gray}{$\pm$ 1.0} & 33.4 \textcolor{gray}{$\pm$ 1.7} \\
MedGemma & 27B & 43.7 \textcolor{gray}{$\pm$ 1.9} & 32.1 \textcolor{gray}{$\pm$ 1.0} & 37.9 \textcolor{gray}{$\pm$ 0.9} \\
Qwen2.5-7B-Inst & 7B & 48.4 \textcolor{gray}{$\pm$ 2.1} & 35.5 \textcolor{gray}{$\pm$ 1.2} & 41.9 \textcolor{gray}{$\pm$ 0.5} \\
Qwen2.5-32B-Inst & 32B & 49.0 \textcolor{gray}{$\pm$ 0.9} & 36.1 \textcolor{gray}{$\pm$ 0.6} & 42.6 \textcolor{gray}{$\pm$ 0.2} \\
Qwen2.5-72B-Inst & 72B & \textbf{49.9} \textcolor{gray}{$\pm$ 2.6} & \textbf{40.6} \textcolor{gray}{$\pm$ 1.6} & \textbf{45.2} \textcolor{gray}{$\pm$ 1.2} \\
Llama-3.3-70B-Inst & 70B & 43.5 \textcolor{gray}{$\pm$ 2.3} & 31.2 \textcolor{gray}{$\pm$ 0.8} & 37.4 \textcolor{gray}{$\pm$ 1.5} \\
\midrule
\textit{Specialized Models} & & & & \\
SFT $\rightarrow$ SFT & 7B & 26.3 \textcolor{gray}{$\pm$ 3.9} & 22.2 \textcolor{gray}{$\pm$ 2.7} & 24.3 \textcolor{gray}{$\pm$ 1.2} \\
SFT $\rightarrow$ GRPO & 7B & 7.1 \textcolor{gray}{$\pm$ 1.1} & 23.2 \textcolor{gray}{$\pm$ 1.3} & 15.2 \textcolor{gray}{$\pm$ 0.7} \\
\textbf{Dementia-R1} & 7B & \underline{44.9} \textcolor{gray}{$\pm$ 1.4} & \underline{39.3} \textcolor{gray}{$\pm$ 0.8} & \underline{42.1} \textcolor{gray}{$\pm$ 1.0} \\
\bottomrule
\end{tabular}
}
\vspace{-0.3cm}
\end{table}

\subsection{Stage 2: Performance of Final Binary Prognosis}
\label{app:stage2_performance}
Building on the clinical indicators established in Stage 1, Stage 2 assesses the model's ultimate capacity to determine the final binary prognosis. 
As summarized in Table~\ref{tab:asan_stage2_performance}, Dementia-R1 achieves the highest overall AUROC on the Asan dataset, demonstrating notable consistency across all time intervals (6--12m, 12--18m, 18--24m).
Similarly, the results on the ADNI benchmark (Table~\ref{tab:adni_stage2_performance}) and the HPH cohort (Table~\ref{tab:haeundae_stage2_performance}) further highlight the model's robustness across diverse prediction horizons. In particular, Dementia-R1 achieves the highest AUROC in the longest-term bucket (36m+ for ADNI; $\geq$24m for HPH), confirming its effectiveness in modeling longitudinal disease trajectories.

\subsection{Qualitative Analysis: Comparative Reasoning}

To demonstrate the impact of our proposed method on reasoning quality, we compare the outputs of Dementia-R1 against the Qwen2.5-32B model using a representative longitudinal case from the AMC cohort. Figure~\ref{fig:note_example} illustrates the input clinical note, which follows a semi-structured SOAP format. In this record, critical signals such as cognitive scores (MMSE, GDS) and medication changes are embedded within the free-text Objective and Plan sections across multiple visits spanning from 2020 to 2023. This presents a complex reasoning challenge, requiring the model to aggregate scattered clinical indicators and correctly reconstruct the patient's disease trajectory from the unstructured narrative.

While both models correctly predict the final diagnosis, their reasoning processes diverge significantly. Figure~\ref{fig:answer_example} presents the reasoning outputs generated by both models. Dementia-R1 effectively structures the longitudinal information by organizing the output into distinct sections for cognitive assessment history, diagnosis, and current status. This structural clarity allows clinicians to rapidly verify the evidence. Notably, the model accurately reconstructs the temporal trajectory of cognitive decline and correctly identifies the drop in MMSE scores from 23 down to 17 alongside the plateau in the most recent visits. Furthermore, it correctly identifies the medication switch involving the discontinuation of Bearcept and the addition of Ebixa, demonstrating precise grounding in the clinical text.

In contrast, despite arriving at the correct label, Qwen2.5-32B produces a dense, unstructured block of text that is difficult to audit for clinical decision-making. More critically, it exhibits significant failures in medical soundness and domain knowledge. First, the baseline hallucinates a pharmacological equivalence by describing "Ebixa (donepezil)" even though Ebixa is memantine, an NMDA receptor antagonist distinct from the cholinesterase inhibitor donepezil. Such hallucinations pose potential safety risks in clinical settings. Second, the baseline misinterprets the Global Deterioration Scale (GDS) scores of 3-4 as indicators of "mild depression" and confuses the dementia staging scale with a depression inventory. These findings underscore that general-purpose reasoning capabilities, even in larger models, do not automatically translate to clinical accuracy. Our results demonstrate that the domain-specific alignment integrated into Dementia-R1 is essential for correcting such misconceptions and ensuring the high reliability required for longitudinal dementia prognosis.

\begin{table*}[t!]
\centering
\small
\caption{\textbf{Stage 1 Performance by Time Gap (AMC).} \textbf{Dementia-R1} achieves superior accuracy in short-to-mid term intervals (0--18 months), validating the effectiveness of the Cold-Start strategy.}
\label{tab:asan_stage1_performance}
\adjustbox{max width=\textwidth}{
\begin{tabular}{lccccc}
\toprule
& \multicolumn{1}{c}{\textbf{Overall}} & \multicolumn{4}{c}{\textbf{Accuracy by Time Gap to Prediction Target}} \\
\cmidrule(lr){3-6}
\textbf{Method} & \textbf{Acc ($\uparrow$)} & \textbf{0--6m} & \textbf{6--12m} & \textbf{12--18m} & \textbf{$>$18m} \\
\midrule
\textit{External LLMs} & & & & & \\
Qwen2.5-32B-Inst & \underline{57.90} \textcolor{gray}{$\pm$ 0.05} & \underline{60.66} \textcolor{gray}{$\pm$ 0.43} & \underline{54.91} \textcolor{gray}{$\pm$ 0.53} & \underline{55.15} \textcolor{gray}{$\pm$ 1.21} & \textbf{50.43} \textcolor{gray}{$\pm$ 5.08} \\
Qwen2.5-7B-Inst & 54.68 \textcolor{gray}{$\pm$ 0.32} & 57.56 \textcolor{gray}{$\pm$ 0.69} & 51.19 \textcolor{gray}{$\pm$ 1.16} & 52.76 \textcolor{gray}{$\pm$ 0.99} & 46.32 \textcolor{gray}{$\pm$ 2.28} \\
\midrule
\textit{Specialized Models} & & & & & \\
SFT $\rightarrow$ SFT & 51.73 \textcolor{gray}{$\pm$ 0.53} & 55.07 \textcolor{gray}{$\pm$ 0.87} & 48.37 \textcolor{gray}{$\pm$ 0.75} & 46.95 \textcolor{gray}{$\pm$ 0.59} & 46.90 \textcolor{gray}{$\pm$ 5.02} \\
SFT $\rightarrow$ GRPO & 55.81 \textcolor{gray}{$\pm$ 0.46} & 59.60 \textcolor{gray}{$\pm$ 0.69} & 51.21 \textcolor{gray}{$\pm$ 1.15} & 52.64 \textcolor{gray}{$\pm$ 0.41} & \underline{49.62} \textcolor{gray}{$\pm$ 5.43} \\
\textbf{Dementia-R1} & \textbf{59.61} \textcolor{gray}{$\pm$ 0.46} & \textbf{63.38} \textcolor{gray}{$\pm$ 0.45} & \textbf{55.25} \textcolor{gray}{$\pm$ 0.69} & \textbf{56.41} \textcolor{gray}{$\pm$ 0.58} & 49.49 \textcolor{gray}{$\pm$ 2.84} \\
\bottomrule
\end{tabular}
}
\end{table*}

\begin{table*}[t!]
\centering
\small
\caption{\textbf{Stage 1 Performance by Time Gap (ADNI).} \textbf{Dementia-R1} demonstrates superior long-term reasoning capability ($>$18 months) compared to larger baselines (32B), validating the efficacy of the proposed RL framework.}
\label{tab:adni_stage1_performance}
\adjustbox{max width=\textwidth}{
\begin{tabular}{lccccc}
\toprule
& \multicolumn{1}{c}{\textbf{Overall}} & \multicolumn{4}{c}{\textbf{Accuracy by Time Gap to Prediction Target}} \\
\cmidrule(lr){3-6}
\textbf{Method} & \textbf{Acc ($\uparrow$)} & \textbf{0--6m} & \textbf{6--12m} & \textbf{12--18m} & \textbf{$>$18m} \\
\midrule
\textit{External LLMs} & & & & & \\
Qwen2.5-32B-Inst & \textbf{77.91} \textcolor{gray}{$\pm$ 0.18} & \textbf{81.26} \textcolor{gray}{$\pm$ 0.27} & \textbf{77.38} \textcolor{gray}{$\pm$ 0.25} & \textbf{76.77} \textcolor{gray}{$\pm$ 0.52} & \underline{80.30} \textcolor{gray}{$\pm$ 0.67} \\
Qwen2.5-7B-Inst & 75.67 \textcolor{gray}{$\pm$ 0.31} & 77.87 \textcolor{gray}{$\pm$ 0.46} & 75.09 \textcolor{gray}{$\pm$ 0.43} & 75.46 \textcolor{gray}{$\pm$ 0.58} & 78.19 \textcolor{gray}{$\pm$ 1.05} \\
\midrule
\textit{Specialized Models} & & & & & \\
SFT $\rightarrow$ SFT & 64.42 \textcolor{gray}{$\pm$ 0.23} & 66.60 \textcolor{gray}{$\pm$ 0.28} & 63.58 \textcolor{gray}{$\pm$ 0.69} & 63.26 \textcolor{gray}{$\pm$ 0.68} & 68.10 \textcolor{gray}{$\pm$ 1.74} \\
SFT $\rightarrow$ GRPO & 67.61 \textcolor{gray}{$\pm$ 0.33} & 71.26 \textcolor{gray}{$\pm$ 1.16} & 67.47 \textcolor{gray}{$\pm$ 0.62} & 65.05 \textcolor{gray}{$\pm$ 0.51} & 70.33 \textcolor{gray}{$\pm$ 1.34} \\
\textbf{Dementia-R1} & \underline{77.04} \textcolor{gray}{$\pm$ 0.28} & \underline{79.44} \textcolor{gray}{$\pm$ 0.92} & \underline{76.30} \textcolor{gray}{$\pm$ 0.51} & \underline{76.23} \textcolor{gray}{$\pm$ 0.52} & \textbf{80.62} \textcolor{gray}{$\pm$ 0.99} \\
\bottomrule
\end{tabular}
}
\end{table*}

\begin{table*}[t!]
\centering
\small
\caption{\textbf{Stage 1 Accuracy by Clinical Index (ADNI).} \textbf{Bold}/\underline{underline} = best/second-best. Mean $\pm$ std over 3 seeds.}
\label{tab:adni_task_performance}
\adjustbox{max width=\textwidth}{
\begin{tabular}{llcccccccc}
\toprule
& & \textbf{Overall} & \textbf{CDRSB} & \textbf{ADAS11} & \textbf{ADAS13} & \textbf{RAVLT} & \textbf{MMSE} & \textbf{ADASQ4} & \textbf{LDEL} \\
\textbf{Method} & \textbf{Size} & \textbf{Acc ($\uparrow$)} & \textbf{Acc ($\uparrow$)} & \textbf{Acc ($\uparrow$)} & \textbf{Acc ($\uparrow$)} & \textbf{Acc ($\uparrow$)} & \textbf{Acc ($\uparrow$)} & \textbf{Acc ($\uparrow$)} & \textbf{Acc ($\uparrow$)} \\
\midrule
\textit{External LLMs} & & & & & & & & & \\
HuatuoGPT-o1 & 8B & 62.17 \textcolor{gray}{$\pm$ 0.94} & 44.70 \textcolor{gray}{$\pm$ 1.34} & 79.10 \textcolor{gray}{$\pm$ 1.44} & 74.08 \textcolor{gray}{$\pm$ 1.12} & 70.55 \textcolor{gray}{$\pm$ 0.53} & 58.91 \textcolor{gray}{$\pm$ 1.89} & 54.00 \textcolor{gray}{$\pm$ 0.40} & 51.96 \textcolor{gray}{$\pm$ 2.00} \\
MedGemma & 27B & 74.37 \textcolor{gray}{$\pm$ 0.38} & 82.52 \textcolor{gray}{$\pm$ 0.36} & 81.45 \textcolor{gray}{$\pm$ 0.68} & 78.76 \textcolor{gray}{$\pm$ 0.42} & 80.34 \textcolor{gray}{$\pm$ 0.39} & 74.55 \textcolor{gray}{$\pm$ 0.58} & 61.72 \textcolor{gray}{$\pm$ 1.14} & 57.32 \textcolor{gray}{$\pm$ 1.39} \\
Qwen2.5-7B-Inst & 7B & 74.87 \textcolor{gray}{$\pm$ 0.48} & 81.35 \textcolor{gray}{$\pm$ 0.43} & 81.95 \textcolor{gray}{$\pm$ 0.91} & 81.22 \textcolor{gray}{$\pm$ 0.87} & 79.99 \textcolor{gray}{$\pm$ 1.22} & 75.05 \textcolor{gray}{$\pm$ 0.69} & 61.09 \textcolor{gray}{$\pm$ 1.78} & 59.93 \textcolor{gray}{$\pm$ 0.93} \\
Qwen2.5-32B-Inst & 32B & \textbf{77.94} \textcolor{gray}{$\pm$ 0.19} & \underline{85.48} \textcolor{gray}{$\pm$ 0.77} & \textbf{86.82} \textcolor{gray}{$\pm$ 0.44} & \textbf{84.28} \textcolor{gray}{$\pm$ 0.55} & 81.83 \textcolor{gray}{$\pm$ 0.80} & \textbf{77.53} \textcolor{gray}{$\pm$ 0.32} & \textbf{65.24} \textcolor{gray}{$\pm$ 0.07} & \underline{60.19} \textcolor{gray}{$\pm$ 0.48} \\
Qwen2.5-72B-Inst & 72B & 77.35 \textcolor{gray}{$\pm$ 0.18} & 85.40 \textcolor{gray}{$\pm$ 0.07} & 84.51 \textcolor{gray}{$\pm$ 0.54} & 83.68 \textcolor{gray}{$\pm$ 0.12} & \textbf{82.66} \textcolor{gray}{$\pm$ 0.15} & 76.98 \textcolor{gray}{$\pm$ 0.63} & 63.77 \textcolor{gray}{$\pm$ 0.98} & \textbf{60.46} \textcolor{gray}{$\pm$ 0.60} \\
Llama-3.3-70B-Inst & 70B & 76.03 \textcolor{gray}{$\pm$ 0.24} & \textbf{85.52} \textcolor{gray}{$\pm$ 0.25} & 82.28 \textcolor{gray}{$\pm$ 0.21} & 81.50 \textcolor{gray}{$\pm$ 0.44} & 81.26 \textcolor{gray}{$\pm$ 0.35} & 75.60 \textcolor{gray}{$\pm$ 0.77} & 62.64 \textcolor{gray}{$\pm$ 0.50} & 59.45 \textcolor{gray}{$\pm$ 0.33} \\
\midrule
\textit{Specialized Models} & & & & & & & & & \\
SFT $\rightarrow$ SFT & 7B & 63.82 \textcolor{gray}{$\pm$ 0.70} & 70.27 \textcolor{gray}{$\pm$ 1.43} & 71.21 \textcolor{gray}{$\pm$ 0.44} & 66.22 \textcolor{gray}{$\pm$ 1.41} & 73.61 \textcolor{gray}{$\pm$ 1.78} & 66.25 \textcolor{gray}{$\pm$ 0.94} & 42.98 \textcolor{gray}{$\pm$ 1.02} & 54.09 \textcolor{gray}{$\pm$ 2.38} \\
SFT $\rightarrow$ GRPO & 7B & 37.20 \textcolor{gray}{$\pm$ 0.55} & 46.60 \textcolor{gray}{$\pm$ 1.59} & 34.94 \textcolor{gray}{$\pm$ 0.12} & 30.71 \textcolor{gray}{$\pm$ 1.51} & 42.81 \textcolor{gray}{$\pm$ 3.94} & 42.10 \textcolor{gray}{$\pm$ 1.01} & 27.92 \textcolor{gray}{$\pm$ 1.12} & 35.03 \textcolor{gray}{$\pm$ 1.84} \\
\textbf{Dementia-R1} & 7B & \underline{77.38} \textcolor{gray}{$\pm$ 0.14} & 83.98 \textcolor{gray}{$\pm$ 0.42} & \underline{85.71} \textcolor{gray}{$\pm$ 0.31} & \underline{84.00} \textcolor{gray}{$\pm$ 0.07} & \underline{82.35} \textcolor{gray}{$\pm$ 0.72} & \underline{77.27} \textcolor{gray}{$\pm$ 0.86} & \underline{64.15} \textcolor{gray}{$\pm$ 0.88} & 60.14 \textcolor{gray}{$\pm$ 0.37} \\
\bottomrule
\end{tabular}
}
\end{table*}

\begin{table*}[t!]
\centering
\small
\caption{\textbf{Stage 2 Fine-tuning Performance by Time Gap (AMC).} We report the breakdown of AUROC and AUPRC across prediction horizons. Dementia-R1 consistently outperforms all baselines across all time gaps. All results represent mean $\pm$ standard deviation across five random seeds.}
\label{tab:asan_stage2_performance}
\adjustbox{max width=\textwidth}{
\begin{tabular}{lcccccccc}
\toprule
& \multicolumn{2}{c}{\textbf{Overall}} & \multicolumn{2}{c}{\textbf{6--12m}} & \multicolumn{2}{c}{\textbf{12--18m}} & \multicolumn{2}{c}{\textbf{18--24m}} \\
\cmidrule(lr){2-3} \cmidrule(lr){4-5} \cmidrule(lr){6-7} \cmidrule(lr){8-9}
\textbf{Method} & \textbf{AUROC} & \textbf{AUPRC} & \textbf{AUROC} & \textbf{AUPRC} & \textbf{AUROC} & \textbf{AUPRC} & \textbf{AUROC} & \textbf{AUPRC} \\
\midrule
\textit{External LLMs} & & & & & & & & \\
HuatuoGPT-o1 & 73.89 \textcolor{gray}{$\pm$ 0.5} & 74.81 \textcolor{gray}{$\pm$ 0.9} & 74.50 \textcolor{gray}{$\pm$ 1.0} & 73.28 \textcolor{gray}{$\pm$ 1.6} & 75.90 \textcolor{gray}{$\pm$ 1.4} & 78.21 \textcolor{gray}{$\pm$ 1.6} & 71.08 \textcolor{gray}{$\pm$ 3.5} & 80.77 \textcolor{gray}{$\pm$ 3.6} \\
Qwen2.5-7B-Inst & 78.59 \textcolor{gray}{$\pm$ 0.8} & 79.24 \textcolor{gray}{$\pm$ 0.9} & 79.41 \textcolor{gray}{$\pm$ 0.9} & 78.48 \textcolor{gray}{$\pm$ 0.7} & 80.27 \textcolor{gray}{$\pm$ 1.8} & 82.07 \textcolor{gray}{$\pm$ 1.8} & 76.62 \textcolor{gray}{$\pm$ 2.6} & 83.80 \textcolor{gray}{$\pm$ 3.2} \\
Qwen2.5-32B-Inst & 79.93 \textcolor{gray}{$\pm$ 1.2} & 79.33 \textcolor{gray}{$\pm$ 1.7} & 79.95 \textcolor{gray}{$\pm$ 1.3} & 78.13 \textcolor{gray}{$\pm$ 2.6} & 82.09 \textcolor{gray}{$\pm$ 1.1} & 82.71 \textcolor{gray}{$\pm$ 0.6} & 74.47 \textcolor{gray}{$\pm$ 3.3} & 79.54 \textcolor{gray}{$\pm$ 2.4} \\
Qwen2.5-72B-Inst & \underline{83.41} \textcolor{gray}{$\pm$ 0.4} & \underline{83.44} \textcolor{gray}{$\pm$ 0.7} & \underline{83.29} \textcolor{gray}{$\pm$ 0.7} & \underline{81.15} \textcolor{gray}{$\pm$ 1.1} & \underline{85.37} \textcolor{gray}{$\pm$ 0.6} & \underline{86.80} \textcolor{gray}{$\pm$ 1.1} & \underline{77.38} \textcolor{gray}{$\pm$ 2.6} & 84.43 \textcolor{gray}{$\pm$ 3.6} \\
Llama-3.3-70B-Inst & 80.70 \textcolor{gray}{$\pm$ 0.3} & 81.11 \textcolor{gray}{$\pm$ 0.7} & 80.39 \textcolor{gray}{$\pm$ 0.5} & 79.03 \textcolor{gray}{$\pm$ 0.9} & 83.00 \textcolor{gray}{$\pm$ 0.7} & 84.52 \textcolor{gray}{$\pm$ 0.7} & 73.61 \textcolor{gray}{$\pm$ 1.9} & 80.58 \textcolor{gray}{$\pm$ 3.2} \\
MedGemma-27B & 74.14 \textcolor{gray}{$\pm$ 0.8} & 71.06 \textcolor{gray}{$\pm$ 1.2} & 74.65 \textcolor{gray}{$\pm$ 1.6} & 70.17 \textcolor{gray}{$\pm$ 1.9} & 75.78 \textcolor{gray}{$\pm$ 0.7} & 73.37 \textcolor{gray}{$\pm$ 1.3} & 71.56 \textcolor{gray}{$\pm$ 1.6} & 76.88 \textcolor{gray}{$\pm$ 2.6} \\
\midrule
\textit{Specialized Models} & & & & & & & & \\
SFT \,\,\,\, w/o Stage 1 & 71.65 \textcolor{gray}{$\pm$ 1.9} & 69.82 \textcolor{gray}{$\pm$ 2.2} & 70.59 \textcolor{gray}{$\pm$ 1.2} & 65.63 \textcolor{gray}{$\pm$ 1.7} & 73.83 \textcolor{gray}{$\pm$ 2.6} & 74.63 \textcolor{gray}{$\pm$ 3.1} & 70.24 \textcolor{gray}{$\pm$ 9.4} & 78.91 \textcolor{gray}{$\pm$ 7.5} \\
GRPO w/o Stage 1 & 81.68 \textcolor{gray}{$\pm$ 0.4} & 81.27 \textcolor{gray}{$\pm$ 0.3} & 81.97 \textcolor{gray}{$\pm$ 0.8} & 79.67 \textcolor{gray}{$\pm$ 1.2} & 84.16 \textcolor{gray}{$\pm$ 0.9} & 85.22 \textcolor{gray}{$\pm$ 1.8} & 76.07 \textcolor{gray}{$\pm$ 2.1} & \underline{84.47} \textcolor{gray}{$\pm$ 2.8} \\
SFT $\rightarrow$ SFT & 77.15 \textcolor{gray}{$\pm$ 1.9} & 72.41 \textcolor{gray}{$\pm$ 2.4} & 77.20 \textcolor{gray}{$\pm$ 2.1} & 69.59 \textcolor{gray}{$\pm$ 2.8} & 78.28 \textcolor{gray}{$\pm$ 1.8} & 75.66 \textcolor{gray}{$\pm$ 2.2} & 72.59 \textcolor{gray}{$\pm$ 7.1} & 79.77 \textcolor{gray}{$\pm$ 7.0} \\
SFT $\rightarrow$ GRPO & 77.75 \textcolor{gray}{$\pm$ 1.7} & 72.69 \textcolor{gray}{$\pm$ 2.3} & 78.72 \textcolor{gray}{$\pm$ 2.5} & 71.23 \textcolor{gray}{$\pm$ 3.8} & 78.18 \textcolor{gray}{$\pm$ 1.2} & 74.34 \textcolor{gray}{$\pm$ 2.0} & 70.52 \textcolor{gray}{$\pm$ 4.3} & 79.03 \textcolor{gray}{$\pm$ 4.1} \\
\textbf{Dementia-R1} & \textbf{84.02} \textcolor{gray}{$\pm$ 0.4} & \textbf{85.73} \textcolor{gray}{$\pm$ 0.4} & \textbf{84.67} \textcolor{gray}{$\pm$ 0.6} & \textbf{85.40} \textcolor{gray}{$\pm$ 0.6} & \textbf{85.39} \textcolor{gray}{$\pm$ 0.4} & \textbf{87.44} \textcolor{gray}{$\pm$ 0.5} & \textbf{78.51} \textcolor{gray}{$\pm$ 2.7} & \textbf{86.15} \textcolor{gray}{$\pm$ 1.7} \\
\bottomrule
\end{tabular}
}
\end{table*}

\begin{table*}[t!]
\centering
\small
\caption{\textbf{Stage 2 Fine-tuning Performance by Time Gap (HPH).} We report the breakdown of AUROC and AUPRC across prediction horizons. All results represent mean $\pm$ standard deviation across five random seeds.}
\label{tab:haeundae_stage2_performance}
\adjustbox{max width=\textwidth}{
\begin{tabular}{lcccccccccc}
\toprule
& \multicolumn{2}{c}{\textbf{Overall}} & \multicolumn{2}{c}{\textbf{6--12m}} & \multicolumn{2}{c}{\textbf{12--18m}} & \multicolumn{2}{c}{\textbf{18--24m}} & \multicolumn{2}{c}{\textbf{$\geq$24m}} \\
\cmidrule(lr){2-3} \cmidrule(lr){4-5} \cmidrule(lr){6-7} \cmidrule(lr){8-9} \cmidrule(lr){10-11}
\textbf{Method} & \textbf{AUROC} & \textbf{AUPRC} & \textbf{AUROC} & \textbf{AUPRC} & \textbf{AUROC} & \textbf{AUPRC} & \textbf{AUROC} & \textbf{AUPRC} & \textbf{AUROC} & \textbf{AUPRC} \\
\midrule
\textit{External LLMs} & & & & & & & & & & \\
HuatuoGPT-o1 & 63.58 \textcolor{gray}{$\pm$ 3.2} & 32.93 \textcolor{gray}{$\pm$ 1.9} & 58.51 \textcolor{gray}{$\pm$ 5.9} & 38.06 \textcolor{gray}{$\pm$ 6.5} & 69.18 \textcolor{gray}{$\pm$ 3.6} & 50.94 \textcolor{gray}{$\pm$ 7.3} & 71.76 \textcolor{gray}{$\pm$ 7.3} & 47.81 \textcolor{gray}{$\pm$ 11.9} & 57.44 \textcolor{gray}{$\pm$ 6.9} & 24.62 \textcolor{gray}{$\pm$ 4.5} \\
Qwen2.5-7B-Inst & 67.90 \textcolor{gray}{$\pm$ 3.2} & 39.28 \textcolor{gray}{$\pm$ 4.1} & 66.12 \textcolor{gray}{$\pm$ 4.4} & 45.42 \textcolor{gray}{$\pm$ 5.3} & 70.79 \textcolor{gray}{$\pm$ 5.4} & 52.81 \textcolor{gray}{$\pm$ 7.6} & 78.70 \textcolor{gray}{$\pm$ 9.0} & 63.06 \textcolor{gray}{$\pm$ 10.2} & 60.07 \textcolor{gray}{$\pm$ 6.5} & 24.60 \textcolor{gray}{$\pm$ 2.5} \\
Qwen2.5-32B-Inst & 68.54 \textcolor{gray}{$\pm$ 1.9} & 39.20 \textcolor{gray}{$\pm$ 1.9} & 74.12 \textcolor{gray}{$\pm$ 4.0} & 55.05 \textcolor{gray}{$\pm$ 9.3} & 72.26 \textcolor{gray}{$\pm$ 4.6} & \underline{63.83} \textcolor{gray}{$\pm$ 3.8} & 74.92 \textcolor{gray}{$\pm$ 8.3} & 50.85 \textcolor{gray}{$\pm$ 7.9} & 55.68 \textcolor{gray}{$\pm$ 8.7} & 24.42 \textcolor{gray}{$\pm$ 5.1} \\
Qwen2.5-72B-Inst & 72.37 \textcolor{gray}{$\pm$ 1.1} & \underline{44.78} \textcolor{gray}{$\pm$ 2.6} & 77.52 \textcolor{gray}{$\pm$ 4.1} & \textbf{58.52} \textcolor{gray}{$\pm$ 4.6} & 77.57 \textcolor{gray}{$\pm$ 1.9} & \textbf{68.05} \textcolor{gray}{$\pm$ 1.7} & 78.28 \textcolor{gray}{$\pm$ 5.9} & 62.97 \textcolor{gray}{$\pm$ 6.9} & 61.52 \textcolor{gray}{$\pm$ 1.4} & 32.25 \textcolor{gray}{$\pm$ 6.1} \\
Llama-3.3-70B-Inst & 67.41 \textcolor{gray}{$\pm$ 1.4} & 34.91 \textcolor{gray}{$\pm$ 0.7} & 72.57 \textcolor{gray}{$\pm$ 3.9} & 55.01 \textcolor{gray}{$\pm$ 7.7} & 74.91 \textcolor{gray}{$\pm$ 2.4} & 55.84 \textcolor{gray}{$\pm$ 3.2} & 68.99 \textcolor{gray}{$\pm$ 5.2} & 31.43 \textcolor{gray}{$\pm$ 2.4} & 56.96 \textcolor{gray}{$\pm$ 3.8} & 28.34 \textcolor{gray}{$\pm$ 3.7} \\
MedGemma-27B & 73.13 \textcolor{gray}{$\pm$ 2.6} & 41.06 \textcolor{gray}{$\pm$ 4.4} & \underline{77.63} \textcolor{gray}{$\pm$ 4.1} & 50.20 \textcolor{gray}{$\pm$ 5.9} & \underline{81.76} \textcolor{gray}{$\pm$ 2.5} & 62.69 \textcolor{gray}{$\pm$ 5.3} & 72.65 \textcolor{gray}{$\pm$ 5.6} & 39.15 \textcolor{gray}{$\pm$ 8.4} & 58.55 \textcolor{gray}{$\pm$ 5.7} & \underline{32.75} \textcolor{gray}{$\pm$ 8.2} \\
\midrule
\textit{Specialized Models} & & & & & & & & & & \\
SFT \,\,\,\, w/o Stage 1 & 63.75 \textcolor{gray}{$\pm$ 1.8} & 30.39 \textcolor{gray}{$\pm$ 3.2} & 64.78 \textcolor{gray}{$\pm$ 9.8} & 38.41 \textcolor{gray}{$\pm$ 13.0} & 61.74 \textcolor{gray}{$\pm$ 5.3} & 42.81 \textcolor{gray}{$\pm$ 7.1} & 75.50 \textcolor{gray}{$\pm$ 4.9} & 37.11 \textcolor{gray}{$\pm$ 3.0} & 57.92 \textcolor{gray}{$\pm$ 13.2} & 24.50 \textcolor{gray}{$\pm$ 9.6} \\
GRPO w/o Stage 1 & 69.96 \textcolor{gray}{$\pm$ 3.4} & 38.69 \textcolor{gray}{$\pm$ 2.4} & 65.99 \textcolor{gray}{$\pm$ 9.7} & 44.93 \textcolor{gray}{$\pm$ 9.5} & 71.32 \textcolor{gray}{$\pm$ 4.7} & 55.65 \textcolor{gray}{$\pm$ 6.8} & \underline{84.96} \textcolor{gray}{$\pm$ 4.7} & \underline{64.54} \textcolor{gray}{$\pm$ 4.3} & \underline{65.48} \textcolor{gray}{$\pm$ 5.4} & 27.15 \textcolor{gray}{$\pm$ 4.0} \\
SFT $\rightarrow$ SFT & 49.14 \textcolor{gray}{$\pm$ 1.2} & 22.47 \textcolor{gray}{$\pm$ 1.7} & 49.10 \textcolor{gray}{$\pm$ 6.8} & 28.04 \textcolor{gray}{$\pm$ 4.5} & 48.41 \textcolor{gray}{$\pm$ 7.1} & 34.50 \textcolor{gray}{$\pm$ 7.0} & 49.87 \textcolor{gray}{$\pm$ 10.4} & 22.16 \textcolor{gray}{$\pm$ 5.4} & 49.73 \textcolor{gray}{$\pm$ 4.8} & 20.33 \textcolor{gray}{$\pm$ 4.2} \\
SFT $\rightarrow$ GRPO & \underline{74.75} \textcolor{gray}{$\pm$ 2.6} & 39.25 \textcolor{gray}{$\pm$ 3.3} & \textbf{79.23} \textcolor{gray}{$\pm$ 4.6} & 51.57 \textcolor{gray}{$\pm$ 7.6} & 80.62 \textcolor{gray}{$\pm$ 4.3} & 61.43 \textcolor{gray}{$\pm$ 10.1} & 75.97 \textcolor{gray}{$\pm$ 6.7} & 39.07 \textcolor{gray}{$\pm$ 9.4} & 62.61 \textcolor{gray}{$\pm$ 6.7} & 26.32 \textcolor{gray}{$\pm$ 7.2} \\
\textbf{Dementia-R1} & \textbf{78.37} \textcolor{gray}{$\pm$ 1.8} & \textbf{51.03} \textcolor{gray}{$\pm$ 2.7} & 75.50 \textcolor{gray}{$\pm$ 3.8} & \underline{56.18} \textcolor{gray}{$\pm$ 4.6} & \textbf{81.96} \textcolor{gray}{$\pm$ 3.2} & 61.35 \textcolor{gray}{$\pm$ 3.5} & \textbf{93.99} \textcolor{gray}{$\pm$ 3.4} & \textbf{84.76} \textcolor{gray}{$\pm$ 6.7} & \textbf{68.60} \textcolor{gray}{$\pm$ 4.9} & \textbf{35.79} \textcolor{gray}{$\pm$ 6.9} \\
\bottomrule
\end{tabular}
}
\end{table*}

\begin{table*}[t!]
\centering
\small
\caption{\textbf{Stage 2 Fine-tuning Performance by Time Gap (ADNI).}
While large-scale general models (e.g., Qwen2.5-32B) show strong performance in certain horizons, \textbf{Dementia-R1} demonstrates superior robustness in the longest-term reasoning ($>$36 months). All results represent mean $\pm$ standard deviation across five random seeds.}
\label{tab:adni_stage2_performance}
\adjustbox{max width=\textwidth}{
\begin{tabular}{l cccccccccccccc}
\toprule
& \multicolumn{2}{c}{\textbf{Overall}} & \multicolumn{2}{c}{\textbf{6--12m}} & \multicolumn{2}{c}{\textbf{12--18m}} & \multicolumn{2}{c}{\textbf{18--24m}} & \multicolumn{2}{c}{\textbf{24--30m}} & \multicolumn{2}{c}{\textbf{30--36m}} & \multicolumn{2}{c}{\textbf{36m+}} \\
\cmidrule(lr){2-3} \cmidrule(lr){4-5} \cmidrule(lr){6-7} \cmidrule(lr){8-9} \cmidrule(lr){10-11} \cmidrule(lr){12-13} \cmidrule(lr){14-15}
\textbf{Method} & \textbf{AUROC} & \textbf{AUPRC} & \textbf{AUROC} & \textbf{AUPRC} & \textbf{AUROC} & \textbf{AUPRC} & \textbf{AUROC} & \textbf{AUPRC} & \textbf{AUROC} & \textbf{AUPRC} & \textbf{AUROC} & \textbf{AUPRC} & \textbf{AUROC} & \textbf{AUPRC} \\
\midrule
\textit{External LLMs} & & & & & & & & & & & & & & \\
HuatuoGPT-o1 & 71.69 \textcolor{gray}{$\pm$ 2.0} & 67.34 \textcolor{gray}{$\pm$ 3.2} & 76.93 \textcolor{gray}{$\pm$ 2.6} & 76.76 \textcolor{gray}{$\pm$ 4.5} & 75.92 \textcolor{gray}{$\pm$ 1.6} & 69.42 \textcolor{gray}{$\pm$ 4.2} & 71.64 \textcolor{gray}{$\pm$ 6.4} & 70.70 \textcolor{gray}{$\pm$ 6.7} & 70.91 \textcolor{gray}{$\pm$ 1.2} & 68.35 \textcolor{gray}{$\pm$ 2.8} & 71.23 \textcolor{gray}{$\pm$ 2.4} & 69.70 \textcolor{gray}{$\pm$ 3.3} & 62.15 \textcolor{gray}{$\pm$ 4.1} & 56.16 \textcolor{gray}{$\pm$ 6.4} \\
Qwen2.5-7B-Inst & 66.23 \textcolor{gray}{$\pm$ 1.9} & 57.17 \textcolor{gray}{$\pm$ 2.4} & 78.26 \textcolor{gray}{$\pm$ 4.5} & 75.61 \textcolor{gray}{$\pm$ 6.6} & 67.74 \textcolor{gray}{$\pm$ 4.3} & 56.48 \textcolor{gray}{$\pm$ 5.3} & 62.61 \textcolor{gray}{$\pm$ 2.3} & 62.58 \textcolor{gray}{$\pm$ 3.0} & 65.42 \textcolor{gray}{$\pm$ 3.1} & 57.23 \textcolor{gray}{$\pm$ 4.3} & 69.28 \textcolor{gray}{$\pm$ 2.3} & 65.42 \textcolor{gray}{$\pm$ 4.0} & 53.42 \textcolor{gray}{$\pm$ 4.5} & 42.69 \textcolor{gray}{$\pm$ 4.7} \\
Qwen2.5-32B-Inst & \underline{82.51} \textcolor{gray}{$\pm$ 0.6} & \underline{83.50} \textcolor{gray}{$\pm$ 0.6} & 83.34 \textcolor{gray}{$\pm$ 1.6} & 83.24 \textcolor{gray}{$\pm$ 2.1} & \textbf{84.90} \textcolor{gray}{$\pm$ 1.2} & \textbf{86.09} \textcolor{gray}{$\pm$ 1.1} & 81.70 \textcolor{gray}{$\pm$ 1.4} & 85.24 \textcolor{gray}{$\pm$ 1.7} & \textbf{83.22} \textcolor{gray}{$\pm$ 1.0} & \textbf{82.90} \textcolor{gray}{$\pm$ 1.6} & \textbf{85.00} \textcolor{gray}{$\pm$ 1.1} & 86.92 \textcolor{gray}{$\pm$ 1.9} & 74.48 \textcolor{gray}{$\pm$ 1.7} & \underline{76.27} \textcolor{gray}{$\pm$ 2.1} \\
Qwen2.5-72B-Inst & 81.45 \textcolor{gray}{$\pm$ 0.6} & \textbf{83.56} \textcolor{gray}{$\pm$ 0.7} & 82.49 \textcolor{gray}{$\pm$ 1.3} & 83.60 \textcolor{gray}{$\pm$ 1.3} & 84.47 \textcolor{gray}{$\pm$ 1.2} & 85.38 \textcolor{gray}{$\pm$ 2.1} & \underline{82.11} \textcolor{gray}{$\pm$ 1.9} & \textbf{87.81} \textcolor{gray}{$\pm$ 0.8} & 82.05 \textcolor{gray}{$\pm$ 0.9} & \underline{82.80} \textcolor{gray}{$\pm$ 1.8} & 83.56 \textcolor{gray}{$\pm$ 1.7} & \underline{87.31} \textcolor{gray}{$\pm$ 1.5} & 69.74 \textcolor{gray}{$\pm$ 5.5} & 73.47 \textcolor{gray}{$\pm$ 3.2} \\
Llama-3.3-70B-Inst & 81.17 \textcolor{gray}{$\pm$ 0.5} & 83.48 \textcolor{gray}{$\pm$ 0.6} & 83.26 \textcolor{gray}{$\pm$ 0.9} & \textbf{85.98} \textcolor{gray}{$\pm$ 1.0} & \underline{84.47} \textcolor{gray}{$\pm$ 0.8} & \underline{85.71} \textcolor{gray}{$\pm$ 1.4} & 80.65 \textcolor{gray}{$\pm$ 0.9} & \underline{86.12} \textcolor{gray}{$\pm$ 0.5} & 81.51 \textcolor{gray}{$\pm$ 1.1} & 82.79 \textcolor{gray}{$\pm$ 1.1} & \underline{83.68} \textcolor{gray}{$\pm$ 0.6} & \textbf{87.37} \textcolor{gray}{$\pm$ 0.8} & 70.20 \textcolor{gray}{$\pm$ 1.5} & 75.41 \textcolor{gray}{$\pm$ 0.8} \\
MedGemma-27B & 78.48 \textcolor{gray}{$\pm$ 1.1} & 71.71 \textcolor{gray}{$\pm$ 1.2} & 75.94 \textcolor{gray}{$\pm$ 3.7} & 69.16 \textcolor{gray}{$\pm$ 4.5} & 81.50 \textcolor{gray}{$\pm$ 1.6} & 75.88 \textcolor{gray}{$\pm$ 3.7} & 77.08 \textcolor{gray}{$\pm$ 2.8} & 72.71 \textcolor{gray}{$\pm$ 2.8} & 79.65 \textcolor{gray}{$\pm$ 2.1} & 71.85 \textcolor{gray}{$\pm$ 4.7} & 75.70 \textcolor{gray}{$\pm$ 3.1} & 74.52 \textcolor{gray}{$\pm$ 2.3} & \underline{79.57} \textcolor{gray}{$\pm$ 3.2} & 72.01 \textcolor{gray}{$\pm$ 4.9} \\
\midrule
\textit{Specialized Models} & & & & & & & & & & & & & & \\
SFT \,\,\,\, w/o Stage 1 & 73.40 \textcolor{gray}{$\pm$ 1.9} & 63.03 \textcolor{gray}{$\pm$ 2.4} & 73.68 \textcolor{gray}{$\pm$ 5.3} & 63.82 \textcolor{gray}{$\pm$ 7.7} & 74.47 \textcolor{gray}{$\pm$ 2.0} & 65.14 \textcolor{gray}{$\pm$ 5.9} & 76.01 \textcolor{gray}{$\pm$ 4.9} & 71.43 \textcolor{gray}{$\pm$ 6.4} & 74.02 \textcolor{gray}{$\pm$ 3.9} & 62.70 \textcolor{gray}{$\pm$ 6.6} & 71.99 \textcolor{gray}{$\pm$ 3.4} & 67.46 \textcolor{gray}{$\pm$ 5.3} & 69.98 \textcolor{gray}{$\pm$ 7.0} & 59.53 \textcolor{gray}{$\pm$ 7.4} \\
GRPO w/o Stage 1 & 79.07 \textcolor{gray}{$\pm$ 0.9} & 77.65 \textcolor{gray}{$\pm$ 1.5} & \underline{83.72} \textcolor{gray}{$\pm$ 2.7} & \underline{85.75} \textcolor{gray}{$\pm$ 3.5} & 81.03 \textcolor{gray}{$\pm$ 1.2} & 77.37 \textcolor{gray}{$\pm$ 2.9} & 78.23 \textcolor{gray}{$\pm$ 3.8} & 80.89 \textcolor{gray}{$\pm$ 4.4} & 79.60 \textcolor{gray}{$\pm$ 1.4} & 76.71 \textcolor{gray}{$\pm$ 3.7} & 79.08 \textcolor{gray}{$\pm$ 2.0} & 80.42 \textcolor{gray}{$\pm$ 3.8} & 70.45 \textcolor{gray}{$\pm$ 4.9} & 66.69 \textcolor{gray}{$\pm$ 3.8} \\
SFT $\rightarrow$ SFT & 79.90 \textcolor{gray}{$\pm$ 0.3} & 73.77 \textcolor{gray}{$\pm$ 1.2} & 77.43 \textcolor{gray}{$\pm$ 2.3} & 72.61 \textcolor{gray}{$\pm$ 4.8} & 83.89 \textcolor{gray}{$\pm$ 3.6} & 77.04 \textcolor{gray}{$\pm$ 5.1} & 81.89 \textcolor{gray}{$\pm$ 1.4} & 82.10 \textcolor{gray}{$\pm$ 5.5} & 79.22 \textcolor{gray}{$\pm$ 2.1} & 71.38 \textcolor{gray}{$\pm$ 2.5} & 77.71 \textcolor{gray}{$\pm$ 1.8} & 76.96 \textcolor{gray}{$\pm$ 2.8} & 78.18 \textcolor{gray}{$\pm$ 4.7} & 72.28 \textcolor{gray}{$\pm$ 7.8} \\
SFT $\rightarrow$ GRPO & 77.88 \textcolor{gray}{$\pm$ 1.7} & 69.09 \textcolor{gray}{$\pm$ 2.6} & 77.41 \textcolor{gray}{$\pm$ 5.3} & 71.21 \textcolor{gray}{$\pm$ 7.6} & 80.99 \textcolor{gray}{$\pm$ 3.6} & 70.29 \textcolor{gray}{$\pm$ 6.4} & 77.42 \textcolor{gray}{$\pm$ 2.3} & 71.70 \textcolor{gray}{$\pm$ 2.6} & 76.84 \textcolor{gray}{$\pm$ 5.6} & 67.88 \textcolor{gray}{$\pm$ 5.7} & 79.02 \textcolor{gray}{$\pm$ 3.8} & 76.30 \textcolor{gray}{$\pm$ 3.3} & 73.35 \textcolor{gray}{$\pm$ 3.5} & 70.50 \textcolor{gray}{$\pm$ 8.1} \\
\textbf{Dementia-R1} & \textbf{83.17} \textcolor{gray}{$\pm$ 1.5} & 82.00 \textcolor{gray}{$\pm$ 1.2} & \textbf{85.39} \textcolor{gray}{$\pm$ 3.5} & 85.29 \textcolor{gray}{$\pm$ 4.1} & 83.87 \textcolor{gray}{$\pm$ 1.9} & 80.96 \textcolor{gray}{$\pm$ 3.5} & \textbf{83.85} \textcolor{gray}{$\pm$ 3.7} & 85.09 \textcolor{gray}{$\pm$ 3.2} & \underline{82.13} \textcolor{gray}{$\pm$ 1.9} & 81.59 \textcolor{gray}{$\pm$ 3.3} & 82.66 \textcolor{gray}{$\pm$ 1.6} & 85.41 \textcolor{gray}{$\pm$ 1.1} & \textbf{83.90} \textcolor{gray}{$\pm$ 2.8} & \textbf{81.13} \textcolor{gray}{$\pm$ 4.1} \\
\bottomrule
\end{tabular}
}
\end{table*}

\subsection{Llama-3.1 Backbone Results}
\label{app:llama_results}
To verify that our framework generalizes beyond the Qwen backbone, we evaluate all training configurations using Llama-3.1-8B-Instruct as the base model. As shown in Table~\ref{tab:asan_llama_results}, Dementia-R1 achieves the highest F1 (77.90\%), AUROC (85.29\%), and AUPRC (85.58\%) among all Llama-based variants, with significant improvements ($p < 0.01$).

\begin{table*}[t!]
\centering
\small
\caption{\textbf{Experimental Results on AMC Dataset (Llama-3.1 Backbone).} We compare Dementia-R1 against Llama-3.1-8B-Instruct and its specialized variants. \textbf{Bold} and \underline{underline} indicate the best and second-best performance. All results represent mean $\pm$ standard deviation across five random seeds. Statistical significance compared to Dementia-R1 is denoted by $^{\ast}$ ($p < 0.05$) and $^{\ast\ast}$ ($p < 0.01$).}
\label{tab:asan_llama_results}
\adjustbox{width=0.98\textwidth}{
\begin{tabular}{lccccccc}
\toprule
\textbf{Method} & \textbf{Size} & \textbf{Accuracy ($\uparrow$)} & \textbf{Precision ($\uparrow$)} & \textbf{Recall ($\uparrow$)} & \textbf{F1 score ($\uparrow$)} & \textbf{AUROC ($\uparrow$)} & \textbf{AUPRC ($\uparrow$)} \\
\midrule
\textit{External LLMs} & & & & & & & \\
Llama-3.1-8B-Inst & 8B & 54.37 \textcolor{gray}{$\pm$ 2.0}\rlap{$^{\ast\ast}$} & 68.94 \textcolor{gray}{$\pm$ 1.8}\rlap{$^{\ast\ast}$} & 58.91 \textcolor{gray}{$\pm$ 1.8}\rlap{$^{\ast\ast}$} & 63.53 \textcolor{gray}{$\pm$ 1.8}\rlap{$^{\ast\ast}$} & 66.66 \textcolor{gray}{$\pm$ 1.4}\rlap{$^{\ast\ast}$} & 63.44 \textcolor{gray}{$\pm$ 1.8}\rlap{$^{\ast\ast}$} \\
\midrule
\textit{Specialized Models} & & & & & & & \\
SFT \,\,\,\, w/o Stage 1 & 8B & 63.49 \textcolor{gray}{$\pm$ 0.8}\rlap{$^{\ast\ast}$} & 60.15 \textcolor{gray}{$\pm$ 0.5}\rlap{$^{\ast\ast}$} & \underline{83.78} \textcolor{gray}{$\pm$ 2.3} & 70.01 \textcolor{gray}{$\pm$ 1.0}\rlap{$^{\ast\ast}$} & 67.19 \textcolor{gray}{$\pm$ 1.4}\rlap{$^{\ast\ast}$} & 66.74 \textcolor{gray}{$\pm$ 1.4}\rlap{$^{\ast\ast}$} \\
GRPO w/o Stage 1 & 8B & 61.30 \textcolor{gray}{$\pm$ 1.8}\rlap{$^{\ast\ast}$} & 59.09 \textcolor{gray}{$\pm$ 1.3}\rlap{$^{\ast\ast}$} & 79.31 \textcolor{gray}{$\pm$ 2.4}\rlap{$^{\ast\ast}$} & 67.72 \textcolor{gray}{$\pm$ 1.6}\rlap{$^{\ast\ast}$} & 70.71 \textcolor{gray}{$\pm$ 1.7}\rlap{$^{\ast\ast}$} & 70.97 \textcolor{gray}{$\pm$ 1.3}\rlap{$^{\ast\ast}$} \\
SFT \,\,\,\, w/o Stage 2 & 8B & 73.43 \textcolor{gray}{$\pm$ 0.6}\rlap{$^{\ast\ast}$} & 70.56 \textcolor{gray}{$\pm$ 0.7}\rlap{$^{\ast\ast}$} & 82.06 \textcolor{gray}{$\pm$ 1.0}\rlap{$^{\ast\ast}$} & 75.87 \textcolor{gray}{$\pm$ 0.6}\rlap{$^{\ast\ast}$} & 75.72 \textcolor{gray}{$\pm$ 0.9}\rlap{$^{\ast\ast}$} & 69.56 \textcolor{gray}{$\pm$ 1.4}\rlap{$^{\ast\ast}$} \\
GRPO w/o Stage 2 & 8B & 74.31 \textcolor{gray}{$\pm$ 0.8}\rlap{$^{\ast\ast}$} & 70.11 \textcolor{gray}{$\pm$ 0.6}\rlap{$^{\ast\ast}$} & \textbf{86.55} \textcolor{gray}{$\pm$ 1.1}\rlap{$^{\ast\ast}$} & \underline{77.46} \textcolor{gray}{$\pm$ 0.7}\rlap{$^{\ast\ast}$} & 84.01 \textcolor{gray}{$\pm$ 0.6}\rlap{$^{\ast\ast}$} & \underline{84.05} \textcolor{gray}{$\pm$ 0.9}\rlap{$^{\ast\ast}$} \\
SFT $\rightarrow$ SFT & 8B & 73.86 \textcolor{gray}{$\pm$ 1.0}\rlap{$^{\ast\ast}$} & 72.46 \textcolor{gray}{$\pm$ 1.0}\rlap{$^{\ast\ast}$} & 78.49 \textcolor{gray}{$\pm$ 0.9}\rlap{$^{\ast\ast}$} & 75.35 \textcolor{gray}{$\pm$ 0.9}\rlap{$^{\ast\ast}$} & 74.30 \textcolor{gray}{$\pm$ 1.0}\rlap{$^{\ast\ast}$} & 69.41 \textcolor{gray}{$\pm$ 0.6}\rlap{$^{\ast\ast}$} \\
SFT $\rightarrow$ GRPO & 8B & \underline{76.02} \textcolor{gray}{$\pm$ 0.5}\rlap{$^{\ast\ast}$} & \underline{77.52} \textcolor{gray}{$\pm$ 0.9}\rlap{$^{\ast\ast}$} & 74.54 \textcolor{gray}{$\pm$ 0.7}\rlap{$^{\ast\ast}$} & 75.99 \textcolor{gray}{$\pm$ 0.4}\rlap{$^{\ast\ast}$} & \underline{84.04} \textcolor{gray}{$\pm$ 0.4}\rlap{$^{\ast\ast}$} & 83.89 \textcolor{gray}{$\pm$ 0.6}\rlap{$^{\ast\ast}$} \\
\textbf{Dementia-R1} & 8B & \textbf{77.39} \textcolor{gray}{$\pm$ 0.6} & \textbf{78.09} \textcolor{gray}{$\pm$ 0.7} & 77.72 \textcolor{gray}{$\pm$ 0.9} & \textbf{77.90} \textcolor{gray}{$\pm$ 0.6} & \textbf{85.29} \textcolor{gray}{$\pm$ 0.2} & \textbf{85.58} \textcolor{gray}{$\pm$ 1.2} \\
\bottomrule
\end{tabular}
}
\end{table*}

\subsection{Leave-One-Out Index Ablation (ADNI)}
\label{app:loo_ablation}

To quantify the contribution of each clinical index learned during Stage~1 pre-training, we perform a leave-one-out ablation on the ADNI dataset. At inference time ($T{=}0$), we remove one index from the input and evaluate the full Stage~2 dementia prognosis performance. Table~\ref{tab:loo_ablation} summarizes the results. Performance remained stable across all removals (F1: 0.74--0.77, AUROC: 0.83--0.85), demonstrating that the model does not over-rely on any single index. Among all indices, removing CDR-SB caused the sharpest decline (F1: 0.74, AUPRC: 0.80), identifying it as the most influential prognostic indicator. Notably, high AUROC was maintained even without memory-specific scales (RAVLT, LDELTOTAL), confirming strong cross-index redundancy.

\begin{table}[!h]
\centering
\small
\caption{\textbf{Leave-One-Out Index Ablation (ADNI, $T{=}0$).} Each row removes the specified index from the input at inference. CDR-SB removal causes the largest performance drop, while overall stability confirms cross-index redundancy.}
\label{tab:loo_ablation}
\adjustbox{max width=\columnwidth}{
\begin{tabular}{lcccccc}
\toprule
\textbf{Removed Index} & \textbf{Acc} & \textbf{Prec} & \textbf{Rec} & \textbf{F1} & \textbf{AUROC} & \textbf{AUPRC} \\
\midrule
MMSE & .79 & .74 & .80 & .77 & .85 & .83 \\
\textbf{CDRSB} & .76 & .70 & .78 & \textbf{.74} & .83 & \textbf{.80} \\
ADAS11 & .79 & .74 & .79 & .76 & .84 & .85 \\
ADAS13 & .79 & .77 & .73 & .75 & .84 & .83 \\
ADASQ4 & .78 & .75 & .74 & .74 & .84 & .85 \\
RAVLT & .80 & .76 & .78 & .77 & .85 & .83 \\
LDELTOTAL & .79 & .75 & .78 & .76 & .85 & .86 \\
\bottomrule
\end{tabular}
}
\end{table}

\subsection{Clinical Index Masking Ablation (AMC, Full Results)}
\label{app:masking_ablation}

Table~\ref{tab:masking_ablation_full} reports the full results of the clinical index masking ablation on the AMC cohort, including overall and bucket-wise performance breakdowns.

\begin{table*}[!t]
\centering
\small
\caption{\textbf{Clinical Index Masking Ablation (AMC).} Performance of Dementia-R1 when clinical indices are masked at inference time. ``None'' denotes the unmasked baseline. The largest drops are consistently associated with MMSE removal.}
\label{tab:masking_ablation_full}
\adjustbox{max width=\textwidth}{
\begin{tabular}{lcccccccc}
\toprule
& \multicolumn{2}{c}{\textbf{Overall}} & \multicolumn{2}{c}{\textbf{6--12m}} & \multicolumn{2}{c}{\textbf{12--18m}} & \multicolumn{2}{c}{\textbf{18--24m}} \\
\cmidrule(lr){2-3} \cmidrule(lr){4-5} \cmidrule(lr){6-7} \cmidrule(lr){8-9}
\textbf{Masked Index(es)} & \textbf{AUROC} & \textbf{AUPRC} & \textbf{AUROC} & \textbf{AUPRC} & \textbf{AUROC} & \textbf{AUPRC} & \textbf{AUROC} & \textbf{AUPRC} \\
\midrule
None (baseline) & \textbf{.8350} & \textbf{.8456} & \textbf{.8480} & \textbf{.8473} & \textbf{.8458} & \textbf{.8606} & \textbf{.7894} & .8641 \\
CDR & .8295 & .8371 & .8378 & .8309 & .8406 & .8614 & .7990 & \textbf{.8818} \\
GDS & .8326 & .8415 & .8380 & .8349 & .8492 & .8684 & .7981 & .8742 \\
GDS + CDR & .8295 & .8350 & .8391 & .8350 & .8424 & .8542 & .7808 & .8492 \\
\midrule
MMSE & .7941 & .8065 & .8065 & .8054 & .8160 & .8356 & .6731 & .7644 \\
MMSE + CDR & .7882 & .7783 & .8022 & .7772 & .8082 & .8061 & .7067 & .7659 \\
MMSE + GDS & .7829 & .7834 & .8056 & .7885 & .7903 & .7993 & .6923 & .7691 \\
ALL & .7807 & .7825 & .7965 & .7840 & .7941 & .8040 & .7135 & .7854 \\
\bottomrule
\end{tabular}
}
\end{table*}

\subsection{Temporal Density Ablation (Full Results)}
\label{app:density_ablation}

Table~\ref{tab:density_ablation_haeundae} reports the full results of the temporal density ablation on the HPH cohort, including overall and bucket-wise performance breakdowns.

\begin{table*}[!t]
\centering
\small
\caption{\textbf{Temporal Density Ablation (HPH).} Performance at 0\% (all indices masked) vs.\ 100\% (all indices retained). The model retains meaningful predictive power even without explicit clinical indices.}
\label{tab:density_ablation_haeundae}
\adjustbox{max width=\textwidth}{
\begin{tabular}{lcccccccccc}
\toprule
& \multicolumn{2}{c}{\textbf{Overall}} & \multicolumn{2}{c}{\textbf{6--12m}} & \multicolumn{2}{c}{\textbf{12--18m}} & \multicolumn{2}{c}{\textbf{18--24m}} & \multicolumn{2}{c}{\textbf{$\geq$24m}} \\
\cmidrule(lr){2-3} \cmidrule(lr){4-5} \cmidrule(lr){6-7} \cmidrule(lr){8-9} \cmidrule(lr){10-11}
\textbf{Density} & \textbf{AUROC} & \textbf{AUPRC} & \textbf{AUROC} & \textbf{AUPRC} & \textbf{AUROC} & \textbf{AUPRC} & \textbf{AUROC} & \textbf{AUPRC} & \textbf{AUROC} & \textbf{AUPRC} \\
\midrule
100\% & .7837 \textcolor{gray}{$\pm$ .0199} & .5103 \textcolor{gray}{$\pm$ .0306} & .7550 \textcolor{gray}{$\pm$ .0427} & .5618 \textcolor{gray}{$\pm$ .0515} & .8196 \textcolor{gray}{$\pm$ .0353} & .6135 \textcolor{gray}{$\pm$ .0387} & .9399 \textcolor{gray}{$\pm$ .0384} & .8476 \textcolor{gray}{$\pm$ .0744} & .6860 \textcolor{gray}{$\pm$ .0553} & .3579 \textcolor{gray}{$\pm$ .0770} \\
0\% & .7359 \textcolor{gray}{$\pm$ .0245} & .5465 \textcolor{gray}{$\pm$ .0207} & .7088 \textcolor{gray}{$\pm$ .0412} & .5365 \textcolor{gray}{$\pm$ .0213} & .7774 \textcolor{gray}{$\pm$ .0256} & .7113 \textcolor{gray}{$\pm$ .0244} & .8487 \textcolor{gray}{$\pm$ .1203} & .7381 \textcolor{gray}{$\pm$ .1070} & .6360 \textcolor{gray}{$\pm$ .0183} & .3658 \textcolor{gray}{$\pm$ .0664} \\
\bottomrule
\end{tabular}
}
\end{table*}

\subsection{Robustness under Naturally Sparse Clinical Indices}
\label{app:sparse_index}

The masking ablation (Figure~\ref{fig:masking_ablation}) removes clinical indices artificially at inference time. To complement it, we evaluate the naturally occurring sparse regime by stratifying the test sets according to how many MMSE/GDS/CDR mentions are actually present in the input history, and reporting performance on the \emph{zero-index} subset, in which the model must rely entirely on free-text narrative.

\paragraph{AMC.}
Of the 935 AMC test samples, 93 contain no clinical index mention at all. Table~\ref{tab:zero_index_amc} reports AUROC on this subset. Dementia-R1 leads all nine baselines, including the direct Stage~1 ablation (GRPO w/o Stage~1, $-2.66$) and every general-purpose model up to 72B.

\begin{table}[!h]
\centering
\small
\caption{\textbf{AMC Zero-Index Subset ($n{=}93$).} AUROC (mean $\pm$ std over 5 seeds) on test samples containing no MMSE/GDS/CDR mention.}
\label{tab:zero_index_amc}
\adjustbox{max width=\columnwidth}{
\begin{tabular}{lcc c}
\toprule
\textbf{Model} & \textbf{Size} & \textbf{AUROC} & \textbf{$\Delta$ vs.\ Ours} \\
\midrule
\textbf{Dementia-R1 (Ours)} & 7B & \textbf{72.24} $\pm$ 1.76 & --- \\
\midrule
GRPO w/o Stage~1 & 7B & 69.58 $\pm$ 2.07 & $-2.66$ \\
SFT~$\rightarrow$~GRPO & 7B & 64.40 $\pm$ 2.97 & $-7.84$ \\
SFT~$\rightarrow$~SFT & 7B & 61.82 $\pm$ 6.07 & $-10.42$ \\
\midrule
Qwen2.5-7B-Instruct & 7B & 66.97 $\pm$ 2.70 & $-5.27$ \\
Qwen2.5-32B-Instruct & 32B & 64.55 $\pm$ 2.71 & $-7.69$ \\
Qwen2.5-72B-Instruct & 72B & 63.60 $\pm$ 1.91 & $-8.64$ \\
Llama-3.3-70B-Instruct & 70B & 62.48 $\pm$ 1.97 & $-9.76$ \\
\midrule
HuatuoGPT-o1 (medical) & 8B & 65.27 $\pm$ 6.01 & $-6.97$ \\
MedGemma (medical) & 27B & 65.82 $\pm$ 3.41 & $-6.42$ \\
\bottomrule
\end{tabular}
}
\end{table}

\paragraph{HPH.}
HPH is naturally the sparsest cohort: 166 of its 190 test samples (87\%) contain no clinical index mention, making it the most direct real-world analogue of the extremely sparse setting. Table~\ref{tab:zero_index_hph} shows that Dementia-R1 again leads all nine baselines. The margin over the direct Stage~1 ablation widens to $-8.76$, indicating that Cold-Start pre-training is most valuable precisely when explicit indices are unavailable and the model must infer trajectory from narrative alone.

\begin{table}[!h]
\centering
\small
\caption{\textbf{HPH Zero-Index Subset ($n{=}166$).} AUROC (mean $\pm$ std over 5 seeds) on test samples containing no MMSE/GDS mention.}
\label{tab:zero_index_hph}
\adjustbox{max width=\columnwidth}{
\begin{tabular}{lcc c}
\toprule
\textbf{Model} & \textbf{Size} & \textbf{AUROC} & \textbf{$\Delta$ vs.\ Ours} \\
\midrule
\textbf{Dementia-R1 (Ours)} & 7B & \textbf{75.24} $\pm$ 2.62 & --- \\
\midrule
GRPO w/o Stage~1 & 7B & 66.48 $\pm$ 4.96 & $-8.76$ \\
SFT~$\rightarrow$~GRPO & 7B & 74.56 $\pm$ 3.08 & $-0.68$ \\
SFT~$\rightarrow$~SFT & 7B & 48.91 $\pm$ 3.10 & $-26.33$ \\
\midrule
Qwen2.5-7B-Instruct & 7B & 63.50 $\pm$ 3.01 & $-11.74$ \\
Qwen2.5-32B-Instruct & 32B & 63.62 $\pm$ 2.99 & $-11.62$ \\
Qwen2.5-72B-Instruct & 72B & 67.36 $\pm$ 1.29 & $-7.88$ \\
Llama-3.3-70B-Instruct & 70B & 65.74 $\pm$ 2.24 & $-9.50$ \\
\midrule
HuatuoGPT-o1 (medical) & 8B & 60.50 $\pm$ 4.19 & $-14.74$ \\
MedGemma (medical) & 27B & 69.70 $\pm$ 3.12 & $-5.54$ \\
\bottomrule
\end{tabular}
}
\end{table}

Taken together with the masking ablation, these results indicate that Dementia-R1 does not merely copy explicit clinical indices: it retains its advantage over every baseline in the sparsest regime on both cohorts.

\subsection{Robustness to Conflicting Longitudinal Signals}
\label{app:conflicting_notes}

Real clinical trajectories are frequently non-monotonic: a patient may cross the dementia threshold at one visit and fall back below it at the next. We isolate this regime explicitly.

\paragraph{Stratification.}
We stratify the 935 AMC test samples by whether the patient's GDS trajectory crosses the dementia cutoff (GDS $\geq$ 4). Excluding patients with fewer than two GDS readings ($n{=}231$), the remaining 704 samples split into \textbf{Stable Non-Dementia} ($n{=}331$), \textbf{Stable Dementia} ($n{=}164$), and \textbf{Crossing} ($n{=}209$), the last of which contains directly contradictory signals across visits. HPH cannot be stratified this way because 87.4\% of its patients lack repeated GDS readings. No model was retrained for this analysis.

\paragraph{Results on the crossing subset.}
Table~\ref{tab:crossing_amc} reports AUROC on the 209 crossing samples. Dementia-R1 leads all nine baselines, and the gap over the direct Stage~1 ablation ($-7.35$) is substantially wider than on the full test set, indicating that Cold-Start pre-training is what allows the model to weigh a transient fluctuation against the surrounding trajectory rather than reacting to the most recent value.

\begin{table}[!h]
\centering
\small
\caption{\textbf{AMC Crossing Subset ($n{=}209$).} AUROC (mean $\pm$ std over 5 seeds) on patients whose GDS trajectory crosses the dementia cutoff, i.e.\ whose visits carry contradictory signals.}
\label{tab:crossing_amc}
\adjustbox{max width=\columnwidth}{
\begin{tabular}{lcc c}
\toprule
\textbf{Model} & \textbf{Size} & \textbf{AUROC} & \textbf{$\Delta$ vs.\ Ours} \\
\midrule
\textbf{Dementia-R1 (Ours)} & 7B & \textbf{77.72} $\pm$ 1.11 & --- \\
\midrule
GRPO w/o Stage~1 & 7B & 70.37 $\pm$ 0.79 & $-7.35$ \\
SFT~$\rightarrow$~GRPO & 7B & 66.34 $\pm$ 4.29 & $-11.38$ \\
SFT~$\rightarrow$~SFT & 7B & 64.79 $\pm$ 3.58 & $-12.93$ \\
\midrule
Qwen2.5-7B-Instruct & 7B & 68.50 $\pm$ 4.22 & $-9.22$ \\
Qwen2.5-32B-Instruct & 32B & 71.53 $\pm$ 2.84 & $-6.19$ \\
Qwen2.5-72B-Instruct & 72B & 74.10 $\pm$ 1.93 & $-3.62$ \\
Llama-3.3-70B-Instruct & 70B & 71.79 $\pm$ 1.09 & $-5.93$ \\
\midrule
HuatuoGPT-o1 (medical) & 8B & 59.17 $\pm$ 1.70 & $-18.55$ \\
MedGemma (medical) & 27B & 56.24 $\pm$ 3.59 & $-21.48$ \\
\bottomrule
\end{tabular}
}
\end{table}

\paragraph{Case studies.}
To characterize \emph{how} the models differ, we examine three crossing patients (seed~1), comparing Dementia-R1 against Qwen2.5-32B-Instruct (the comparator used in the neurologist evaluation, Sec.~\ref{sec:human_eval}) and against GRPO w/o Stage~1 (which isolates the effect of Stage~1). We report the predicted probability $p$ of dementia.

\begin{itemize}
    \item \textbf{Case A} --- transient spike (gold $=0$; GDS $3 \rightarrow 4 \rightarrow 3$). Dementia-R1 correctly discounts the spike ($p{=}0.001$) by integrating spouse-reported memory improvement with a sustained MMSE, whereas Qwen2.5-32B ($p{=}1.000$) and GRPO w/o Stage~1 ($p{=}0.989$) both over-call on the single elevated reading.

    \item \textbf{Case B} --- alternation (gold $=0$; GDS alternating between 3 and 4). Dementia-R1 ($p{=}0.000$) anchors on a K-MMSE of 27, above the impairment threshold, and attributes the fluctuation to severe depression; both baselines over-call ($p \approx 1.000$).

    \item \textbf{Case C} --- apparent recovery (gold $=1$; GDS $3 \rightarrow 4 \rightarrow 3$, with a vascular dementia diagnosis). Dementia-R1 ($p{=}0.924$) integrates the vascular dementia diagnosis with the surrounding multimodal evidence, while GRPO w/o Stage~1 ($p{=}0.001$) produces a false negative by anchoring on an isolated ``MMSE 24, normal'' mention.
\end{itemize}

Across all three patterns --- spike, alternation, and apparent recovery --- Dementia-R1 integrates non-monotonic cues rather than following the most recent index value, and Case~C attributes this behavior specifically to Stage~1 Cold-Start pre-training.

\subsection{LLM Extraction Quality Audit}
\label{app:extraction_audit}

To validate the fidelity of the LLM-based extraction pipeline used for constructing verifiable rewards (Sec.~\ref{sec:3.1}), we conducted a human expert audit in two rounds.

\paragraph{Initial audit (200 notes / 600 slots).}
A clinician independently reviewed 200 randomly sampled clinical notes, verifying the extracted values for all three target indices (MMSE, GDS, CDR), yielding 600 slot-level comparisons. Table~\ref{tab:extraction_audit} summarizes the results. The overall slot-level accuracy was 98.5\%, confirming that the extracted rewards are high-fidelity and sufficient for robust RL optimization.

\begin{table}[!h]
\centering
\small
\caption{\textbf{LLM Extraction Audit Results (Initial).} Slot-level accuracy of the Qwen2.5-14B-Instruct extractor on 200 randomly sampled clinical notes (600 total slots).}
\label{tab:extraction_audit}
\begin{tabular}{lcc}
\toprule
\textbf{Clinical Index} & \textbf{Slots} & \textbf{Accuracy (\%)} \\
\midrule
MMSE & 200 & 97.5 \\
GDS & 200 & 98.5 \\
CDR & 200 & 99.5 \\
\midrule
\textbf{Overall} & 600 & 98.5 \\
\bottomrule
\end{tabular}
\end{table}

\paragraph{Expanded audit (600 notes).}
To verify that this estimate is not an artifact of the sample size, we expanded the audit to 600 clinical notes, auditing 200 independently sampled notes per clinical index. The aggregate slot-level accuracy was 98.17\% (589/600 correct, 11 errors), consistent with the initial estimate. Per-index results are reported in Table~\ref{tab:extraction_audit_expanded}.

\begin{table}[!h]
\centering
\small
\caption{\textbf{LLM Extraction Audit Results (Expanded).} Per-index accuracy of the Qwen2.5-14B-Instruct extractor over 600 audited clinical notes (200 per index).}
\label{tab:extraction_audit_expanded}
\adjustbox{max width=\columnwidth}{
\begin{tabular}{lcccc}
\toprule
\textbf{Clinical Index} & \textbf{Notes} & \textbf{Correct} & \textbf{Errors} & \textbf{Accuracy (\%)} \\
\midrule
MMSE & 200 & 194 & 6 & 97.00 \\
GDS & 200 & 196 & 4 & 98.00 \\
CDR & 200 & 199 & 1 & 99.50 \\
\midrule
\textbf{Total} & 600 & 589 & 11 & 98.17 \\
\bottomrule
\end{tabular}
}
\end{table}

\paragraph{Error type breakdown.}
We categorized the 11 errors observed in the expanded audit (Table~\ref{tab:extraction_error_types}). All are case-level and random rather than systematic; we observed no directional bias that could be amplified during RL training.

\begin{table}[!h]
\centering
\small
\caption{\textbf{Error Type Breakdown (Expanded Audit).} The 11 extraction errors are case-level random errors rather than a systematic, directional bias.}
\label{tab:extraction_error_types}
\begin{tabular}{p{0.30\columnwidth}cp{0.38\columnwidth}}
\toprule
\textbf{Error type} & \textbf{Count} & \textbf{Description} \\
\midrule
Fabrication & 3 & A value is extracted although only the index keyword appears in the note. \\
False negative & 4 & The value is missed because it falls outside the extraction boundary window. \\
Misattribution & 4 & An overloaded abbreviation is resolved to the wrong scale (e.g., ``GDS'' as Geriatric Depression Scale rather than Global Deterioration Scale). \\
\midrule
\textbf{Total} & \textbf{11} & \\
\bottomrule
\end{tabular}
\end{table}

\paragraph{Representative failure modes.}
From the initial audit we identified three representative failure modes among the nine misextracted slots:

\begin{enumerate}
    \item \textbf{Scale confusion.} The extractor misidentified a Barthel Index score as the GDS value when both appeared in close proximity within the ``Objective'' section (e.g., ``CDR 0.5 (3) / GDS 4 / Barthel Index 14/20 / S-IADL 11/45'').

    \item \textbf{Temporal misattribution.} When clinical indices and laboratory results from different dates co-occurred in the same note, the extractor assigned a value to the wrong visit date. For instance, ``2021/11/23 MMSE 12 CDR 0.5'' was erroneously attributed to a subsequent lab date (2022-04-05) listed immediately below.

    \item \textbf{Contextual misparse.} In notes where the ``MMSE :'' field was followed by detailed neurological examination subscores (e.g., ``phonemic 8 / 30 secs'') rather than a total score, the extractor incorrectly inferred MMSE = 30 from the denominator notation.
\end{enumerate}

These errors were rare (1.83\% of slots in the expanded audit) and predominantly affected MMSE extraction, which operates on the finest-grained scale. The tolerance-aware reward ($\delta = \pm 2$ for MMSE) further mitigates the impact of such boundary errors during RL training. Appendix~\ref{app:noise_robustness} additionally shows that the downstream prognosis is insensitive to extraction noise far exceeding this observed error rate.

\subsection{Robustness to Extraction Noise}
\label{app:noise_robustness}

Because Stage~1 rewards are produced by an auxiliary extractor, a natural concern is that extraction errors could cascade into and be amplified by RL training. We test this directly with a noise-injection stress test.

\paragraph{Setup.}
On a 1{,}000-sample subset, we corrupted the Stage~1 clinical indices for 30\%, 50\%, and 70\% of samples. Each injected error was placed \emph{above} the reward tolerance margin of the corresponding scale, so that the corrupted label is guaranteed to be treated as incorrect by $R_{\text{cold}}$ (Eq.~\ref{eq:reward_cold}). We then re-ran Stage~1 GRPO for each noise level and fine-tuned Stage~2 with identical hyperparameters.

\paragraph{Downstream performance is essentially unaffected.}
As shown in Table~\ref{tab:noise_downstream}, the final dementia AUROC stays within $\pm$0.5 percentage points of the noise-free baseline even when 70\% of the Stage~1 labels are corrupted beyond tolerance.

\begin{table}[!h]
\centering
\small
\caption{\textbf{Downstream Robustness to Stage~1 Extraction Noise.} Best dementia AUROC after injecting above-tolerance noise into the Stage~1 clinical indices.}
\label{tab:noise_downstream}
\begin{tabular}{lcc}
\toprule
\textbf{Injected noise} & \textbf{Best dementia AUROC} & \textbf{$\Delta$ vs.\ 0\%} \\
\midrule
0\% & 0.7945 & --- \\
30\% & 0.7926 & $-0.19$ pp \\
50\% & 0.7929 & $-0.16$ pp \\
70\% & 0.7902 & $-0.43$ pp \\
\bottomrule
\end{tabular}
\end{table}

\paragraph{Mechanism: noisy samples are filtered, not learned.}
GRPO standardizes rewards within each prompt's group of $G$ completions, $A_i = (r_i - \bar{r}) / \sigma_r$ (Eq.~\ref{eq:advantage}). A corrupted label is typically implausible given the patient's longitudinal record, so the model's prior steers most of the $G$ completions away from it. When \emph{all} $G$ completions miss the corrupted target, every reward is $0$, hence $r_i - \bar{r} = 0$ and $A_i = 0$ for all $i$: the gradient contribution of that prompt vanishes and the sample is effectively dropped rather than learned. Table~\ref{tab:noise_gradient} provides direct evidence: as the injected noise grows, the mean training reward falls while the fraction of auto-filtered (zero-gradient) prompts rises by roughly 13 percentage points.

\begin{table}[!h]
\centering
\small
\caption{\textbf{Automatic Filtering of Noisy Prompts under GRPO.} As noise increases, more prompts collapse to a zero advantage and contribute no gradient.}
\label{tab:noise_gradient}
\adjustbox{max width=\columnwidth}{
\begin{tabular}{lcc}
\toprule
\textbf{Injected noise} & \textbf{Mean training reward} & \textbf{Zero-gradient prompts} \\
\midrule
0\% & 0.268 & 0.712 \\
30\% & 0.198 & 0.770 \\
50\% & 0.152 & 0.823 \\
70\% & 0.116 & 0.840 \\
\bottomrule
\end{tabular}
}
\end{table}

\paragraph{Practical significance.}
The measured slot-level error rate of our extractor is 1.83\% (Appendix~\ref{app:extraction_audit}), roughly $38\times$ smaller than the 70\% stress level at which the downstream AUROC still degrades by less than half a point. Cascading extraction error is therefore negligible in practice.

\paragraph{Method versus extractor.}
A related question is how much of the reported gain comes from the method rather than from a strong extractor. The same Qwen2.5-14B-Instruct extractor supplies the clinical indices for \emph{all} specialized baselines in Tables~\ref{tab:main_results}--\ref{tab:adni_results} (SFT and GRPO w/o Stage~1, SFT~$\rightarrow$~SFT, SFT~$\rightarrow$~GRPO). With the extractor held constant, Dementia-R1 still outperforms the strongest 7B baseline (SFT~$\rightarrow$~GRPO) by $+6.27$ AUROC on AMC (84.02 vs.\ 77.75) and $+3.62$ on HPH (78.37 vs.\ 74.75), isolating the gain to the training framework rather than to the extraction pipeline.

\subsection{Training Implementation Details}
\label{app:implementation}

We implemented our framework using PyTorch. All experiments were conducted on four NVIDIA H100 (80GB) GPUs.

\paragraph{Reinforcement Learning (Dementia-R1)}
For the RL stage, we utilized the Open-R1 framework. We employed DeepSpeed ZeRO-3 and vLLM (colocate mode) to optimize memory usage for processing long clinical narratives. We set the per-device batch size to 1 with a gradient accumulation of 2 and a group size of $G=8$ (effective batch size of 8). The model was trained for 5,000 steps in Bfloat16 precision with a 2,000-token completion limit.

\subsection{Pretraining Target Ablation (Figures)}
\label{app:target_ablation_figures}

\begin{figure}[H]
    \centering
    \includegraphics[width=\columnwidth]{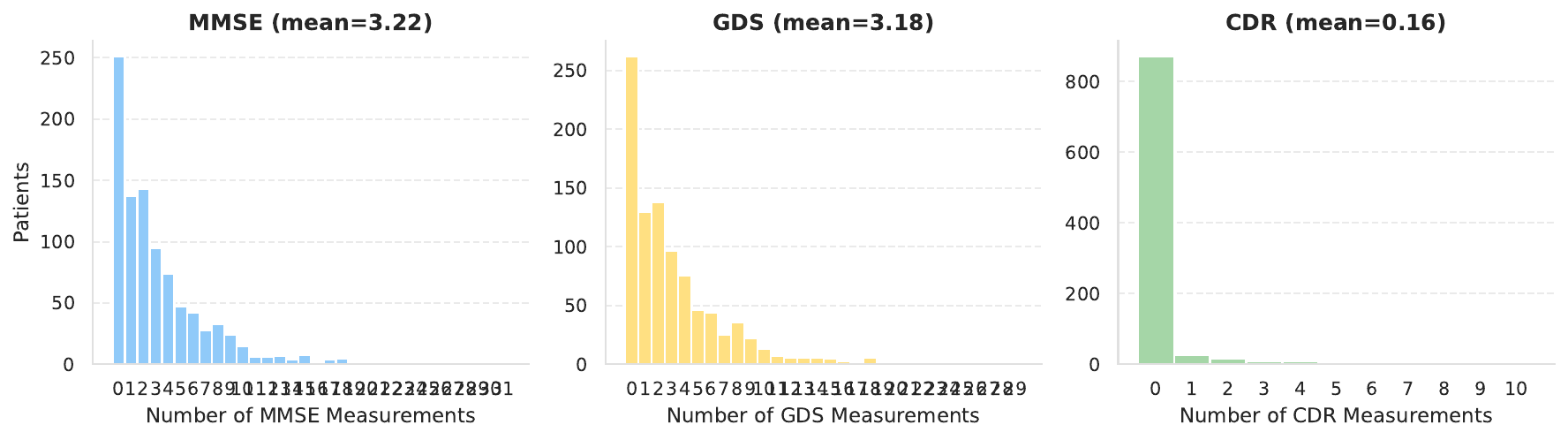}
    \vspace{-0.5cm}
    \caption{\textbf{Clinical Index Distribution (AMC Test Set).} Number of MMSE, GDS, and CDR measurements per patient. CDR annotations are substantially sparser than MMSE and GDS.}
    \label{fig:index_distribution}
    \vspace{-0.3cm}
\end{figure}

\begin{table}[H]
\centering
\small
\caption{\textbf{Pretraining Target Ablation.} Overall AUROC (\%) when Stage~1 uses individual vs.\ combined targets. Best per cohort in \textbf{bold}.}
\label{tab:target_ablation}
\begin{tabular}{lcc}
\toprule
\textbf{Target} & \textbf{AMC} & \textbf{HPH} \\
\midrule
All (combined) & 79.68 & 70.76 \\
MMSE & 78.82 & 69.95 \\
\textbf{GDS} & \textbf{81.03} & \textbf{70.88} \\
CDR & 78.88 & -- \\
\bottomrule
\end{tabular}
\end{table}

   

\subsection{Human evaluation protocol}
\label{app:human_eval}
To assess the clinical quality of reasoning, we conducted a blinded human evaluation with medical experts using a pairwise comparison protocol.
For each case, experts were presented with two anonymized model responses (Model~A and Model~B) generated for the same patient record and prediction task.
For each evaluation criterion, experts were asked to select one of three options: Model~A, Model~B, or Tie.

Each model pair was evaluated using 10 question-answer cases per comparison across two cohorts (AMC and HPH), yielding 100 total judgment pairs. Judgments were collected independently for the following five clinically motivated dimensions:

\begin{enumerate}
    \item \textbf{Temporal Reasoning Accuracy:} Which response appropriately interprets changes in symptoms and the rate of progression by comparing earlier records with the most recent records?
    \item \textbf{Evidence Grounding:} Which response cites evidence that is explicitly present in the original clinical notes and does not introduce information that is absent from the records?
    \item \textbf{Clinically Relevant Evidence Selection:} Which response avoids being influenced by clinically irrelevant details or overlooking key evidence, and instead bases its reasoning on diagnostically important evidence from the clinical notes?
    \item \textbf{Medical Soundness:} Which response is more medically sound with respect to dementia diagnostic criteria and clinical judgment, in terms of both the reasoning process and the final conclusion?
    \item \textbf{Overall Clinical Utility:} When used as reference material in real-world clinical practice, which response is more reliable and more helpful for reducing clinical decision-making time?
\end{enumerate}

\begin{figure*}[!t]
    \centering
    \begin{tcolorbox}[colback=gray!5!white, colframe=gray!75!black, title={Longitudinal Clinical Note Input (AMC)}, fonttitle=\bfseries]
    \scriptsize
2023-08-30:\\
Subjective\\
   Follow-up with Professor **, Constipation, pletaal dosage reduced
   Sometimes forgetful, but there are times when it's okay
   Caregiver observes that there is a slight decline
   Handles all household chores personally
   Forgets what they went for when crossing the room\\
Objective\\
   F/79y; Date of Birth (anonymized): ****/**/**\\
   2020/08/27 MMSE 23 GDS 4\\
   2022/02/03 MMSE 21 GDS 3\\
   2022/09/01 MMSE 17 GDS 4\\
   2023-08-30 MMSE; 17 (recall 2) GDS; 4\\
   Right-handed\\
   Highest Education Level; Illiterate\\
Assessment\\
    \# major neurocognitive disorder
   * VaD
   * MTA2/3 D3P3 (2022/09)\\
Plan\\
   Add ebixa and discontinue bearcept
   LICA\\
          Pletaal tab [50mg]                                1   TAB   DP     1 time    35 days   PO  \\
          Lexapro tab [10mg]                                1   TAB   N     1 time    35 days   PO  \\
          Ebixa tab [10mg]                                  0.5   TAB  BNP     2 times    35 days   PO\\
2023-10-04:\\
Subjective\\
   Follow-up patient of Professor \\
   Scheduled for LICA after dementia team consultation on 2023/08/30\\
   Discontinued bearcept and added ebixa\\
   Bowel movements improved after changing the medication\\
Objective\\
   F/79y; Date of Birth (anonymized): ****/**/**\\
   2020/08/27 MMSE 23 GDS 4\\
   2022/02/03 MMSE 21 GDS 3\\
   2022/09/01 MMSE 17 GDS 4\\
   2023/08/30 MMSE 17 (recall 2) GDS 4 (illiterate)\\
   2022-08-21 eGFR(CKD-EPI) (Qn), Blood 69 ml/min/1.73m$^2$\\
Assessment\\
   \# major neurocognitive disorder\\
   * VaD\\
   * MTA2/3 D3P3 (2022/08)\\
Plan\\
   * Reduced pletaal due to incontinence (outpatient of Professor **)\\
   LICA on 07/07\\
   Increase ebixa dosage, reduce back to half tablet if side effects occur\\
          Pletaal tab [50mg]                                1   TAB   DP     1 time   56 days   PO\\
          Lexapro tab [10mg]                                1   TAB   N     1 time   56 days   PO\\
          Ebixa tab [10mg]                                  1   TAB   BNP   2 times   56 days   PO\\
2023-12-02:\\
Subjective\\
   Post LICA visit\\
   No gastrointestinal side effects with current medication\\
   mood: so so\\
Objective\\
   F/79y; Date of Birth (anonymized): ****/**/**\\
   Unlearned\\
   2020/08/27 MMSE 23 GDS 4\\
   2022/02/03 MMSE 21 GDS 3\\
   2022/09/01 MMSE 17 GDS 4\\
   2023/08/30 MMSE 17 GDS 4 (recall 2)\\
   2023/11/24                    GDS 3 CDR 0.5 SB 1.0 BI 20 SIADL 5 NPI 2\\
   Note> Z score -1.5 or lower in some cognitive domains.
   The test results suggest a retrieval deficit in verbal memory and a deficit in visual memory. Other functions such as frontal/executive functions, attention, language and related functions, and visuoconstruction ability are all within normal levels.
   Therefore, bilateral frontal lobe dysfunction is suggested.
   The patient does not report any decline in ADL and is currently in an amnestic mild cognitive impairment state.
   The patient has shown cognitive decline since around '18, which has gradually worsened, and with multiple vascular risk factors and test results indicating a retrieval deficit, subcortical vascular cognitive impairment needs to be considered first. Given the patient's advanced age and the more pronounced memory impairment compared to other cognitive domains, Alzheimer's disease underlying pathology or a combined condition should also be considered, and the influence of hypothyroidism cannot be completely ruled out.
   A correlation with neuroimaging studies is recommended for an accurate diagnosis, and follow-up is recommended if cognitive function decline worsens. Treatment for depression is also recommended.\\
   2022-08-21        eGFR(CKD-EPI) (Qn), Blood  69  ml/min/1.73m$^2$  \\
Assessment\\
   \# impending dementia\\
   * VaD
   * MTA2/3 D3P3 (2022/08)
Plan\\
   * Reduce pletaal due to fecal incontinence (outpatient visit with Professor **)\\
   ditto\\
   Continued lexapro after consultation\\
   Consultation to change medication if fecal incontinence persists\\
          Pletaal tab [50mg]                                1   TAB   DP     1 time    91 days   PO  \\
          Lexapro tab [10mg]                                1   TAB    N     1 time    91 days   PO  \\
          Ebixa tab [10mg]                                  1   TAB  BNP     2 times    91 days   PO\\
    \end{tcolorbox}
    \caption{Example of longitudinal clinical notes. To protect patient privacy, all personal identifiers (e.g., names) were removed, and dates were randomly shifted.}
    \label{fig:note_example}
\end{figure*}

\begin{figure*}[!t]
    \centering
    \begin{tcolorbox}[colback=gray!5!white, colframe=gray!75!black, title={Qualitative Comparison of Longitudinal Reasoning for 6-12 Month Dementia Prognosis}, fonttitle=\bfseries]
    \small
    \textbf{[Dementia-R1 output]} \\
The patient's clinical notes and cognitive assessments provide a clear picture of their cognitive decline over time. Here's the step-by-step reasoning:\\
\\
1. **Cognitive Assessments Over Time:**\\
   - 2020/08/27: MMSE 23, GDS 4 (Severe cognitive decline)\\
   - 2022/02/03: MMSE 21, GDS 3 (Moderate cognitive decline)\\
   - 2022/09/01: MMSE 17, GDS 4 (Severe cognitive decline)\\
   - 2023/08/30: MMSE 17 (recall 2), GDS 4 (Severe cognitive decline)\\
\\
2. **Clinical Dementia Rating (CDR):**\\
   - The CDR scores are consistently 0.5 or 1, indicating very mild to mild dementia.\\
\\
3. **Symptoms and Functional Decline:**\\
   - The patient reports forgetfulness, sometimes forgetting what they went for when crossing the room.\\
   - Caregiver observes a slight decline.\\
   - The patient handles all household chores personally but shows signs of memory impairment.\\
\\
4. **Diagnosis and Plan:**\\
   - The patient is diagnosed with VaD (Vascular Dementia) and MTA2/3 D3P3.\\
   - The patient is currently in an amnestic mild cognitive impairment state, with a retrieval deficit in verbal and visual memory.\\
   - The patient has shown cognitive decline since around 2019, which has gradually worsened.\\
\\
5. **Current Status:**\\
   - The most recent MMSE score (17) and GDS score (4) indicate severe cognitive decline.\\
   - The patient's CDR score is 0.5 or 1, indicating very mild to mild dementia.\\
   - The patient's condition has not improved over the past year, and the decline is consistent with a diagnosis of dementia.\\
\\
Given the consistent cognitive decline, severe MMSE and GDS scores, and the diagnosis of VaD, it is clear that the patient is likely to have dementia.\\
    \vspace{0.8em}
    \hrule
    \vspace{0.8em}

    \textbf{[QWEN 2.5 32B  output]} \\
1. The patient's MMSE scores have been declining over time: 23 in 2020, 21 in 2022, and 17 in 2022 and 2023. A score of 17 is indicative of mild to moderate cognitive impairment.\\
2. The patient has been diagnosed with major neurocognitive disorder, specifically vascular dementia (VaD), and is classified as MTA2/3 D3P3.\\
3. The caregiver observes a slight decline in the patient's cognitive function, and the patient sometimes forgets what they went for when crossing the room, which is a sign of cognitive impairment.\\
4. The patient's GDS score has been consistently around 3-4, indicating mild depression, which can be associated with cognitive decline.\\
5. The patient's cognitive test results suggest a retrieval deficit in verbal memory and a deficit in visual memory, with other cognitive functions within normal levels.\\
6. The patient is in an amnestic mild cognitive impairment state, with cognitive decline since around 2019, which has gradually worsened.\\
7. The patient has multiple vascular risk factors, and the test results indicate a retrieval deficit, suggesting subcortical vascular cognitive impairment.\\
8. The patient's advanced age and more pronounced memory impairment compared to other cognitive domains suggest Alzheimer's disease underlying pathology or a combined condition.\\
9. The patient is being treated with Ebixa (donepezil), which is commonly used in the treatment of Alzheimer's disease and other forms of dementia
    \end{tcolorbox}
    \caption{A comparison of the reasoning processes generated by Dementia-R1 and the Qwen2.5-32B for the clinical case shown in Figure~\ref{fig:note_example}. While Dementia-R1 produces a structured and grounded rationale, the baseline generates a dense narrative containing medical hallucinations.}
    \label{fig:answer_example}
\end{figure*}

\begin{figure*}[!t]
    \centering
    \begin{tcolorbox}[colback=gray!5!white, colframe=gray!75!black, title={Longitudinal Clinical Note Input (HPH)}, fonttitle=\bfseries]
    \scriptsize
\textbf{=== Initial notes ===}\\
2019-11-05: Department[Neurology] BP(S)[*] BP(D)[86] Hb[15.3] PLT[254] AST[28] ALT[27] Glucose[81] Chief complaint[Left leg tremor onset 2 months ago] Pain presence[None] Current disease status[left finger tapping: decreased left foot tapping: decreased left side rigidity left resting tremor on foot Parkinson's dz Tremor dominant HY 1] Medical history[Hypertension] Medical history[Diabetes] Medical history[Hyperlipidemia] Diagnosis[Parkinson's disease] Treatment plan[FP CIT PET]\\[0.3em]
\textbf{=== Follow-up notes ===}\\
2019-11-20: Department[Neurology] S[] O[Explained FP CIT PET scan] A[Parkinson's disease Tremor dominant HY 1] P[repeat initial Parkinson's disease]\\
2019-12-04: S[Feeling nauseous after taking medication.] O[150/81 finger tapping: intact gait: normal -- ``on'' state] A[Parkinson's disease HY stage 1 S/E madopar - vomiting] P[repeat]\\
2020-01-15: S[Body shakes 2--3 times a day.] O[136/89 bradykinesia (+) gait: normal] P[repeat next follow-up to confirm tremor]\\
2020-03-11: S[Shaking intermittently.] O[mild finger tapping decreased overall bradykinesia 123/75] A[Parkinson's disease HY stage 1 S/E madopar - vomiting] P[tepra increase trihexine increase]\\
2020-06-03: S[Shaking in legs due to lack of movement.] O[135/75 left foot tremor at rest, finger tapping: equivocal shuffling gait] A[Parkinson's disease HY stage 1 S/E madopar - vomiting] P[Increase madopar 100-50-100 trihexine increase explained side effects]\\
2020-07-01: S[Body feels a little less stiff, right leg shakes] O[127/72 right foot tremor at rest (+) gait: normal] A[Parkinson's disease HY stage 1] P[repeat]\\
2020-08-26: S[Leg shakes, dizzy.] O[128/69 bradykinesia (-) right foot tremor - no change] A[Parkinson's disease HY stage 1] P[tepra $\rightarrow$ indenol change]\\
2020-11-16: S[Shaking in arms and legs is difficult. Intermittent headaches. Feeling suicidal.] O[116/70] A[Parkinson's disease HY stage 1] P[trihexine additional ssri additional increase madopar if necessary psychiatry consultation]\\
2020-12-07: S[Shaking intermittently. Memory is very poor.] O[126/74 Bradykinesia (-/-) tremor (-)] A[Parkinson's disease HY stage 1] P[trihexine decrease psychiatry consultation discontinue depression medication]\\
2021-02-01: S[Memory decline is severe. Crying a lot, too scared to see psychiatry, mmse=28] O[119/64] P[psychiatry consultation absolutely necessary]\\
2021-03-09: S[Psychiatry consultation made me feel a little better.] O[left side bradykinesia (+) 132/68] A[Parkinson's disease HY stage 1] P[add azilect]\\
2021-05-06: S[medication only] O[140/78 bradykinesia (+) resting tremor on left hand] A[Parkinson's disease HY stage 1] P[prescribe amantadine]\\
2021-06-28: S[feeling tired and having difficulty working in the restaurant.] O[122/63 mild bradykinesia gait: mild slow] A[Parkinson's disease HY stage 1 S/E xanax - somnolence] P[discontinue xanax]\\
2021-08-23: S[feeling sleepy.] O[131/76] A[Parkinson's disease HY stage 1] P[repeat]\\
2021-12-07: A[Parkinson's disease HY stage 1] P[diagnosis letter]\\
2022-02-07: S[feeling tremors.] O[140/70 left side bradykinesia (+)] A[Parkinson's disease HY stage 1] P[add trihexyphenidyl]\\
2022-04-04: S[significant memory decline. Short-term memory decline] O[137/71] A[Parkinson's disease HY stage 1] P[discontinue trihexyphenidyl (memory decline) mmse]\\
2022-05-30: S[back pain, leg pain. Left leg pain - orthopedic] O[164/90] A[Parkinson's disease HY stage 1] P[adjust to bid, follow up with orthopedic]\\
2022-09-26: S[back pain fluctuates, antalgic gait, memory decline affecting work] O[139/81] A[Parkinson's disease HY stage 1] P[repeat, check for dementia symptoms]\\
2023-02-20: S[severe depression, significant blood pressure fluctuations] O[132/61] P[add trihexyphenidyl, add lexapro, ANS function test at next follow-up]\\
2023-05-08: S[feeling tremors, having difficult thoughts due to depression] P[increase lexapro, increase trihexyphenidyl]\\
2023-06-26: S[Feels like blood is gushing in the head. Sensory symptom on the right parietal area. Feels heavy. Poor memory.] O[110/63] A[Parkinson's disease HY stage 1] P[REPEAT New sensory symptom BRAIN MRI]\\
2023-08-21: S[Feels like head is flowing. Lower back pain. Dragging the feet. Memory decline.] O[mri results explained] A[Parkinson's disease HY stage 1] P[Etivil Buspar prescribed MMSE prescribed]\\[0.3em]
\textbf{=== Autonomic nervous system tests ===}\\
2019-11-28: Blood pressure fluctuation [Abnormal] Valsalva ratio [Abnormal] Conclusion [Adrenergic and cardiovagal dysfunction. No orthostatic hypotension. CASS Cardiovagal score - 1 Adrenergic score - 2] Head-up tilt table test [Normal HUT]\\
2023-05-08: Blood pressure fluctuation [Abnormal] Valsalva ratio [Abnormal] Conclusion [Adrenergic and cardiovagal dysfunction. No orthostatic hypotension.] Head-up tilt table test [Normal HUT]\\[0.3em]
\textbf{=== K-MMSE records ===}\\
2019-11-28: 27 \quad 2021-02-01: 28 \quad 2022-05-30: 29\\[0.3em]
\textbf{=== GDS records ===}\\
2019-11-28: 2 \quad 2021-02-01: 2 \quad 2022-05-30: 2\\[0.3em]
\textbf{=== Brain MRI ===}\\
2019-11-20: Absence of septum pellucidum. Probable small vessel disease.\\[0.3em]
\textbf{=== PER-CT ===}\\
2019-11-05: Decreased FP-CIT uptake in dorsal posterior putamen (Rt.$\ll$Lt.), probable Parkinson's disease.\\[0.3em]
\textbf{=== Gait analysis ===}\\
2019-11-20: The Cadence is within normal range. The Double Support is within normal range. The Single Support is within normal range. The Step Length is more symmetrical than average. The Step Time is within normal range. The Stride Length is within normal range. The Stride Time is more symmetrical than average. The Walking Speed is within normal range.\\[0.3em]
\textbf{=== Motor symptom description ===}\\
Speech is normal. Facial expression is slightly impaired. Rest tremor is normal in the face and chin. Rest tremor is normal in the right arm. Rest tremor is normal in the left arm. Rest tremor is normal in the right leg. Rest tremor is slightly impaired in the left leg. Action tremor is normal in the right arm. Action tremor is normal in the left arm. Stiffness is normal in the neck. Stiffness is normal in the right arm. Stiffness is normal in the left arm. Stiffness is normal in the right leg. Stiffness is normal in the left leg. Right finger tapping is normal. Left finger tapping is slightly impaired. Right hand movement is normal. Left hand movement is normal. Rapid right hand movement is normal. Rapid left hand movement is mildly impaired. Left foot tapping is slightly impaired. Right foot tapping is normal. Rising from chair is normal. Standing ability is slightly impaired. Gait is normal. Postural stability is slightly impaired. Movement speed is normal.
    \end{tcolorbox}
    \caption{Example of longitudinal clinical notes from the HPH cohort for PDD prediction. Unlike the AMC cohort's SOAP-format notes, HPH records consist of structured fields (S/O/A/P) combined with separate clinical index records, autonomic tests, neuroimaging, gait analysis, and motor symptom descriptions. To protect patient privacy, all personal identifiers were removed, and dates were randomly shifted.}
    \label{fig:hph_note_example}
\end{figure*}

\begin{figure*}[!t]
    \centering
    \begin{tcolorbox}[colback=gray!5!white, colframe=gray!75!black, title={Qualitative Comparison of Longitudinal Reasoning for 6--12 Month PDD Prognosis (HPH)}, fonttitle=\bfseries]
    \small
    \textbf{[Dementia-R1 output]} \\
1. Cognitive Assessments:\\
\quad - MMSE Score: 2019-11-28: 27; 2021-02-01: 28; 2022-05-30: 29\\
\quad - GDS Score: 2019-11-28: 2; 2021-02-01: 2; 2022-05-30: 2\\
\\
2. Cognitive Decline Indicators:\\
\quad - The MMSE scores are relatively stable and within the normal range (27--29), which suggests no significant cognitive decline.\\
\quad - The GDS scores are consistently at 2, indicating very mild cognitive decline.\\
\\
3. Functional Assessments:\\
\quad - Gait Analysis: The gait analysis shows that the patient's gait is normal, and there are no significant impairments noted.\\
\quad - Motor Symptoms: The patient's motor symptoms are generally stable, with some minor impairments noted, primarily in the left leg and hand.\\
\\
4. Other Clinical Indicators:\\
\quad - The patient has experienced some memory decline, but the memory decline is not severe, and the patient's overall cognitive function remains within a normal range.\\
\quad - The patient has also experienced some psychological issues, including depression and suicidal thoughts, which could be contributing to the perception of cognitive decline.\\
\\
5. Conclusion: Given the stable MMSE scores (27--29) and the consistent GDS scores (2), the patient does not show significant cognitive decline. The minor motor and gait impairments are not severe enough to suggest a diagnosis of PDD at this time.\\
\\
\textbf{Answer: 0}
    \vspace{0.8em}
    \hrule
    \vspace{0.8em}

    \textbf{[QWEN 2.5 32B output]} \\
The patient has a history of Parkinson's disease (PD) and has been followed over several years. The notes indicate that the patient has experienced tremors, bradykinesia, and other motor symptoms typical of PD. Additionally, the patient has reported memory decline and cognitive changes.\\
\\
1. Cognitive Assessments: MMSE scores have been relatively stable over time: 27 in 2019, 28 in 2021, and 29 in 2022. These scores are within the range of mild cognitive impairment but not severely impaired. GDS scores have consistently been 2, indicating very mild cognitive decline.\\
\\
2. Motor Symptoms: The patient has had tremors and bradykinesia. The HY stage has remained at 1, indicating mild motor impairment. Gait analysis shows normal walking speed and cadence.\\
\\
3. ANS Tests: The ANS tests have shown adrenergic and cardiovagal dysfunction, which can be associated with PD but are not specific indicators of dementia.\\
\\
4. Recent Notes: Recent notes (e.g., 2023) indicate significant memory decline and difficulty with work. However, the MMSE score of 29 and GDS score of 2 still indicate only very mild cognitive impairment.\\
\\
Given that the cognitive assessments have not shown severe cognitive impairment, the patient is unlikely to have developed PDD yet. However, the recent memory decline and cognitive symptoms warrant further monitoring.\\
\\
\textbf{Answer: 0}
    \end{tcolorbox}
    \caption{A comparison of the reasoning processes generated by Dementia-R1 and Qwen2.5-32B for the HPH clinical case shown in Figure~\ref{fig:hph_note_example}. Both models correctly predict non-progression to PDD, but Dementia-R1 provides a more structured analysis that systematically separates cognitive, functional, and motor assessments, while the baseline produces a less organized narrative.}
    \label{fig:hph_answer_example}
\end{figure*}

\clearpage






\subsection{Prompt Templates}
\label{app:prompts}

\paragraph{Stage 1 (Cold-Start Pre-training).}
Figures~\ref{fig:prompt_asan_pretrain} and~\ref{fig:prompt_adni_pretrain} present the prompt templates for the AMC and ADNI pre-training tasks, respectively. In this stage, the model is trained to predict verifiable intermediate clinical indices (e.g., MMSE/GDS/CDR) extracted from unstructured notes or structured records. All templates enforce a unified \texttt{<think>} / \texttt{<answer>} format, enabling reliable parsing of the predicted value for training.

\paragraph{Stage 2 (Task Fine-tuning).}
Figures~\ref{fig:prompt_asan}, \ref{fig:prompt_haeundae}, and~\ref{fig:prompt_adni} present the prompt templates for the cohort-specific downstream tasks of dementia prognosis on AMC, Parkinson's disease dementia (PDD) prediction on HPH, and MCI-to-dementia conversion prediction on ADNI. These templates retain the same constrained output format as Stage~1 and guide the model to base its prediction on longitudinal evidence across the provided history.

\paragraph{Other prompts and reuse across settings.}
Figure~\ref{fig:prompt_rationale_gen} shows a separate prompt used to generate teacher rationales for constructing CoT-supervised data for SFT baselines. We reuse the same task prompts across all experimental pipelines, including SFT$\rightarrow$SFT, SFT$\rightarrow$GRPO, GRPO$\rightarrow$GRPO, and single-stage baselines. The pipelines differ only in the optimization procedure and in whether teacher-generated rationales are included.



\begin{figure*}[!t]
    \centering
    \begin{tcolorbox}[colback=gray!5!white, colframe=gray!75!black, title={Pre-training Task: AMC (Future Score Prediction)}, fonttitle=\bfseries]
    \small
    \textbf{[System Prompt]} \\
    A conversation between a User and an Assistant. The User provides clinical notes and metadata describing a patient's condition before a cognitive or functional assessment. Each input explicitly specifies the target scale (GDS, MMSE, or CDR), the time interval, and the required output format. The Assistant must carefully read the provided instructions, understand which scale is being predicted, and output the correct numerical value according to the described scoring rule.

    \vspace{0.8em}
    \hrule
    \vspace{0.8em}
    
    \textbf{[Input Data]} \\
    \textbf{Instruction:} You are given longitudinal clinical notes recorded BEFORE a cutoff relative to a cognitive assessment. The most recent included note lies \texttt{<TIME\_INTERVAL>} prior to the anchor assessment date. \\
    \textbf{Task:} Predict the target score (\textbf{Example: MMSE}) for the anchor assessment. \\
    \textbf{Format:} Output step-by-step reasoning in \texttt{<think>} tags and the final value in \texttt{\textbackslash boxed\{\}} within \texttt{<answer>} tags.

    \vspace{0.5em}
    \textbf{Scoring Indicators Glossary:} \\
    - \texttt{MMSE}: Integer score ranging from 0 to 30 (Higher = better global cognition). \\
    - \texttt{GDS}: Global Deterioration Scale from 1 to 7 (Higher = more severe impairment). \\
    - \texttt{CDR}: Clinical Dementia Rating global score chosen from \{0, 0.5, 1, 2, 3\}.
    
    \vspace{0.5em}
    === Clinical note === \\
    \texttt{<CLINICAL\_NOTE>}
    \end{tcolorbox}
    \caption{Pre-training prompt template for the AMC dataset. While \textbf{MMSE} is shown as an example, the model is pre-trained to predict various global cognitive scores, including \textbf{GDS} and \textbf{CDR}, based on unstructured clinical notes.}
    \label{fig:prompt_asan_pretrain}
\end{figure*}

\begin{figure*}[!t]
    \centering
    \begin{tcolorbox}[colback=gray!5!white, colframe=gray!75!black, title={Pre-training Task: ADNI (Future Score Prediction)}, fonttitle=\bfseries]
    \small
    \textbf{[System Prompt]} \\
    A conversation between a User and an Assistant. The User provides longitudinal structured ADNI clinical, cognitive, imaging, and biomarker data across multiple visits. The Assistant must predict the future score or diagnosis at the NEXT visit within a specified time window. Target tasks include \texttt{MMSE}, \texttt{CDRSB}, \texttt{ADAS11}, \texttt{ADAS13}, \texttt{ADASQ4}, \texttt{RAVLT\_learning}, and \texttt{LDELTOTAL}. Respond only in the specified \texttt{<think>} and \texttt{<answer>} format.

    \vspace{0.8em}
    \hrule
    \vspace{0.8em}
    
    \textbf{[Input Data]} \\
    \textbf{Instruction:} You are given longitudinal records for a single participant. All visits occur before the target visit. \\
    \textbf{Task:} Predict the target score (\textbf{Example: MMSE}) at the NEXT visit. \\
    \textbf{Constraint:} Time gap bucket = 2--3 months. \\
    \textbf{Format:} Output step-by-step reasoning in \texttt{<think>} tags and the final predicted value in \texttt{\textbackslash boxed\{\}} within \texttt{<answer>} tags.

    \vspace{0.5em}
    \textbf{Variable Glossary:} \\
    - \texttt{PTEDUCAT/APOE4}: Education years / Number of APOE $\epsilon$4 alleles. \\
    - \texttt{CDRSB/ADAS13/MMSE/MOCA}: Clinical severity and cognitive scores (Higher CDRSB/ADAS = worse; Higher MMSE/MOCA = better). \\
    - \texttt{RAVLT/LDELTOTAL}: Memory scores (Lower = poorer memory). \\
    - \texttt{FAQ}: Functional Activities Questionnaire (Higher = worse daily function). \\
    - \texttt{ABETA/TAU/PTAU}: CSF biomarkers for amyloid and tau pathology. \\
    - \texttt{Ventricles/Hippocampus/WholeBrain}: MRI volumetric measures (Structural atrophy).
    
    \vspace{0.5em}
    === Clinical Assessment Data === \\
    \texttt{2006-12-11: <<<VISIT 1/2>>>} \\
    \texttt{ABETA: 446.8, ADAS13: 25.0, MMSE: 27, CDRSB: 0.5, LDELTOTAL: 12, ...} \\
    \texttt{-----(Longitudinal history continues)-----} \\
    \texttt{(Prediction target: MMSE score at the next visit)}
    \end{tcolorbox}
    \caption{Pre-training prompt template for the ADNI dataset. The model predicts future indicators (e.g., MMSE, CDRSB) by analyzing longitudinal structured assessment data. The input includes a variable glossary to assist in interpreting clinical indicators.}
    \label{fig:prompt_adni_pretrain}
\end{figure*}
\begin{figure*}[!t]
    \centering
    \begin{tcolorbox}[colback=gray!5!white, colframe=gray!75!black, title={Fine-tuning Task: AMC (Dementia Prognosis)}, fonttitle=\bfseries]
    \small
    \textbf{[System Prompt]} \\
    A conversation between a User and an Assistant. The User provides longitudinal clinical notes and metadata describing a patient's condition. The Assistant must determine whether the patient is likely to be diagnosed with dementia. Output 0 if the patient is unlikely to have dementia, and 1 if the patient is likely to have dementia. Respond only in the specified \texttt{<think>} and \texttt{<answer>} format.

    \vspace{0.8em}
    \hrule
    \vspace{0.8em}
    
    \textbf{[Input Data]} \\
    \textbf{Instruction:} You are given longitudinal clinical notes collected BEFORE a cutoff relative to a dementia diagnosis date. The interval to the diagnosis date is: \texttt{<TIME\_INTERVAL>} (e.g., 12--18m). \\
    \textbf{Task:} Predict whether the patient is likely to have dementia (0: unlikely, 1: likely). \\
    \textbf{Format:} Output step-by-step reasoning in \texttt{<think>} tags and the final answer in \texttt{\textbackslash boxed\{0 or 1\}} within \texttt{<answer>} tags.

    \vspace{0.5em}
    \textbf{Scoring Indicators Glossary:} \\
    - \texttt{CDR (Global Score)}: 0 (No dementia), 0.5 (Very mild), 1 (Mild), 2 (Moderate), 3 (Severe). Higher = worse. \\
    - \texttt{MMSE (Total Score)}: Integer from 0 to 30. Higher = better cognitive function; Lower = more impairment. \\
    - \texttt{GDS (Global Deterioration Scale)}: 1 (No decline) to 7 (Very severe cognitive decline). Higher = worse.
    
    \vspace{0.5em}
    === Clinical note === \\
    \texttt{<CLINICAL\_NOTE>}
    \end{tcolorbox}
    \caption{Fine-tuning prompt template for the AMC dataset. The task requires detecting dementia presence based on unstructured clinical notes and integrated scoring indicators.}
    \label{fig:prompt_asan}
\end{figure*}

\begin{figure*}[!t]
    \centering
    \begin{tcolorbox}[colback=gray!5!white, colframe=gray!75!black, title={Fine-tuning Task: HPH (PDD Prediction)}, fonttitle=\bfseries]
    \small
    \textbf{[System Prompt]} \\
    A conversation between a User and an Assistant. The User provides longitudinal clinical notes and metadata describing a patient's condition. The clinical notes may contain standardized cognitive and functional assessment information. When present, these indicators should be interpreted as follows:

    \vspace{0.3em}
    \textbf{Scoring Indicators Glossary:} \\
    - \texttt{MMSE (Total Score)}: Integer from 0 to 30. Higher = better cognitive function; Lower = more impairment. \\
    - \texttt{GDS (Global Deterioration Scale)}: 1 (No decline) to 7 (Very severe cognitive decline). Higher = worse.

    \vspace{0.3em}
    The Assistant must determine whether the patient is likely to be diagnosed with dementia. Output 0 if the patient is unlikely to have dementia, and 1 if the patient is likely to have dementia. Respond only in the specified \texttt{<think>} and \texttt{<answer>} format.

    \vspace{0.8em}
    \hrule
    \vspace{0.8em}

    \textbf{[Input Data]} \\
    \textbf{Instruction:} You are given longitudinal clinical notes collected BEFORE a cutoff relative to the patient's last clinical assessment. The interval between the last assessment date and the most recent available note is: \texttt{<TIME\_INTERVAL>} (e.g., $\geq$24m). \\
    \textbf{Task:} Predict whether the patient diagnosed with Parkinson's disease will develop Parkinson's disease dementia (PDD) (0: unlikely, 1: likely). \\
    \textbf{Format:} Output step-by-step reasoning in \texttt{<think>} tags, justifying the decision using relevant clinical indicators (e.g., MMSE score, GDS, gait metrics, postural instability), and the final answer in \texttt{\textbackslash boxed\{0 or 1\}} within \texttt{<answer>} tags.

    \vspace{0.5em}
    === Clinical notes (EN) === \\
    \texttt{<CLINICAL\_NOTE>}
    \end{tcolorbox}
    \caption{Fine-tuning prompt template for the HPH dataset. The task requires predicting Parkinson's disease dementia (PDD) progression based on longitudinal clinical notes including cognitive assessments, motor symptoms, and gait analysis.}
    \label{fig:prompt_haeundae}
\end{figure*}

\begin{figure*}[!t]
    \centering
    \begin{tcolorbox}[colback=gray!5!white, colframe=gray!75!black, title={Fine-tuning Task: ADNI Cohort (MCI Conversion Prediction)}, fonttitle=\bfseries]
    \small
    \textbf{[System Prompt]} \\
    A conversation between a User and an Assistant. The User provides longitudinal clinical assessment data and metadata describing a patient's cognitive and functional trajectory. The Assistant must determine whether the patient has progressed from a baseline status of Mild Cognitive Impairment (MCI) to dementia by the time of the final diagnosis. Output 0 if the final diagnosis is non-dementia (MCI or CN), and 1 if the patient has converted to dementia. Use trends across longitudinal data (cognition, function, severity scores) for reasoning. Respond only in the specified \texttt{<think>} and \texttt{<answer>} format.

    \vspace{0.8em}
    \hrule
    \vspace{0.8em}
    
    \textbf{[Input Data]} \\
    \textbf{Instruction:} You are given longitudinal clinical assessment data for a patient with baseline MCI. Records are collected before a cutoff set prior to the patient's last diagnostic assessment. The interval between the last diagnosis and the most recent visit is: \texttt{<TIME\_INTERVAL>} (e.g., 6--12m). \\
    \textbf{Task:} Predict whether the patient has progressed to dementia (0: non-dementia, 1: converted). \\
    \textbf{Format:} Output step-by-step reasoning in \texttt{<think>} tags and the final answer in \texttt{\textbackslash boxed\{0 or 1\}} within \texttt{<answer>} tags.

    \vspace{0.5em}
    \textbf{Variable Glossary:} \\
    - \texttt{PTEDUCAT/APOE4}: Education years / Number of APOE $\epsilon$4 alleles. \\
    - \texttt{CDRSB/ADAS13/MMSE/MOCA}: Clinical severity and cognitive scores (Higher CDRSB/ADAS = worse; Higher MMSE/MOCA = better). \\
    - \texttt{RAVLT/LDELTOTAL}: Memory scores (Lower = poorer memory). \\
    - \texttt{FAQ}: Functional Activities Questionnaire (Higher = worse daily function). \\
    - \texttt{ABETA/TAU/PTAU}: CSF biomarkers for amyloid and tau pathology. \\
    - \texttt{Ventricles/Hippocampus/WholeBrain}: MRI volumetric measures (Structural atrophy).
    
    \vspace{0.5em}
    === Clinical Assessment Data === \\
    \texttt{2011-11-28: <<<VISIT 1/7>>>} \\
    \texttt{CDRSB: 0.5, ADAS13: 14.0, MMSE: 28, FAQ: 0, Hippocampus: 6521, ...} \\
    \texttt{-----(Longitudinal history continues)-----}
    \end{tcolorbox}
    \caption{Fine-tuning prompt template for the ADNI dataset. The model predicts MCI-to-dementia conversion using longitudinal trends of clinical indices and biomarkers.}
    \label{fig:prompt_adni}
\end{figure*}

\begin{figure*}[!t]
    \centering
    \begin{tcolorbox}[colback=gray!5!white, colframe=gray!75!black, title={Diagnostic Rationale Generation}, fonttitle=\bfseries]
    \small
    \textbf{[System Prompt]} \\
    You are an AI assistant that generates step-by-step reasoning paths.

    \vspace{0.8em}
    \hrule
    \vspace{0.8em}
    
    \textbf{[User Prompt]} \\
Problem:
\{problem\}

Answer:
\{answer\}

Task: Generate a clear step-by-step reasoning path that explains how to solve the problem and arrive at the answer.

Reasoning:
    \end{tcolorbox}
    \caption{Prompt template for generating diagnostic rationales for Supervised Fine-Tuning.}
    \label{fig:prompt_rationale_gen}
\end{figure*}

\end{document}